\documentclass{article}

\usepackage{PRIMEarxiv}
\usepackage{indentfirst}
\usepackage{amsmath}
\usepackage{algorithm}
\usepackage{algorithmic}
\usepackage[utf8]{inputenc} 
\usepackage[T1]{fontenc}    
\usepackage{hyperref}       
\usepackage{url}            
\usepackage{booktabs}       
\usepackage{amsfonts}       
\usepackage{nicefrac}       
\usepackage{microtype}      
\usepackage{lipsum}
\usepackage{fancyhdr}       
\usepackage{graphicx}       
\usepackage{svg}
\usepackage{subfig}
\graphicspath{{media/}}     

\title{A Knowledge Distillation-Based Backdoor Attack in Federated Learning
}

\author{
  Yifan Wang \\
  School of Computer Science and Technology \\
  University of Science and Technology of China \\
  Hefei Anhui\\
  \texttt{yifanW@mail.ustc.edu.cn} \\
  \And
  Wei Fan \\
  School of Computer Science and Technology \\
  University of Science and Technology of China \\
  Hefei Anhui\\
  \texttt{slimfun@mail.ustc.edu.cn} \\
  \And
  Keke Yang \\
  School of Computer Science and Technology \\
  University of Science and Technology of China \\
  Hefei Anhui\\
  \texttt{ykk@mail.ustc.edu.cn} \\
  \And
  Naji Alhusaini \\
  School of Computer Science and Technology \\
  University of Science and Technology of China \\
  Hefei Anhui\\
  \texttt{husaini@ustc.edu.cn} \\
  \And
  Jing Li \\
  School of Computer Science and Technology \\
  University of Science and Technology of China \\
  Hefei Anhui\\
  \texttt{lj@ustc.edu.cn} \\
}

\begin{document}
\maketitle

\begin{abstract}

Federated Learning (FL) is a novel framework of decentralized machine learning. Due to the decentralized feature of FL, it is vulnerable to adversarial attacks in the training procedure, \eg, backdoor attacks. A backdoor attack aims to inject a backdoor into the machine learning model such that the model will make arbitrarily incorrect behavior on the test sample with some specific backdoor trigger. Even though a range of backdoor attack methods of FL has been introduced, there are also methods defending against them. Many of the defending methods utilize the abnormal characteristics of the models with backdoor or the difference between the models with backdoor and the regular models. To bypass these defenses, we need to reduce the difference and the abnormal characteristics. We find a source of such abnormality is that backdoor attack would directly flip the label of data when poisoning the data. However, current studies of the backdoor attack in FL are not mainly focus on reducing the difference between the models with backdoor and the regular models.

In this paper, we propose Adversarial Knowledge Distillation(ADVKD), a method combine knowledge distillation with backdoor attack in FL. With knowledge distillation, we can reduce the abnormal characteristics in model result from the label flipping, thus the model can bypass the defenses. Compared to current methods, we show that ADVKD can not only reach a higher attack success rate, but also successfully bypass the defenses when other methods fails. To further explore the performance of ADVKD, we test how the parameters affect the performance of ADVKD under different scenarios. According to the experiment result, we summarize how to adjust the parameter for better performance under different scenarios. We also use several methods to visualize the effect of different attack and explain the effectiveness of ADVKD.

\end{abstract}


\section{Introduction}

Federated Learning(FL)\cite{FedAVG} is a new framework of decentralized model training. In FL, participants can use their local training data to collaboratively train a machine learning model while keeping the local training data never leaves its owner. As FL utilizes participants' private data, the final model often has a better performance. And because the local training data never leaves its owner, FL prevents the leakage of local training data and thus protects the privacy of participants. Benefit from this, there are several practical applications of FL in high privacy requirements areas, such as speech recognition\cite{FL_APP_SpeechRecognition_1}, text prediction on mobile device\cite{FL_APP_WordPrediction_1,FL_APP_WordPrediction_2} and some medical applications\cite{FL_APP_Medical_1,FL_APP_Medical_2,FL_APP_Medical_3}.

Despite this, as FL can not guarantee that all the participants are honest, many studies show FL is vulnerable to attacks from malicious participants \cite{How2Backdoor,AdversarialLens,DBA,EdgecaseAttack} in the training procedure. The attacks of FL training can be classified into untargeted attacks and targeted attacks. The goal of the untargeted attacks is to damage the machine learning model's performance. Conversely, the targeted attack or backdoor attack aims to inject a backdoor into the model by submitting abnormal updates. The backdoor in the model can control the output of the victim model on the test sample with some specific trigger, and the victim model with backdoor can still perform regularly on the original FL task.

The adversary in the backdoor attack of FL can not directly modify the global model, so it can only submit the local model updates with backdoor to indirectly affect the global model. Because of the backdoor, these model updates may show some differences comparing with regular model updates. On the other hand, many current studies on defending backdoor attacks\cite{Krum, Bulyan, FLAME, FoolsGold} focus on discriminate the model updates with backdoor by identifying their abnormal characteristics. So if the characteristics of the model updates with backdoor are similar to a regular model update, it would be hard for the defense on server to identify them. Even the server can discard all the updates to avoid backdoor, the global model would never be updated, which is conflict with the target of FL. According to this, the model updates with backdoor can bypass the defense by reducing their abnormal characteristics. However, existing methods are not mainly focus on reducing the abnormal characteristics in model updates. We theoretically and experimentally found that current backdoor attack methods often directly flip the label when poisoning the local dataset, which would cause the neurons important to original task get punished and thus the model would move towards the direction contrary to normal training. Inspired by this finding, we propose adversarial knowledge distillation(ADVKD), introduce the knowledge distillation into the scenario of the backdoor attack in FL to reduce the abnormal characteristics in the model updates with backdoor. We design two attack strategy to handle the backdoor attack tasks under different scenarios. The experiment results on different datasets and different defending methods have shown that the proposed method can outperform the baselines. Besides, we also experimentally show that ADVKD does reduced the abnormal characteristics in the model updates with backdoor.

We summarize our contributions as follows:
\begin{itemize}
  \item We theoretically and experimentally analyze the reason why the updates from malicious participants would show some abnormal characteristics and fail to bypass defenses.
  \item We bridge knowledge distillation with backdoor attack in FL setting and propose a novel backdoor attack method ADVKD, which can reduce the difference between the updates from malicious participants and the updates from benign participants so that the backdoor attack can bypass the defenses/robust aggregations on the server.
  \item According to the experiment results on different datasets with different aggregation methods, we find that our method can successfully bypass the defenses/robust aggregations on the server. Comparing with the results of other backdoor attack methods, we find that ADVKD is more effective and stealthy, as its model update is more similar to a regular model update.
  \item We analyze the effect of the parameters of ADVKD by experiments, and summarize the method of adjusting the parameters of ADVKD under different scenarios for better performance.
  \item We visualize and compare the effect of different attack methods on the local model update or global model by several ways, and the results explain the effectiveness of ADVKD.
\end{itemize}

\section{Background and Related Work}

\subsection{Federated Learning}

In traditional machine learning, datasets are gathered together for further processing and training. However, gathering all the data in a central database is undesirable for security and privacy reasons. To solve this contradiction, Google has proposed a new machine learning framework called federated learning\cite{FedAVG}. In federated learning, each participant uses the local dataset to train the local model and sends the updated model parameters of the local model to the server for aggregation to a new global model. Then, the server sends the new global model to participants as their local model in the next round of training. This procedure repeats until the model converges. Overall, the target of federated learning is to minimize $f(\omega)$, \ie the average of the loss functions of the global model on each participant:
\begin{equation}
  \min_{\omega}{f(\omega)} = \frac{1}{N}\sum_{i=1}^{N}{f_{i}(\omega)}
\end{equation}
\begin{equation}
  f_{i}{(\omega)} = \sum_{j=1}^{|D_{i}|}{L(x_{i,j}, y_{i,j}; \omega)}
\end{equation}
where $N$ is the number of participants, $\omega$ is parameters of global model. $D_{i}$ is the private dataset of participant i and $(x_{i,j}, y_{i,j})$ are data points in the dataset. $f_{i}$ is the loss on $i^{th}$ participant with loss function $L$, \eg cross entropy loss or other loss functions.

In federated learning with N participants and a central server, the training procedure in each round can be divided into the following steps:
\begin{itemize}
  \item \textbf{Participant Selection:} Denote t as current iteration round, server select m participants and send current global model parameters $\omega^{t}$ to the subset of participants.
  \item \textbf{Local Training:} For each selected participant $i \in \{1...m\}$, denote $D_{i}$ and $\omega^{t}_{i}$ as local dataset and local model of participant i. After receive a new global model $\omega^{t}$ from the server, selected participants apply parameters of $\omega^{t}$ to local model and train it with their local datasets. After training, each participant get the updated local model with parameters $\omega^{t+1}_{i}$. Finally, each participant send the update of model parameters $\Delta\omega^{t+1}_{i} = \omega^{t+1}_{i} - \omega^{t}$ to server.
  \item \textbf{Model Aggregation:} After the server has received updates from all the selected participants, the server aggregate all the updates together to get new global model parameters $\omega^{t+1}$.
  \begin{equation}
    \omega^{t+1} = \omega^t + \frac{\eta}{m}\sum^{m}_{i=1}{\Delta\omega^{t+1}_{i}}
  \end{equation}
  where $\eta$ is the server learning rate.
\end{itemize}

\subsection{Backdoor Attacks}

In the backdoor attack of deep learning, the adversary has two targets: keeping the overall performance of model and keeping model outputs adversary-desired result (\eg an adversary-selected label or wrong label) when the model receives samples with backdoor trigger (\eg an input picture with some specific pattern). In existing works\cite{GU_BADNETS,CHEN_DataPoisoning,LIU_Trojaning}, they have proposed different ways to transform pristine sample into a new sample with backdoor features and join them into training procedure to achieve the adversary's goal. As the deep learning model often works like a black box, even though the model has been injected with backdoor, it is still hard for a user to detect the backdoor in the model without any knowledge of the backdoor trigger. The optimization target of the adversary is:

\begin{equation}
  \min_{\omega}((1 - \alpha)\sum_{i=1}^{|D|}{L(x_{i},y_{i};\omega)} + \alpha\sum_{i=1}^{|D|}{L(R(x_{i}, y_{i});\omega)}) \label{backdoor_optimization_problem}
\end{equation}
where $\alpha$ is a trade-off parameter that determines the adversary cares more about the accuracy on original task or backdoor task, $D$ is the training dataset, $\omega$ is the parameters of the model, $L$ is the loss function and $R$ is the data poisoning function which can convert a pristine sample into a poisoned sample.

Different works on the backdoor attack of traditional deep learning have different scenarios. Several works assume the adversarial can control the training dataset and have full access to the training procedure (\eg, the outsourced training scenario mentioned in Gu \etal\cite{GU_BADNETS}). Other works assume the adversary can not access the training dataset but can only access the trained public model (\eg, the autonomous driving scenario in Liu \etal\cite{LIU_Trojaning}). In contrast, some works consider a stricter condition: the adversary has no knowledge of the model and its parameters, training procedure, and training dataset but can only insert a few poisoned samples into the training dataset to achieve the backdoor goals (\eg, Chen \etal\cite{CHEN_DataPoisoning}).

In the federated learning scenario, the adversary can only control a fraction of participants, which leads to a new problem in backdoor attack. In every round of federated learning, the updates with backdoor submitted by the adversary would be aggregated with other updates from benign participants on the server. Thus, backdoor parameters would be scaled down and perturbed by benign updates, damaging backdoor performance on the final global model. And as the participants in a single round of federated learning are randomly selected, it's possible that no adversary-controlled participant being selected in some rounds. Thus, the global model would forget the previously injected backdoor after these rounds. Bagdasaryan \etal\cite{How2Backdoor} proposed an attacking method that replaces the global model with an adversarial-desired one after single round by directly scaling up the update by a parameter $\gamma$ before submitting it. Bhagoji \etal\cite{AdversarialLens}'s method modifies the optimization target in (\ref{backdoor_optimization_problem}), changes the weight of model performance $(1-\alpha)$ to $1$ and change the weight of backdoor performance $\alpha$ to one likes $\gamma$ in \cite{How2Backdoor} to ensure the backdoor-related parameter updates can resist the effect of scaling down and benign updates. Xie \etal\cite{DBA} proposed distributed backdoor attack, which introduced the "distributed" feature into backdoor attack by dividing the original backdoor trigger into several smaller new backdoor triggers and distributing these new triggers to adversarial-controlled participants; different adversarial participants would conduct a backdoor attack with a different small trigger. This approach enhances stealth and persistence. Wang \etal\cite{EdgecaseAttack}'s work associates adversarial sample\cite{AdversarialSample} with backdoor attack, proofs the existence of backdoor attack and proposes an edge-case attack by using features of data distribution.

\subsection{Robust Federated Learning}

In federated learning, the server can not interfere with the local training procedure of participants as it would violate privacy. So the aggregation algorithm should be replaced with a more robust one to build a robust federated learning system. Many current works propose different robust aggregation algorithms. Some works apply geometric median\cite{GeoMed,RFA}, coordinate-median or trimmed mean\cite{TrimmedMean} in aggregation for more robustness towards outliers. Blanchard \etal\cite{Krum} proposed Krum, which views updates from participants as a vector and selects the most representative one (have the smallest sum of the Euclidean distance to nearest n-f-2 updates) as the aggregated update in this round. Its variety Multi-Krum broaden the range of selection to top-k representative updates and use the average of top-k as aggregated update. Mhamdi \etal\cite{Bulyan}'s method, called bulyan, combines Krum with trimmed mean by running Krum multiple times to select reliable updates and then running trimmed mean on these reliable updates to remove the effect of outliers. These algorithms aim to handle untargeted attacks which can damage model performance.

On the other hand, Fung \etal\cite{FoolsGold}'s FoolsGold algorithm calculates the cosine similarity among participants' updates and assigns different weights to different participants in weighted average to weaken the effect of similar updates, hence adversary can not use Sybil attack to affect the direction of model update. However, for coordinated adversarial participants, they can make an internal aggregation, then coordinate to generate several new model updates that have small cosine similarity with each other, but can be the same to the result of internal aggregation after aggregated. Sun \etal\cite{CanYou}'s work shows federated learning can be more robust by applying Norm-Clip and WeakDP. Norm-Clip would compare the norm of local model updates $\Vert\Delta\omega^{t}_{i}\Vert_2$ with a threshold $M$, then multiply the model updates with $\frac{M}{\Vert \Delta\omega^{t+1}_{i} \Vert_2}$ to limit their effect if their norm exceed the threshold. WeakDP is adding a Gaussian noise to the aggregated model update. However, if the threshold $M$ is too small or the noise is too large, it would damage the model performance. On the contrary, if the threshold $M$ is too large or the noise is too small, the defense would fail to defend attacks. For different models, different datasets and different hyperparameters, the suitable threshold $M$ and noise are different. Nguyen \etal\cite{FLAME} proposed FLAME framework, which can adaptively adjust the threshold $M$ and noise in every round. It also uses HDBSCAN\cite{HDBSCAN} with cosine distance as distance metric to clustering model updates, and only selects the largest cluster for following aggregation.

\section{Adversarial Knowledge Distillation}


\subsection{Attacker Ability}

In federated learning backdoor attacks, a fraction of participants are adversarial or their device (\eg, mobile phone) are compromised by the adversary. According to the assumption in \cite{How2Backdoor}, the adversary can (1) control the local training data on each compromised participant, (2) control the training procedure on each compromised participant, (3) modify the model update from these compromised participants before submitting it and (4) fetch information of current global model parameters to adapt the training procedure in each round.

On the other hand, the adversary cannot disturb any benign participant on their training procedure and update submission, nor the aggregation of the server. So we can assume the update from each benign participant is correct.

\subsection{Motivation}

\begin{figure}[htbp]
  \centering
  \subfloat[backdoor pattern]{
    \includegraphics[height=2.5cm]{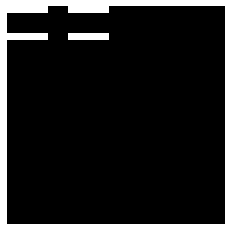}
    \label{fig:fig_backdoor_pattern_a}
  }

  \subfloat[clean samples and clean samples with backdoor pattern]{
    \includegraphics[height=3cm]{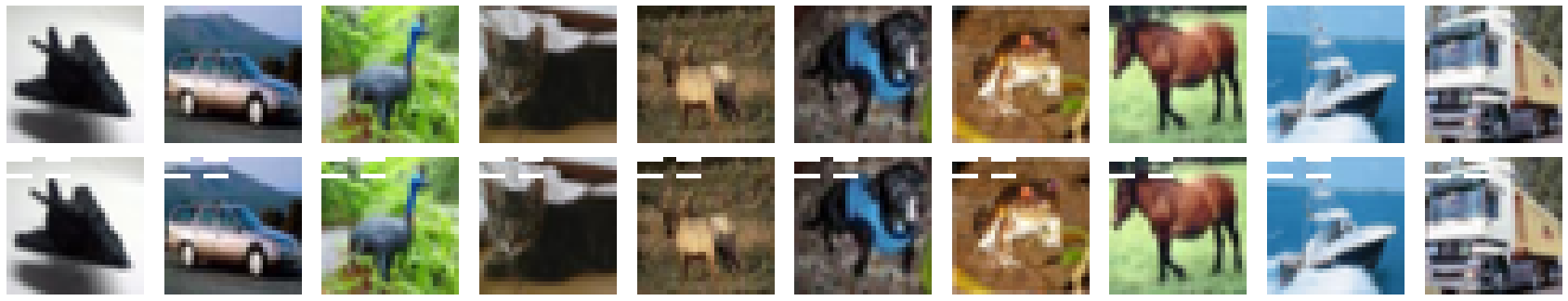}
    \label{fig:fig_backdoor_pattern_b}
  }

  \subfloat[Cat, Dog and two kinds of mixture of cat and dog]{
    \includegraphics[height=2cm]{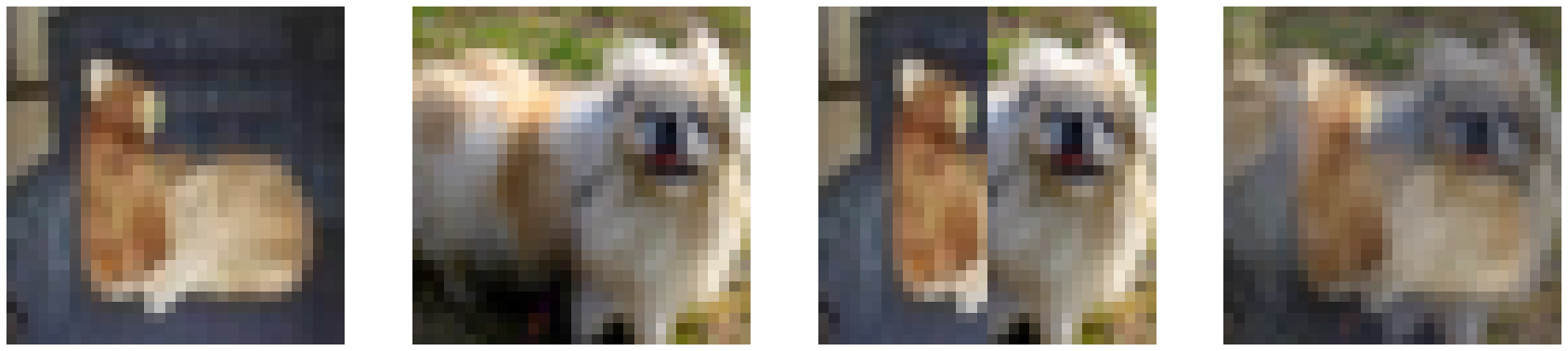}
    \label{fig:fig_backdoor_pattern_c}
  }
  \caption{Backdoor pattern and backdoor samples}
  \label{fig:fig_backdoor_pattern}
\end{figure}

In a backdoor attack, the adversary expect the model would return a specific result when receiving an input sample with a backdoor trigger. In other word, the adversary expect the model classifies the content of the backdoor trigger into the target class. For example, in image classification, if the adversary use pattern in Fig\ref{fig:fig_backdoor_pattern_a} as a backdoor trigger, the target of the adversary is to make sure the model would classify the pattern into the target class.

However, in an actual backdoor attack, the backdoor trigger is usually embedded into normal samples, not directly used as a sample, so the sample has normal and backdoor trigger content (\eg, Fig\ref{fig:fig_backdoor_pattern_b}). Such pictures can be viewed as a mixture of "backdoor pattern sample" and clean sample, which is similar to "mixup"\cite{MIXUP} or "cutmix"\cite{CUTMIX} in data augmentation, likes in fig \ref{fig:fig_backdoor_pattern_c}. In "mixup" and "cutmix", we can generate an interpolation of a different sample (mixup) or replace a portion of one sample with one from another sample (cutmix), then we also need to generate a new label for the new sample by "mix" their label together to present this new sample has features, semantic or information from original samples. Nevertheless, in a backdoor attack, the adversary often changes the label of a poisoned sample(embedded with backdoor trigger) to the adversary-desired one. So we can get an equation of gradient on a poisoned sample; here we set the loss function as CrossEntropyLoss and use one-hot code as a label.

\begin{equation}
  L(R(x, y); \omega) = - \sum_{i=1}^{N_{class}} y_{i} \log \frac{e^{l_{i}}}{\sum_{j=1}^{N_{class}}e^{l_{j}}} = - \log \frac{e^{l_{target}}}{\sum_{j=1}^{N_{class}}e^{l_{j}}}
  \label{BackdoorCrossEntropyLoss}
\end{equation}
\begin{equation}
  \frac{\partial L(R(x, y); \omega)}{\partial l_{target}} = \frac{e^{l_{target}} - \sum_{j=1}^{N_{class}}e^{l_{j}}}{\sum_{j=1}^{N_{class}}e^{l_{j}}} < 0
  \label{BackdoorCrossEntropyLoss_target}
\end{equation}
\begin{equation}
  \frac{\partial L(R(x, y); \omega)}{\partial l_{label}} = \frac{e^{l_{label}}}{\sum_{j=1}^{N_{class}}e^{l_{j}}} > 0
  \label{BackdoorCrossEntropyLoss_label}
\end{equation}
where $l_{i}$ is the logit of class $i$, $N_{class}$ is the number of classes, $y_{i}$ is the component of label (a one-hot code as previously defined) on class $i$, $target$ and $label$ is adversary-desired class and original class.

We can find in equation\ref{BackdoorCrossEntropyLoss_target} and equation\ref{BackdoorCrossEntropyLoss_label} that after the label was changed by adversary in backdoor attack, the gradient of loss function(CrossEntropyLoss) has the same direction with gradient of $l_{label}$ and has different direction with gradient of $l_{target}$. Hence, when we optimize the loss function by using gradient descent, the model's parameters would move toward the direction of increasing $l_{target}$, decreasing $l_{label}$. However, when a benign participant optimizing model, the model's parameters would move toward increasing $l_{label}$. As such difference exists, with training going on, the difference between model updates would get larger and larger.

\begin{figure}[hbtp]
  \centering
  \subfloat[Euclidean Distance]{
    \includegraphics[width=0.35\textwidth]{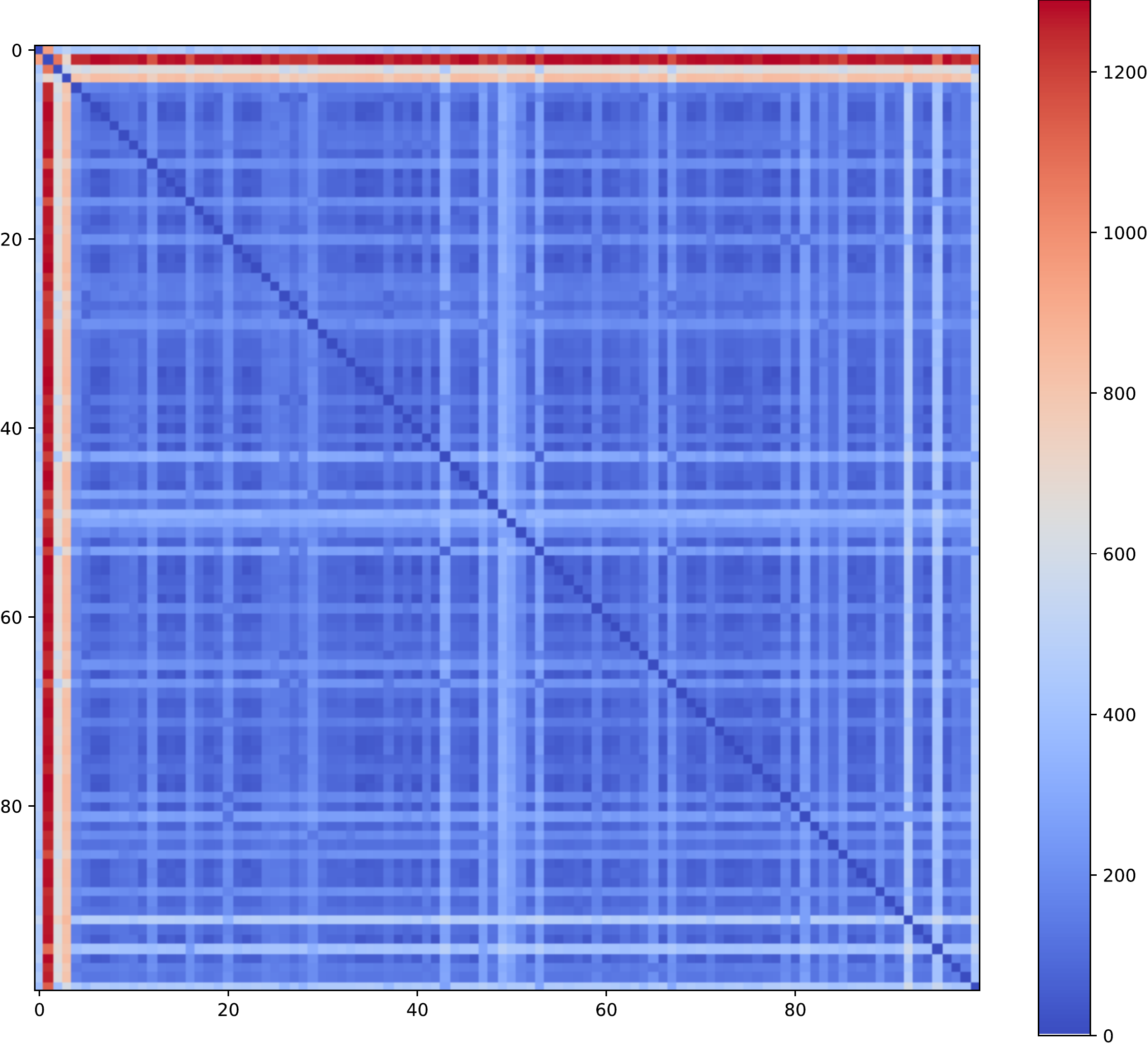}
    \label{fig:fig_dmat_CE}
  }
  \subfloat[Cosine Distance]{
    \includegraphics[width=0.35\textwidth]{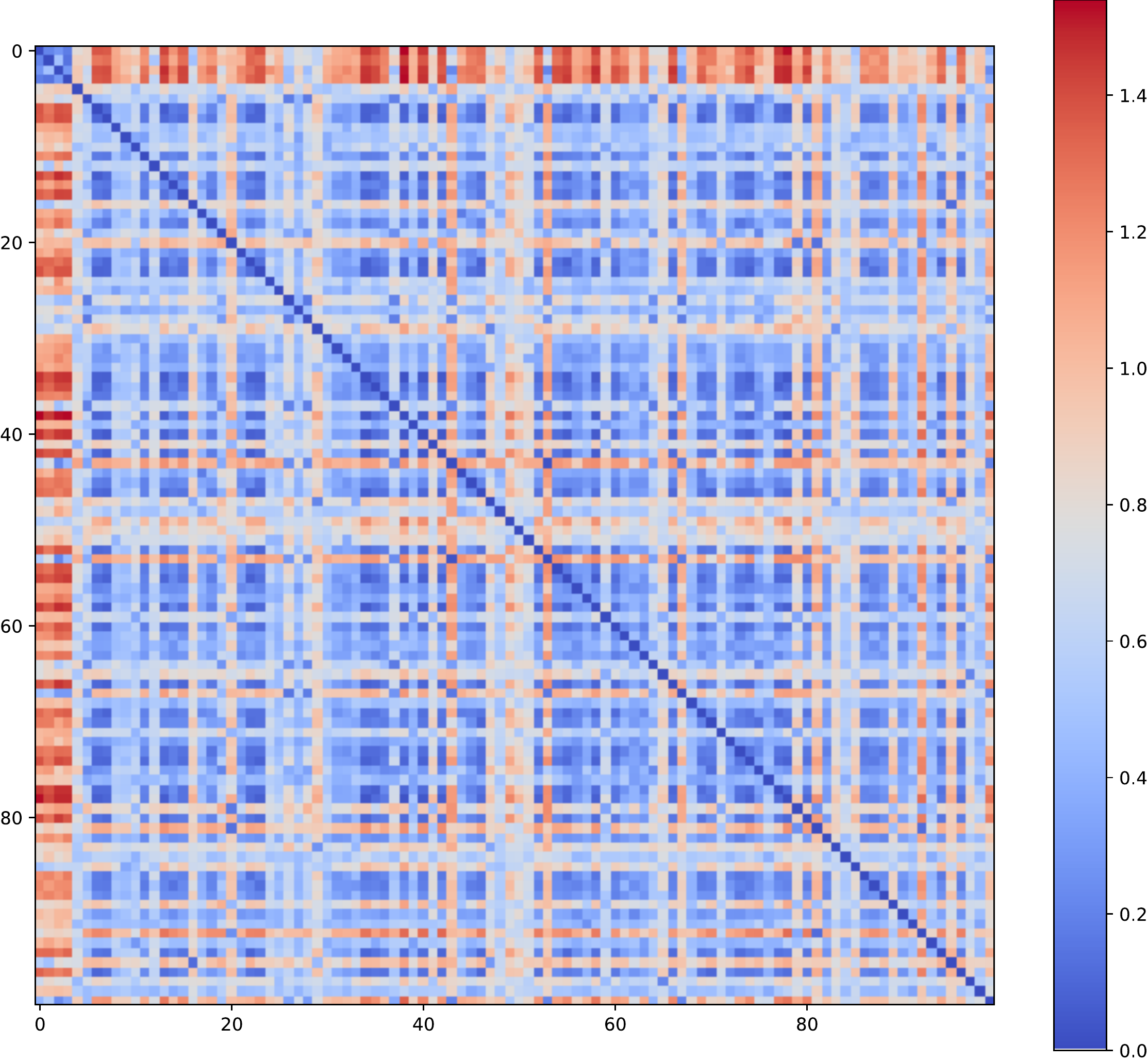}
    \label{fig:fig_cosmat_CE}
  }
  \caption{Heat Map of Euclidean distance and cosine distance between 100 model updates in the same round}
  \label{fig:fig_mat_CE}
\end{figure}

We can prove this problem by experiment. We use ResNet-18 with CIFAR10 dataset to simulate one single round of federated learning with 4 adversary participants and 96 benign participants (100 in total) and calculate the Euclidean distance and cosine distance between these model updates. The result is shown in Fig\ref{fig:fig_dmat_CE} and \ref{fig:fig_cosmat_CE}, the Euclidean distance and cosine distance between the first four updates (from adversary participants) and other updates is considerable, which confirms the problem previously mentioned does exist.

Both analysis and experiment above show that directly changing labels would damage the stealth of backdoor attack and make it more vulnerable to defenses on server. Motivated by this finding, we proposed a new backdoor attack method which can reduce the effect of this problem.

\subsection{Algorithm Design}

As mentioned above, directly flipping the label of the poisoned sample to the target label would cause the final model update to become an outlier. Inspired by this, we try to use knowledge distillation in training procedures to reduce the punishment on neurons related to the original label and get a smoother output, which we call adversarial knowledge distillation(ADVKD). In knowledge distillation\cite{KnowledgeDistillation}, there will be a teacher model to generate logit for each sample. Then the student model optimizes the KL-divergence loss to align model output and transfer knowledge into the student model. However, as an adversary in federated learning, it's hard to get a well-trained model as a teacher. Inspired by teacher-free knowledge distillation(Tf-KD)\cite{TeacherFreeKnowledgeDistillation}, we select the current global model (as we can not get a better pre-trained model) as the teacher to perform knowledge distillation with the local model. So we use the current global model parameters $\omega^{t}$ and the local dataset $D_{local}$ to generate logits as output of teacher model, and we combine it with the local dataset $D_{local}$ to generate a new local dataset with soft target $D_{local}^{soft}$. By using $D_{local}^{soft}$ we can calculate cross entropy and KL divergence loss and optimize both losses.

\paragraph{Implementation of ADVKD}

As shown in algorithm \ref{alg:ADVKD}, at the beginning of one round, we assign the current global model parameters to the local model (line 2). We use current global model and local dataset to generate corresponding logits and merge the logits into the local dataset to generate a new local dataset with a soft target for following knowledge distillation (line 3). In training, for each epoch, we would poison (part of) data in the batch and update the model parameters by optimizing model loss on the poisoned batch to guide the model towards higher backdoor accuracy. When training epochs finished, we calculate the difference between the local model parameters and the global model parameters and submit the difference as a model update.

\begin{algorithm}[htb]
  \caption{ Adversarial Knowledge Distillation(ADVKD) }
  \label{alg:ADVKD}
  \begin{algorithmic}[1]
    \STATE \textbf{INPUT:} Model $f(\cdot; \omega)$, model parameters $\omega^{t}$, local dataset $D_{local} = {(x_1, y_1), ..., (x_{|D_{local}|}, y_{|D_{local}|})}$
    \STATE $\omega_{i}^{t+1} \leftarrow \omega^{t}$
    \STATE $D^{soft}_{local}$ $ \leftarrow GenerateSoftTarget(f(\cdot; \omega), \omega^{t}, D_{local})$
    \FOR{$epoch \in [1..E]$}
      \FOR{each batch $\mathcal{B} = \{(x_1, y_1, y^{soft}_1), ..., (x_{b}, y_{b}, y^{soft}_{b})\} \subset D^{soft}_{local}$}
        \STATE $\mathcal{B}^{'} \leftarrow PoisonBatch(\mathcal{B})$
        \STATE $\omega_{i}^{t+1} \leftarrow \omega_{i}^{t+1} - lr\cdot\nabla((1-\alpha)L_{CE}(\mathcal{B^{'}}; \omega_{i}^{t+1}) + \alpha  L_{KD}(\mathcal{B}^{'}; \omega_{i}^{t+1}))$
      \ENDFOR
    \ENDFOR
    \RETURN $\omega_{i}^{t+1} - \omega^{t}$
  \end{algorithmic}
\end{algorithm}

\paragraph{Generating Poisoned Soft Target}

In ADVKD, for every batch, we will poison part of the samples in the batch and change the corresponding label into the target of an adversary, which is similar to the data poisoning procedure of previous works\cite{GU_BADNETS,DBA}. However, in our method, a batch not only contains sample data and labels but also holds the soft targets for knowledge distillation, and the value of soft targets are set to the logits produced by the current global model by default. It becomes a new problem to handle the soft target poisoning. We propose two different ways to solve this problem:
\begin{itemize}
  \item \textbf{ADVKD-REG(Regularization)}: Do not make any modifications on soft targets. By using the logit produced by the current global model to conduct knowledge distillation(optimizing KL-divergence loss) as a regularization, we can prevent the local model bias too much to backdoor target so that the model update would not become an outlier and thus keep the stealth of backdoor attack. Nevertheless, it might also weaken the effect of the backdoor and cause the attack fails.
  \item \textbf{ADVKD-ENH(Enhancement)}: Do some modification on the original value of soft targets (logit of global model) to boost the performance of backdoor. For the model in Tf-KD, except for conducting knowledge distillation with the model itself (self-training/Tf-KD$_{self}$), we can also conduct knowledge distillation with a manually-designed model output (Tf-KD$_{reg}$) which is similar to the smoothed label in Label Smoothing Regularization(LSR)\cite{LabelSmoothingRegularization}. Inspired by this method, we can modify the soft target of the poisoned part of batch, \eg increasing its value on backdoor target, to encourage model gains better backdoor accuracy. But if the modification method is not appropriate, this method would degenerate to traditional minimizing cross entropy loss. So, it needs more experiment results to determine the detail modification method and its parameters.
\end{itemize}

For the second method, we design a detail strategy to generate the poisoned soft target: let $l^{clean}$ and $l^{poison}$ be the logits of current model on original sample and poisoned sample, $l_{i}$ be the logit on $i_{th}$ class, $l_{target}$ and $l_{label}$ be the logit on backdoor target class and original label, the adversary can generate poisoned soft target $y_{i}^{soft\_poison}$ as described in equation\ref{PoisonedSoftTargetGen}, in which $\gamma$ and $\beta$ are parameters.

\begin{equation}
  y_{i}^{soft\_poison} = \begin{cases}
    l_{i}^{clean} & i \neq target \\
    \\
    \begin{aligned}
      & l_{label}^{clean} + \\
      & max((l_{label}^{clean} - \min_{j}(l_{j}^{clean})) * \gamma + (l_{target}^{poison} - l_{label}^{poison}), (l_{label}^{clean} - \min_{j}(l_{j}^{clean})) * \beta)
    \end{aligned} & i = target
  \end{cases}
  \label{PoisonedSoftTargetGen}
\end{equation}

In the poisoned soft target mentioned above, the value of soft targets on non-backdoor-target class keeps original logit but the value of soft targets on backdoor target class is the sum of logit on original label with an increment to encourage model moves toward the adversary-desired one. The increment is the larger one between two values. The first value $(l_{label}^{clean} - \min_{j}(l_{j}^{clean})) * \gamma + (l_{target}^{poison} - l_{label}^{poison})$ is consists of two parts: $(l_{target}^{poison} - l_{label}^{poison})$ presents the difference between the logit of poisoned sample on backdoor target and original label, add this value to logit of original label could introduce the relation of the logit of backdoor target and original label on poisoned sample into soft target, then we add $(l_{label}^{clean} - \min_{j}(l_{j}^{clean})) * \gamma$, which is a non-negative value that can strengthen the backdoor in model to be better than the backdoor in current model and the $\gamma$ can be viewed as the step length of strengthen. The second value $(l_{label}^{clean} - \min_{j}(l_{j}^{clean})) * \beta$ is always a non-negative value and can keep $y_{target}^{soft\_poison}$ not smaller than $y_{label}^{soft\_poison}$, especially when $(l_{target}^{poison} - l_{label}^{poison})$ is negative, so that can keep model holds backdoor after training.

As mentioned above, different from directly generating probability distribution by labels in Tf-KD$_{reg}$, we decide the soft target of a sample by considering output of current global model on this sample, which can not only control the strength of backdoor in model to avoid model update becomes outlier by adjusting $\gamma$ to adjust training step, but also keep the difference of samples of the same class by using logit.

\section{Experiments}

\subsection{Experiment Setup}

\subsubsection{Datasets and Models}

We demonstrate three public datasets, including Fashion-MNIST dataset, EMNIST dataset and CIFAR10 dataset. For Fashion-MNIST and EMNIST dataset, we use a CNN model with two convolutional layers followed by two fully connected layers. For CIFAR10 dataset, we use ResNet-18. In the experiments of Fashion-MNIST and EMNIST dataset, according to the conclusion in \cite{How2Backdoor} that it's better to attack when the global is converging, we start the attack in 10th round as the model is about to converge. In the experiments of CIFAR10 dataset, we use a pre-trained model in the beginning. The details of these datasets are shown in Table \ref{tab:datasets}.

\begin{table}[htbp]
  \caption{Details of each dataset}
    \centering
    \begin{tabular}{lllll}
      \toprule
      Datasets      & Model     & Number of Classes & Training/Testing Examples & Data shape \\
      \midrule
      Fashion-MNIST & CNN       & 10                & 60000/10000   & 28*28 \\
      EMNIST        & CNN       & 10                & 240000/40000  & 28*28 \\
      CIFAR10       & ResNet-18 & 10                & 50000/10000   & 32*32*3 \\
      \bottomrule
    \end{tabular}
    \label{tab:datasets}
  \end{table}

In experiments, we assume there are 100 participants in total and divide dataset into 100 local datasets, we set the number of participants to be selected in each round of federated learning to 12. When dividing dataset into small local datasets of participants, we consider Non-IID scenario as it is close to the realistic scenario. For the Non-IID scenario, the Dirichlet distribution is used to divide training data to guarantee the heterogeneous in data distribution, and its parameter $\alpha$ is set to 0.5 by default in following experiments.

\subsubsection{Backdoor Attacks, Defending Methods and Metrics}

In experiments, we consider two baseline methods: a naive method introduced in \cite{How2Backdoor} and the DBA method introduced in \cite{DBA}. Similar to the setting in \cite{DBA}, for DBA method, we set the number of adversary to 4 and also divide the backdoor trigger into 4 parts, for centralized methods, we set the number of adversary to 1 by default. Two different backdoor attack scenarios are mentioned in \cite{How2Backdoor}: single-shot attack and multi-shot attack. Here we only consider the multi-shot attack scenario as the adversary participants in single-shot scenario would multiply model update with a large factor to get a better backdoor performance in the only chance, and it would be detected by some defending methods, \eg Krum\cite{Krum}, which is contrary to our target of avoid becoming an outlier. To perform the difference of different methods in a shorter time, we conduct a complete attack procedure in every round.

When evaluating the performance of ADVKD, we will apply ADVKD-ENH and ADVKD-REG to two baseline methods for evaluation.

As mentioned above, the aggregation procedure on server may not only be FedAvg but some more robust methods. So in this paper, we consider Multi-Krum\cite{Krum} and FLAME\cite{FLAME}, and evaluate the performance of our method and  other baselines under such robust aggregation.

We use the attack success rate(ASR) to evaluate the performance of attack, as shown in equation\ref{ASR}.

\begin{equation}
  ASR(\omega, D_{test}, R) = \frac{1}{|D_{test}|}\sum^{|D_{test}|}_{i=1}\mathbb{I}(g(R(x_{i});\omega) = y_{target})
  \label{ASR}
\end{equation}

where $\mathbb{I}$ is an indicator function which returns 1 when the condition is true otherwise be 0, $R$ is data-poisoning function which can inject backdoor pattern into clean samples, $g$ is the model and $\omega$ represents model parameters.

\subsection{Experiment Results}

\paragraph{No Defense(FedAvg) Scenario}

In the scenario of FedAvg without any defense, we conduct naive method, naive method with ADVKD, DBA and DBA with ADVKD on training. For the parameter $\alpha$ of ADVKD, we select 0.7, 0.5 and 0.3, and add $\alpha = 0.9$ for EMNIST dataset. For ADVKD-ENH, we set its parameters $\gamma = 2$, $\beta = 0.5$. We test the ASR of different attack method and their effect on the model's performance.

\begin{figure}[htbp]
  \centering
  \subfloat[ASR of Naive Methods]{
    \includegraphics[width=0.48\textwidth]{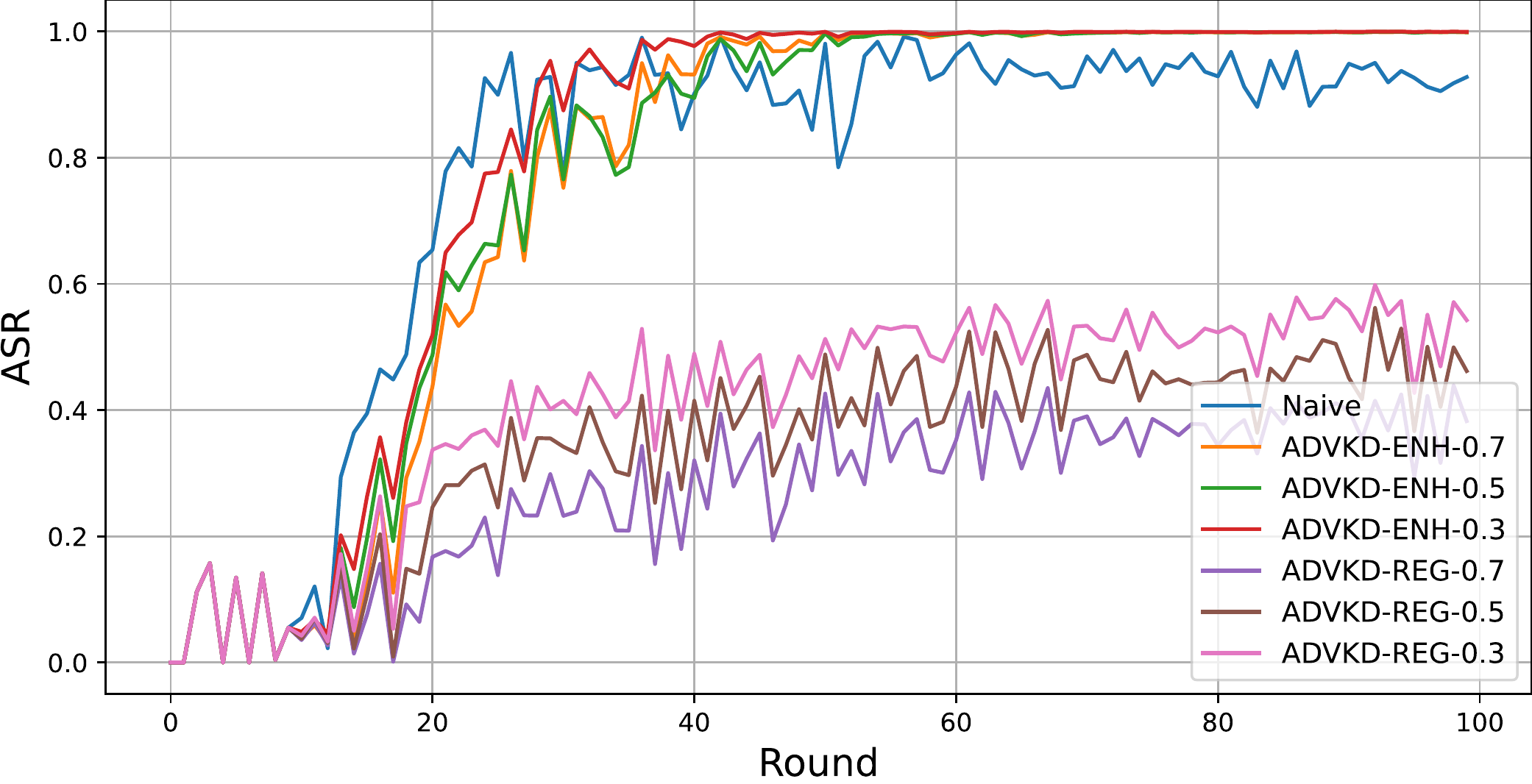}
    \label{fig:fig_Attack_NoDef_fmnist_a}
  }
  \subfloat[Accuracy of Naive Methods]{
    \includegraphics[width=0.48\textwidth]{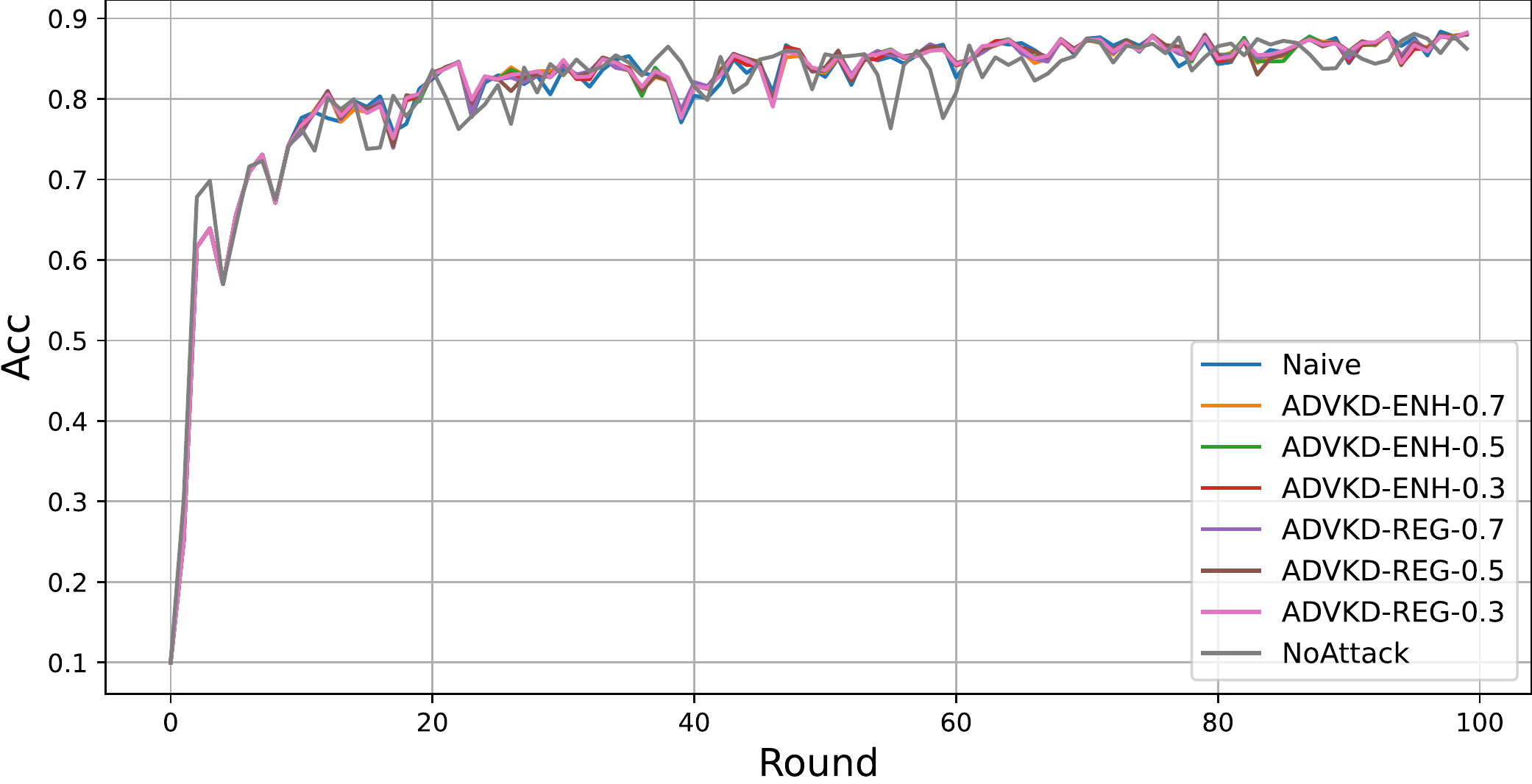}
    \label{fig:fig_Attack_NoDef_fmnist_acc_a}
  }
  
  \subfloat[ASR of DBA Methods]{
    \includegraphics[width=0.48\textwidth]{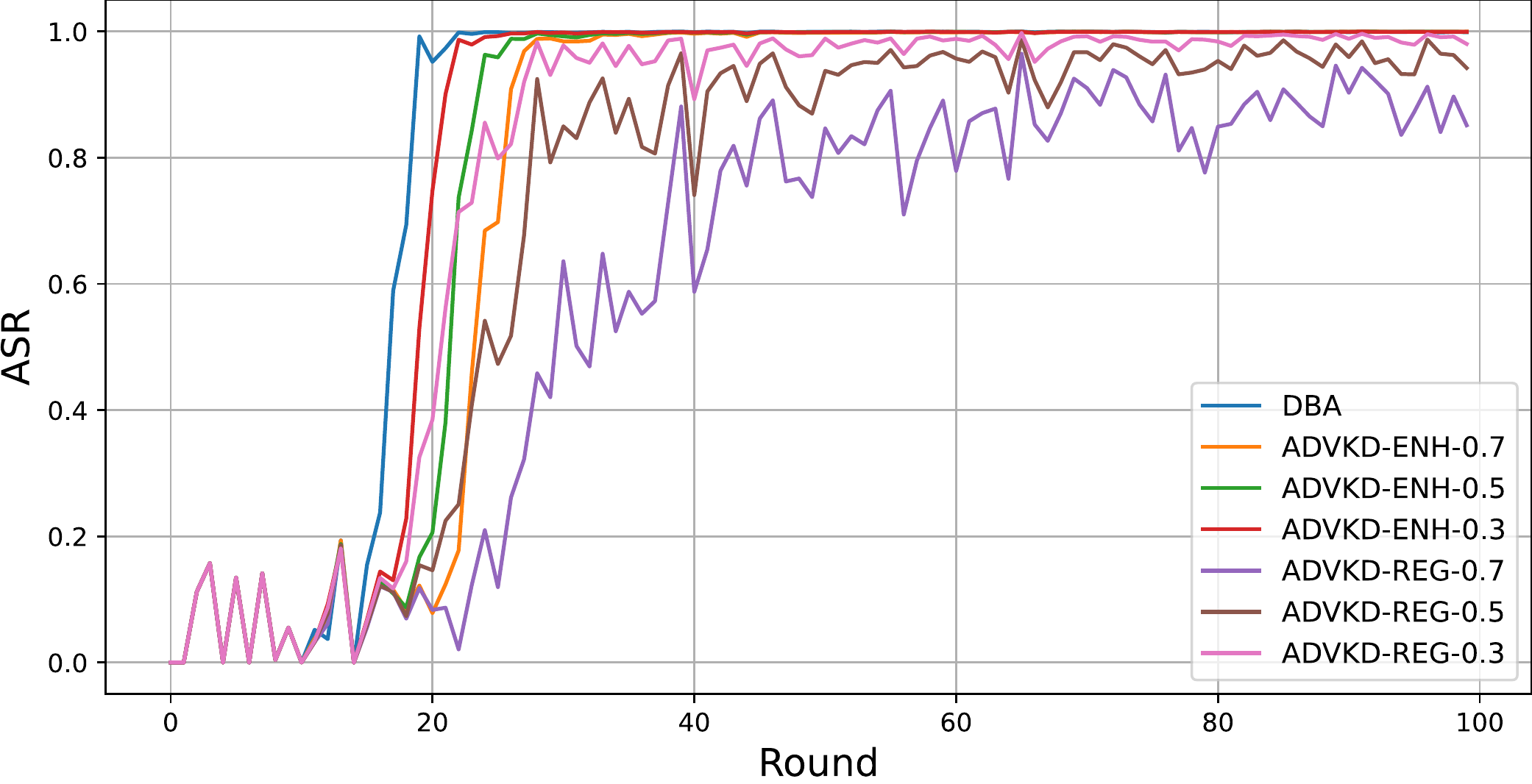}
    \label{fig:fig_Attack_NoDef_fmnist_b}
  }
  \subfloat[Accuracy of DBA Methods]{
    \includegraphics[width=0.48\textwidth]{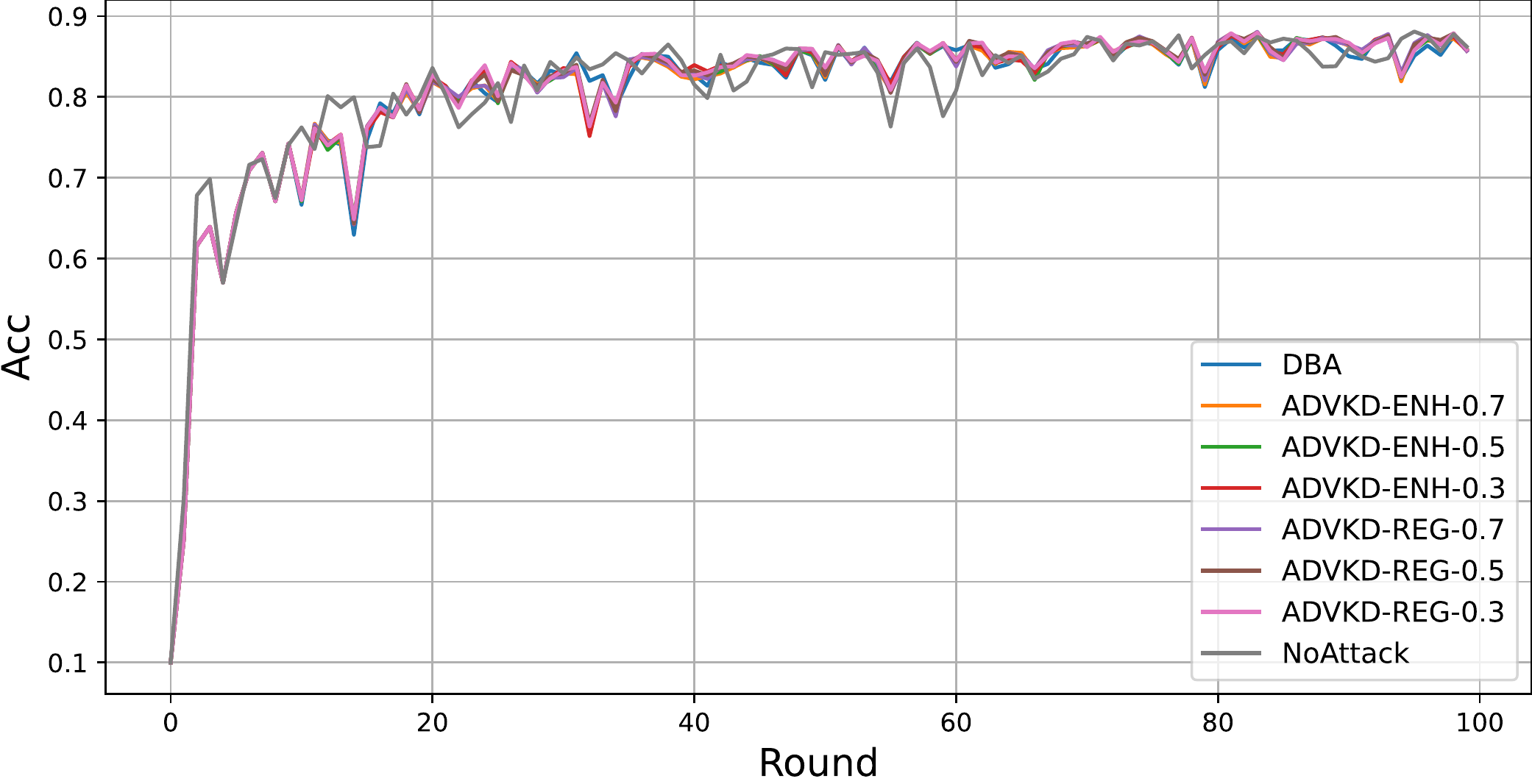}
    \label{fig:fig_Attack_NoDef_fmnist_acc_b}
  }
  \caption{Fashion-MNIST in FedAvg}
  \label{fig:fig_Attack_NoDef_fmnist}
\end{figure}

\begin{figure}[htbp]
  \centering
  \subfloat[ASR of Naive Methods]{
    \includegraphics[width=0.48\textwidth]{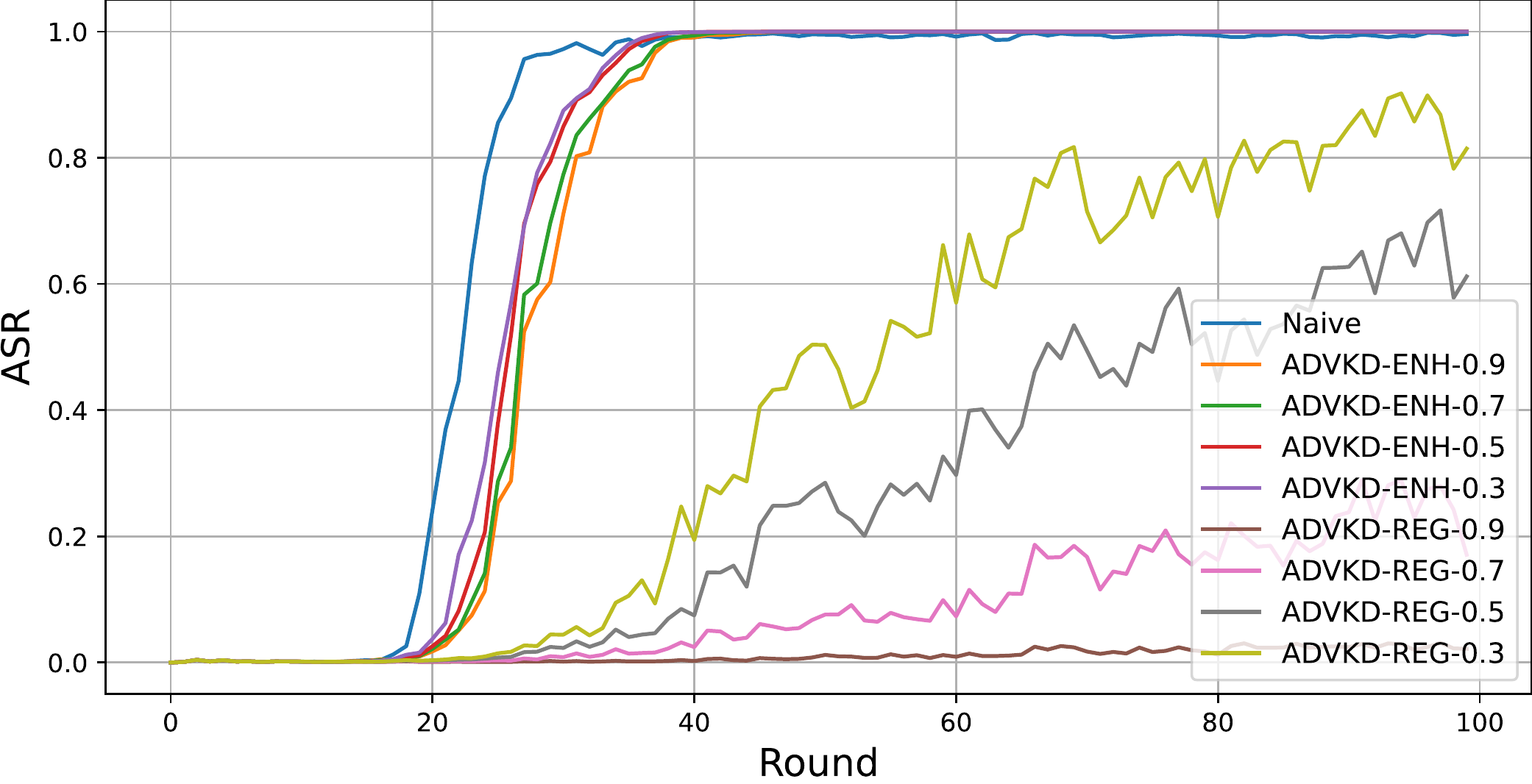}
    \label{fig:fig_Attack_NoDef_emnist_a}
  }
  \subfloat[Accuracy of Naive Methods]{
    \includegraphics[width=0.48\textwidth]{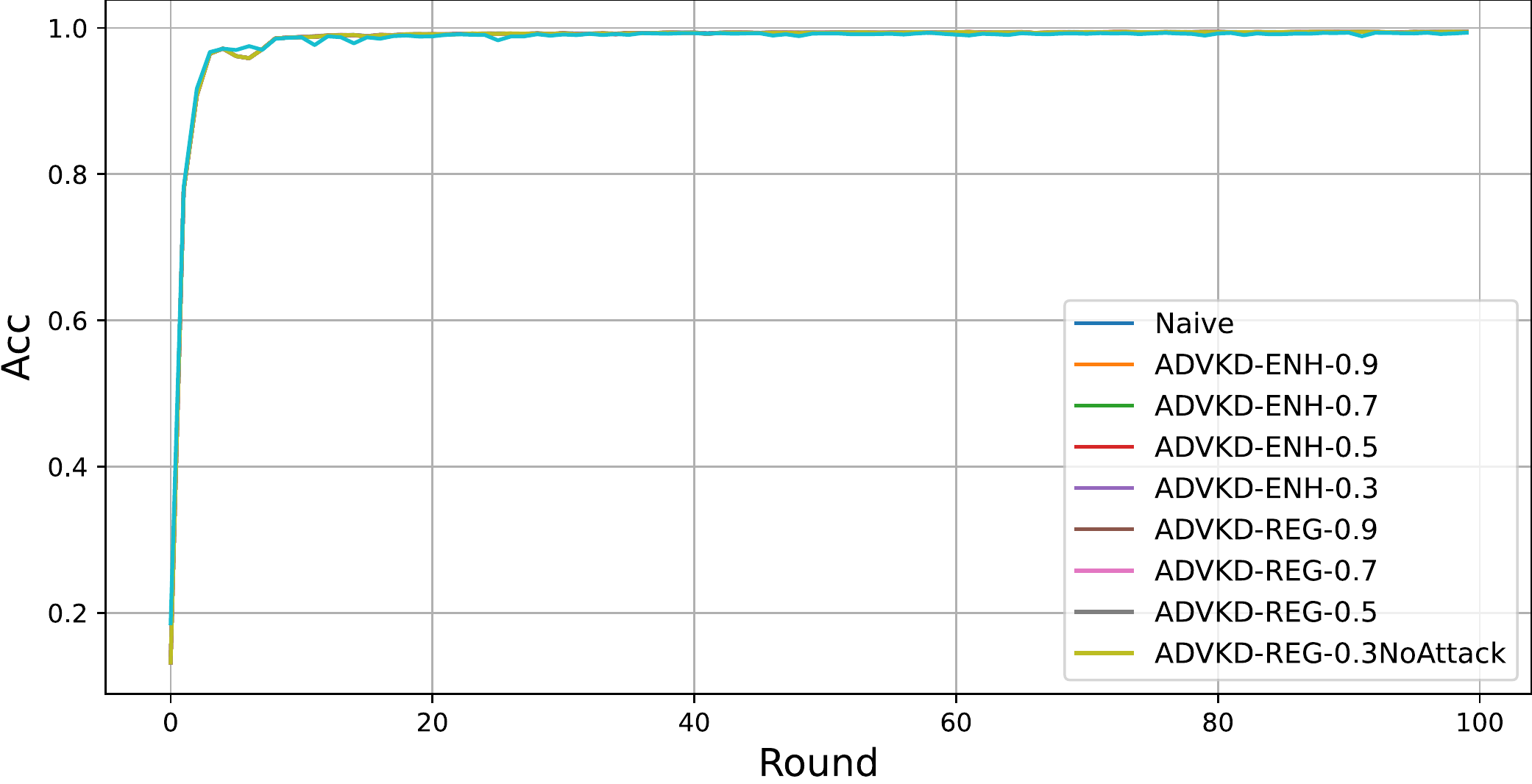}
    \label{fig:fig_Attack_NoDef_emnist_acc_a}
  }

  \subfloat[ASR of DBA Methods]{
    \includegraphics[width=0.48\textwidth]{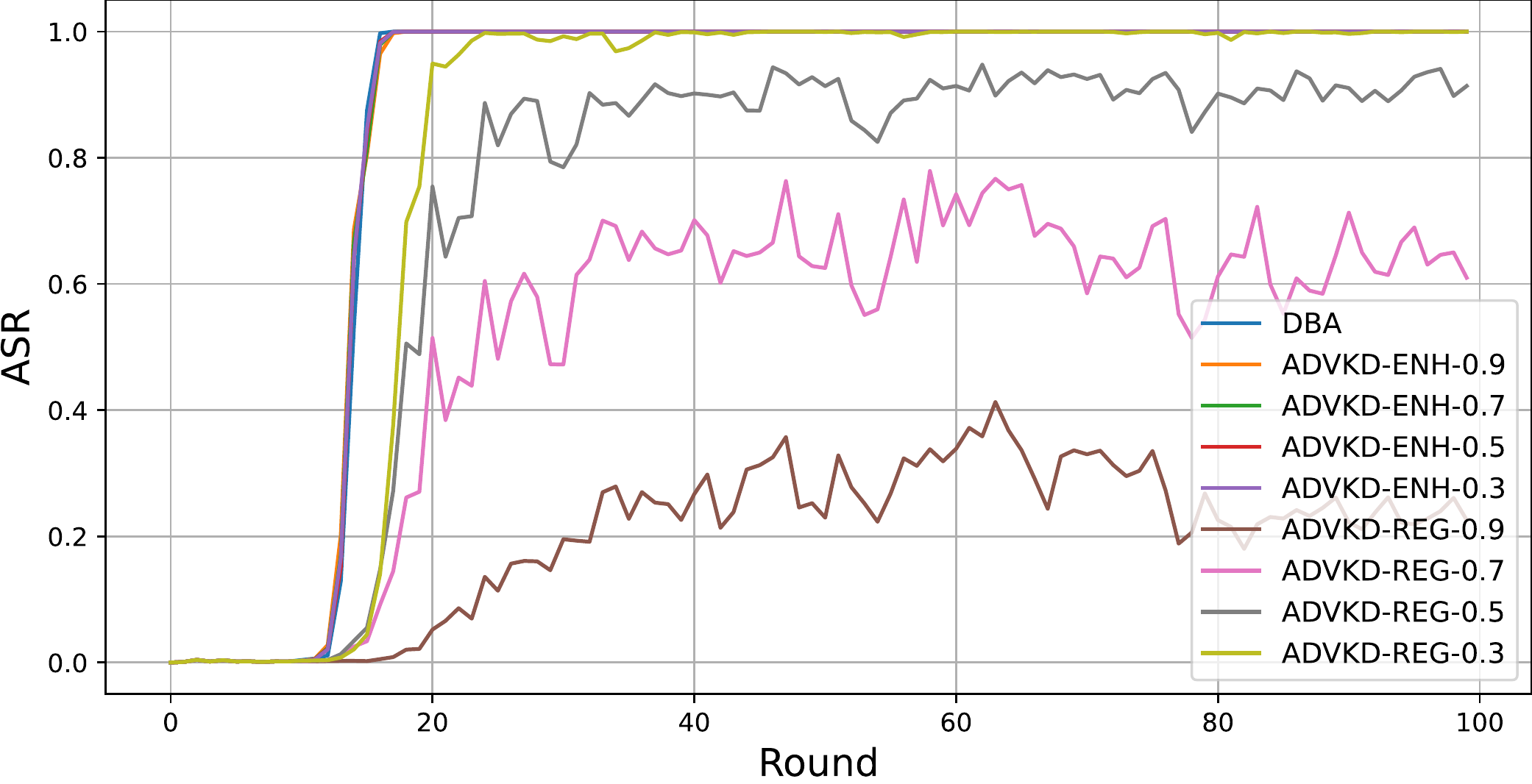}
    \label{fig:fig_Attack_NoDef_emnist_b}
  }
  \subfloat[Accuracy of DBA Methods]{
    \includegraphics[width=0.48\textwidth]{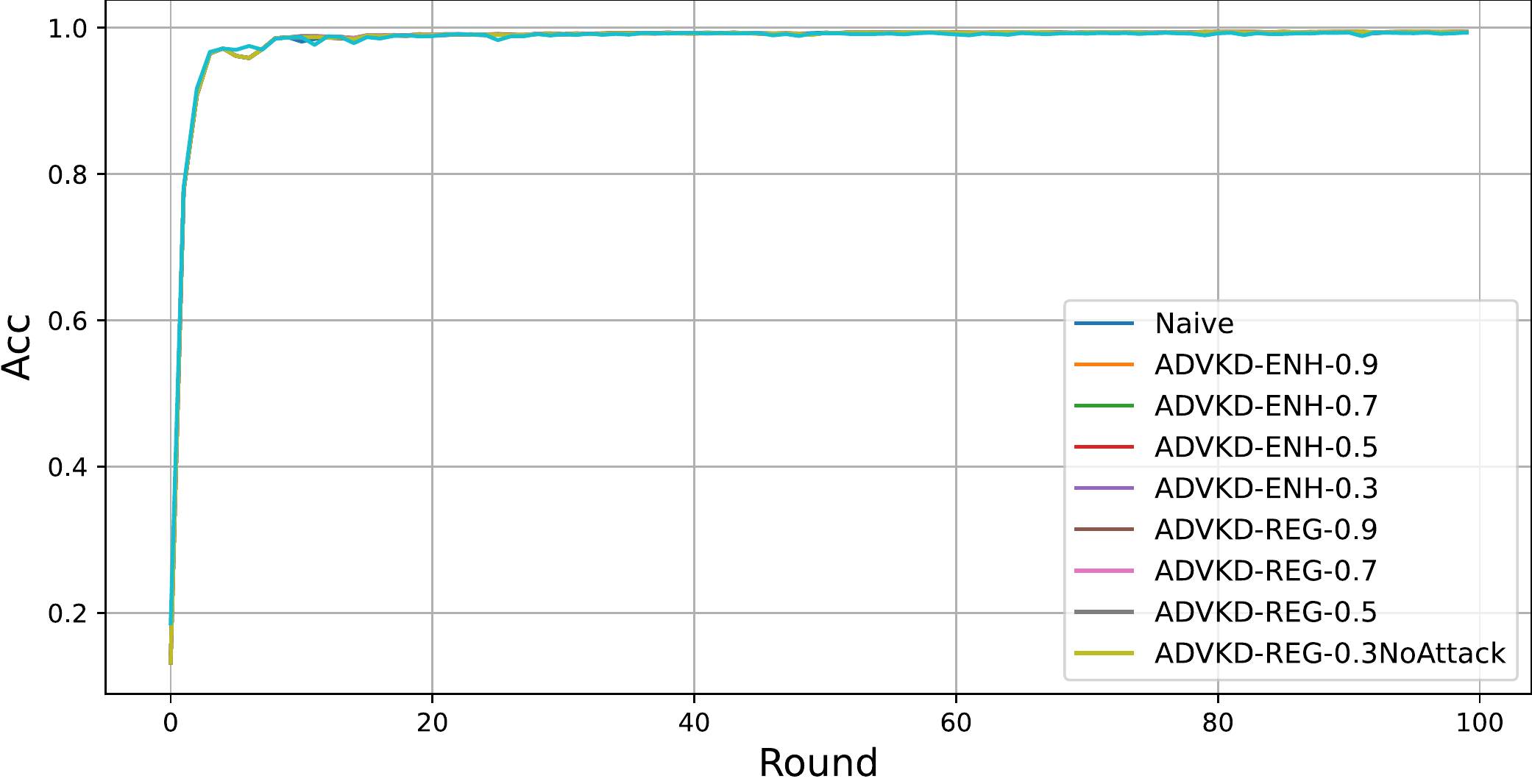}
    \label{fig:fig_Attack_NoDef_emnist_acc_b}
  }
  \caption{EMNIST in FedAvg}
  \label{fig:fig_Attack_NoDef_emnist}
\end{figure}

The result of Fashion-MNIST and EMNIST dataset are shown in Fig\ref{fig:fig_Attack_NoDef_fmnist} and Fig\ref{fig:fig_Attack_NoDef_emnist}. Fig\ref{fig:fig_Attack_NoDef_fmnist_a}, Fig\ref{fig:fig_Attack_NoDef_fmnist_acc_a}, Fig\ref{fig:fig_Attack_NoDef_emnist_a} and Fig\ref{fig:fig_Attack_NoDef_emnist_acc_a} are the ASR and the model accuracy of Naive method and Naive+ADVKD on these two datasets. Fig\ref{fig:fig_Attack_NoDef_fmnist_b}, \ref{fig:fig_Attack_NoDef_fmnist_acc_b}, Fig\ref{fig:fig_Attack_NoDef_emnist_b} and Fig\ref{fig:fig_Attack_NoDef_emnist_acc_b} are the results of DBA method and DBA+ADVKD. As we can see, the final ASR of Naive method and DBA are close to 100\% but the ASR of DBA grows faster than Naive method, which is consistent with result in \cite{DBA}. The results of ADVKD-ENH are close to original methods. But the ASR of ADVKD-REG are always lower than original methods, and it becomes worse when $\alpha$ gets larger. Such result confirms our intuition as the regularization in ADVKD-REG would damage the performance of backdoor. We can also find that these attack didn't make any obvious effect on the model accuracy.

\begin{figure}[htbp]
  \centering
  \subfloat[ASR of Naive Methods]{
    \includegraphics[width=0.48\textwidth]{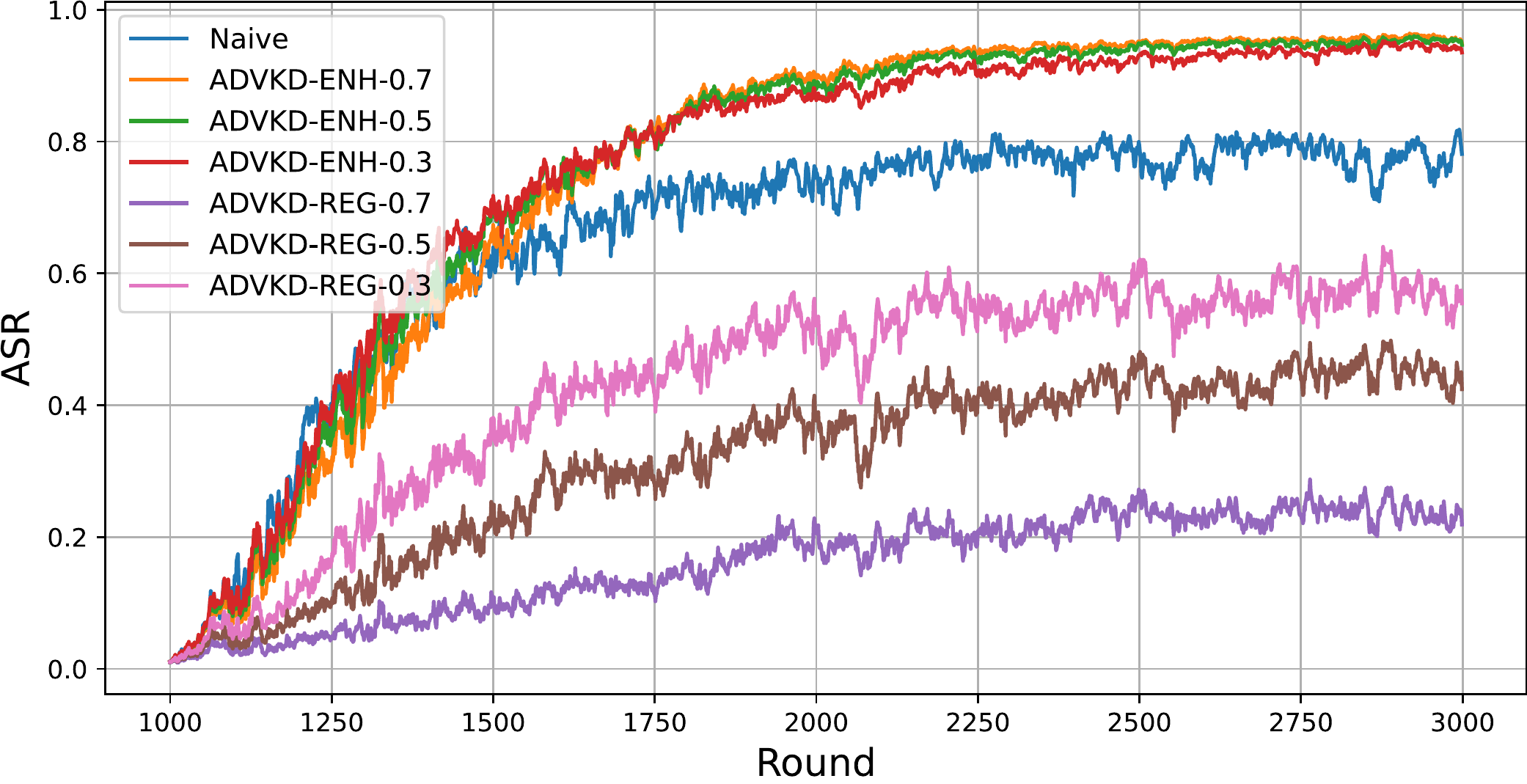}
    \label{fig:fig_Attack_NoDef_a}
  }
  \subfloat[Accuracy of Naive Methods]{
    \includegraphics[width=0.48\textwidth]{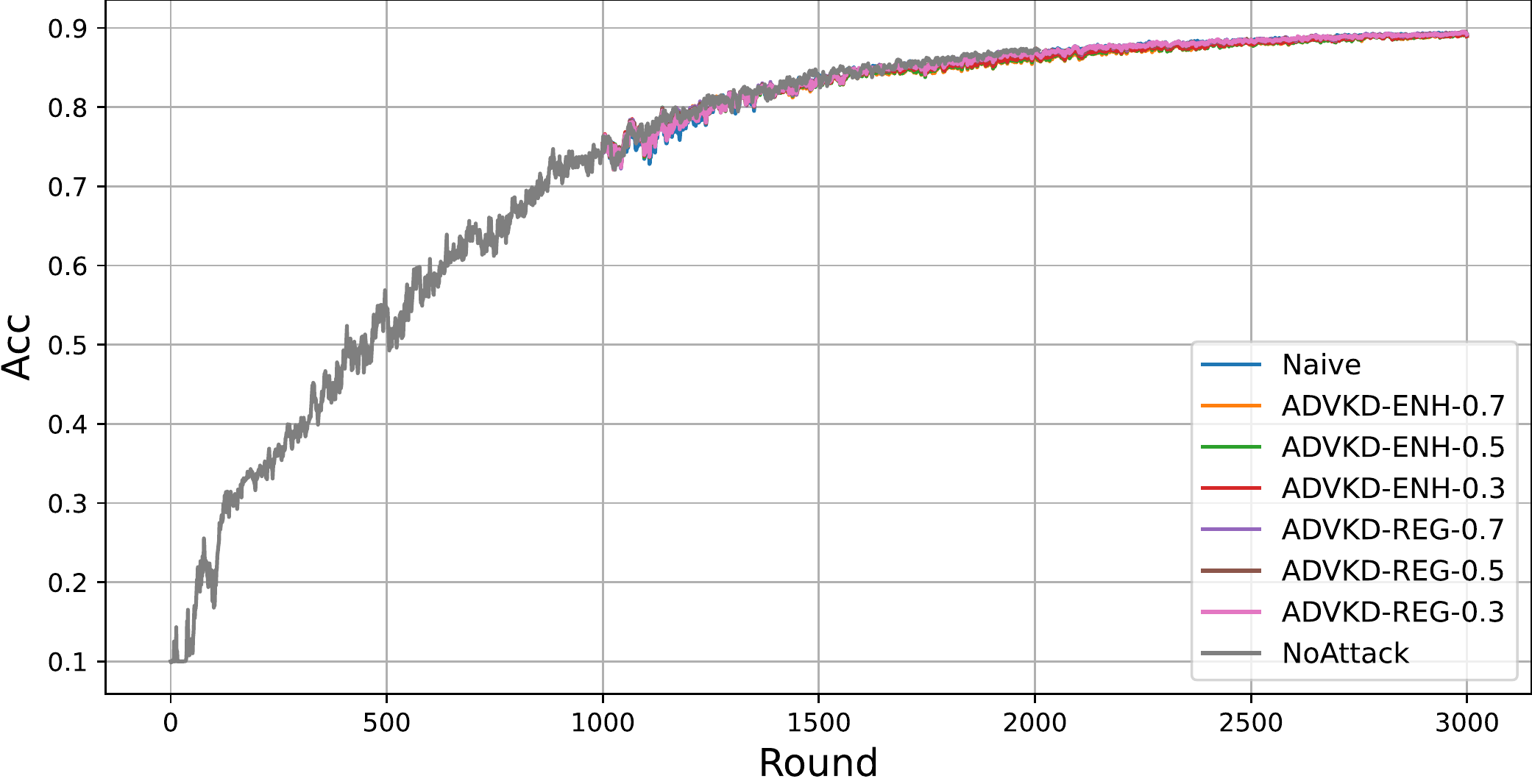}
    \label{fig:fig_Attack_NoDef_acc_a}
  }

  \subfloat[ASR of DBA Methods]{
    \includegraphics[width=0.48\textwidth]{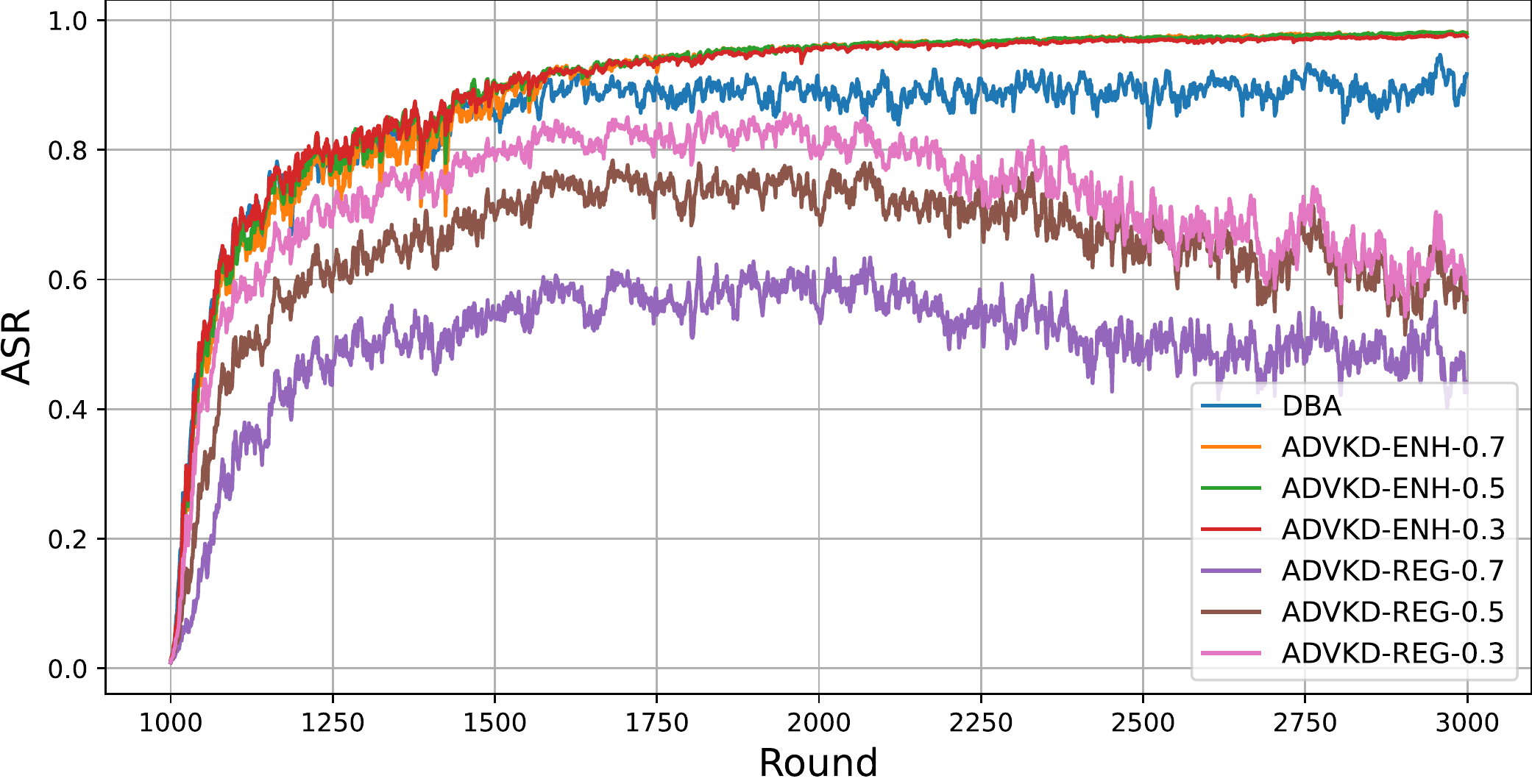}
    \label{fig:fig_Attack_NoDef_b}
  }
  \subfloat[Accuracy of DBA Methods]{
    \includegraphics[width=0.48\textwidth]{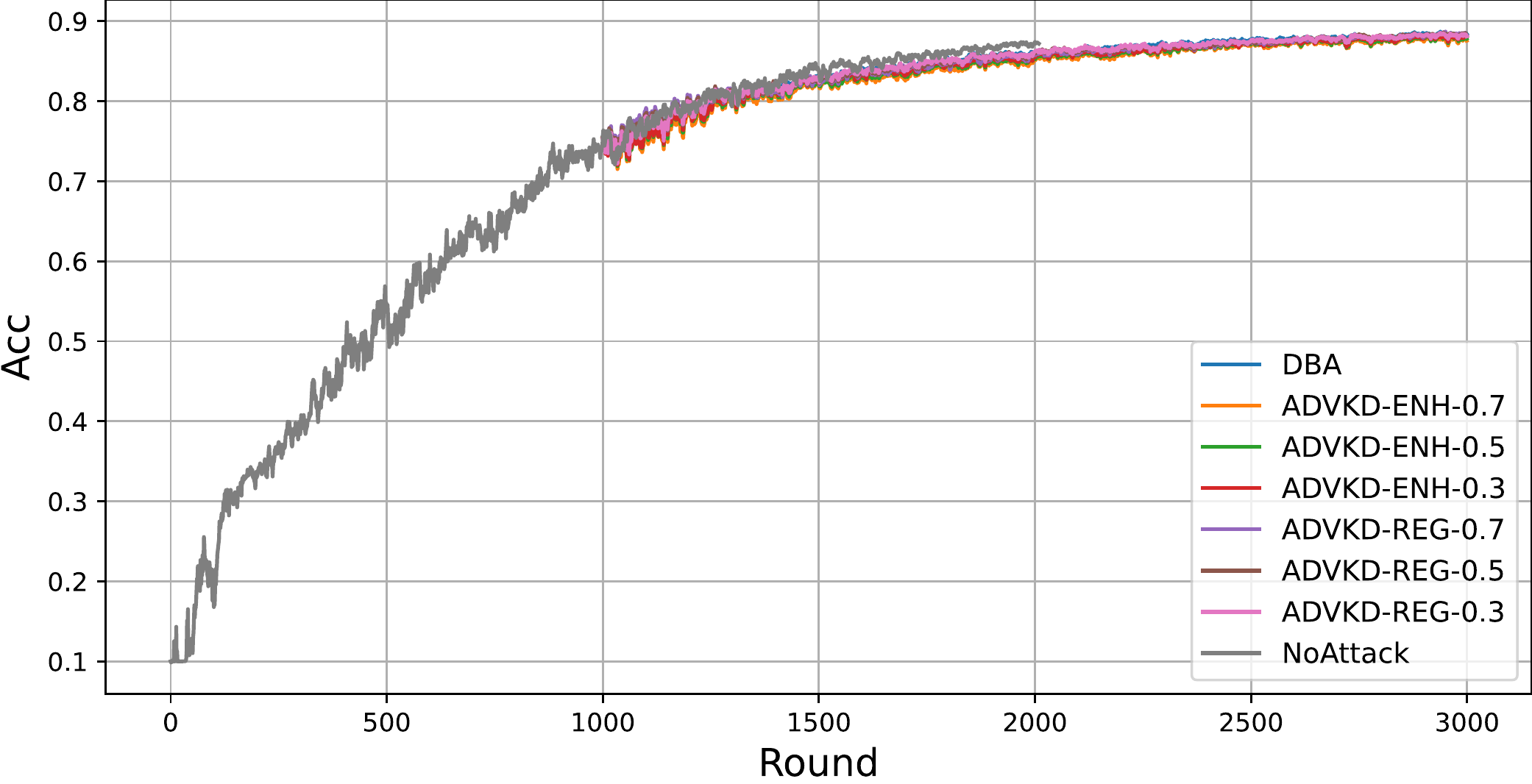}
    \label{fig:fig_Attack_NoDef_acc_b}
  }
  \caption{CIFAR10 in FedAvg}
  \label{fig:fig_Attack_NoDef}
\end{figure}

The results on CIFAR10 dataset are shown in Fig\ref{fig:fig_Attack_NoDef}. The model accuracy is still not affected by attacks. However, different from previous two datasets, the classification task of CIFAR10 dataset is more difficult, and it's harder to inject backdoor into model, so the final ASR of Naive method and DBA only reach about 80\% and 90\%. The ASR of ADVKD-ENH with two baseline methods can reach 95\% and 98\%, which is higher than original methods, confirming knowledge distillation can lead model moving toward a local minimum with better backdoor performance.

\paragraph{Multi-Krum}

In Multi-Krum, we need to satisfy the assumption that the number of participants $n$ in a round and the number of adversary $f$ satisfies $2f + 2 < n$. So we set $n=12, f=4$ and set the parameter $m$ of Multi-Krum to $n-f=8$.

\begin{figure}[htbp]
  \centering
  \subfloat[ASR of Naive Methods]{
    \includegraphics[width=0.48\textwidth]{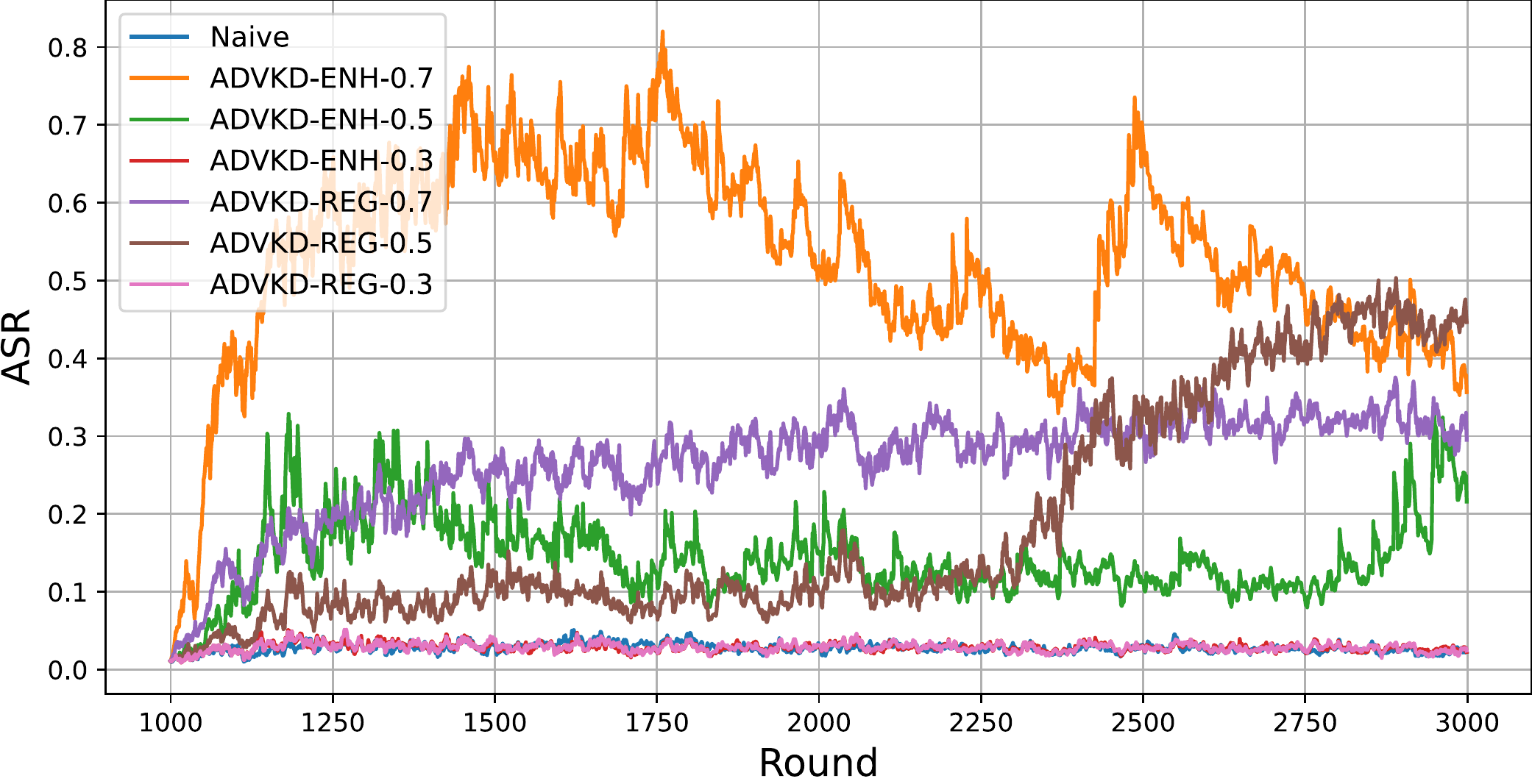}
    \label{fig:fig_Attack_Krum_a}
  }
  \subfloat[Count of Naive Methods]{
    \includegraphics[width=0.48\textwidth]{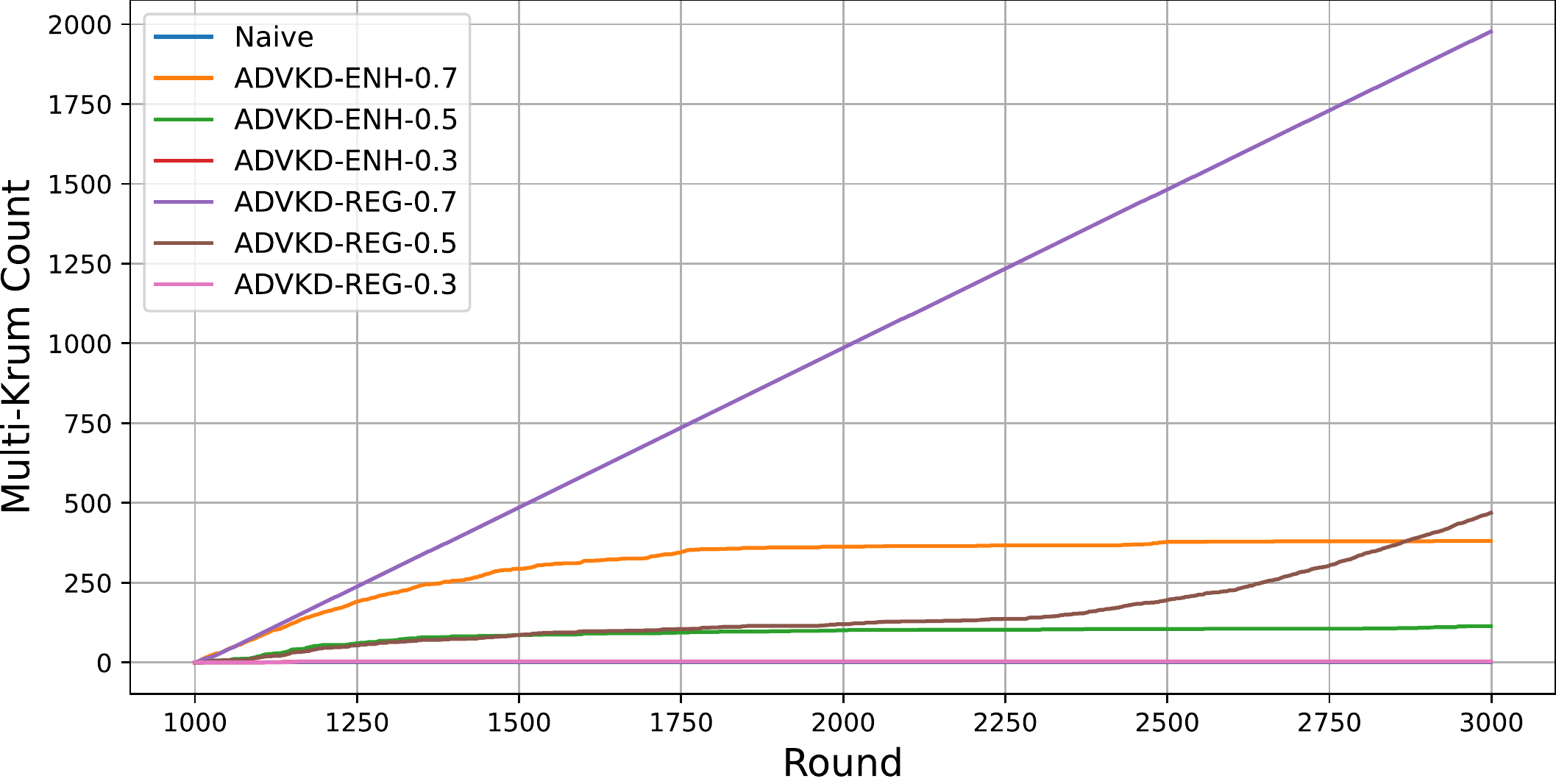}
    \label{fig:fig_Attack_Krum_b}
  }

  \subfloat[ASR of DBA Methods]{
    \includegraphics[width=0.48\textwidth]{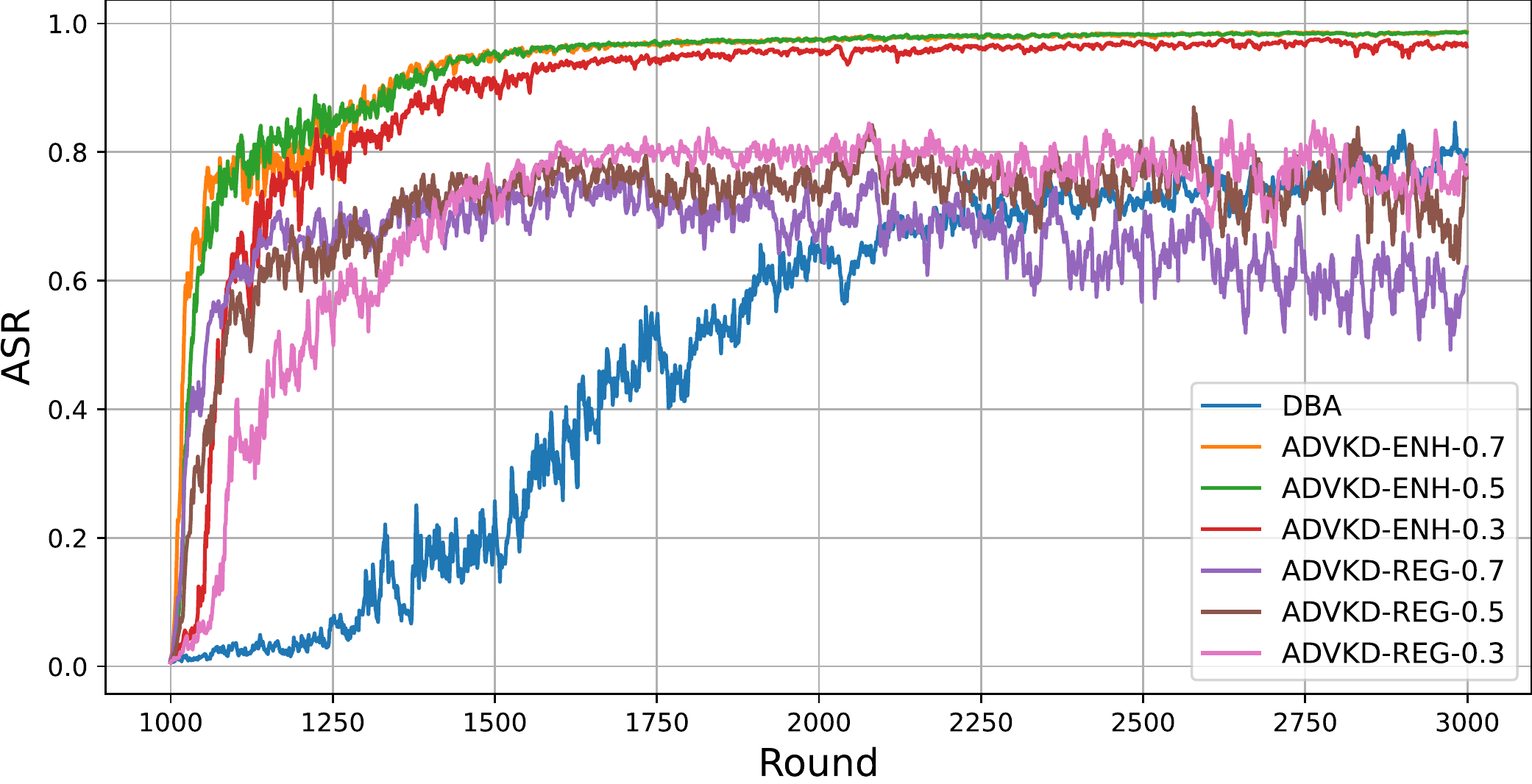}
    \label{fig:fig_Attack_Krum_c}
  }
  \subfloat[Count of DBA Methods]{
    \includegraphics[width=0.48\textwidth]{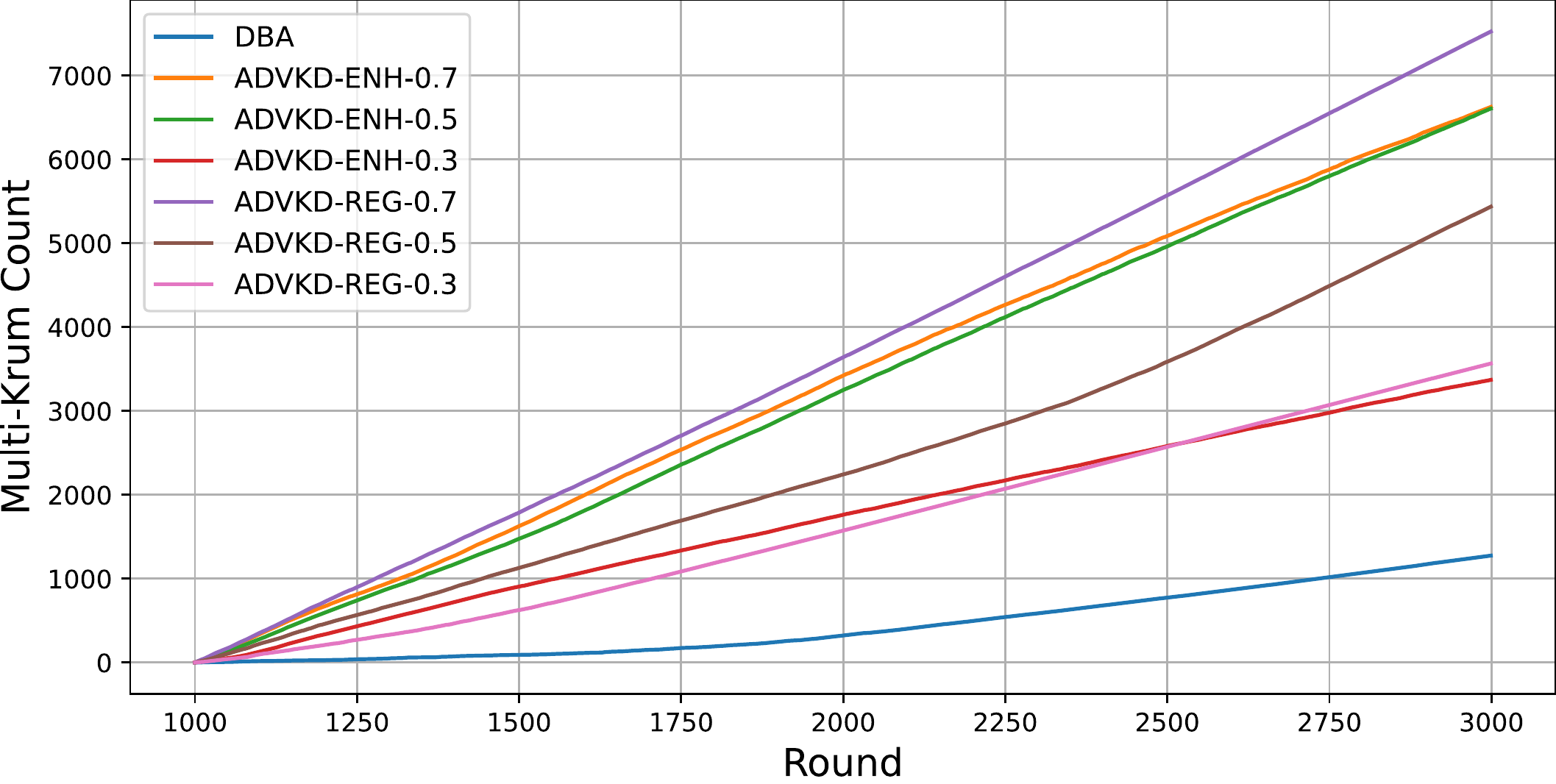}
    \label{fig:fig_Attack_Krum_d}
  }
  \caption{CIFAR10 in Multi-Krum}
  \label{fig:fig_Attack_Krum}
\end{figure}

The result of experiment on CIFAR10 dataset are shown in Fig\ref{fig:fig_Attack_Krum}. Fig\ref{fig:fig_Attack_Krum_a} and Fig\ref{fig:fig_Attack_Krum_b} show the ASR and accumulated number of adversary selected by Multi-Krum in every round of naive method and naive+ADVKD. We can see naive method have never been selected by Multi-Krum so that its ASR is close to 0, so do ADVKD when its parameter $\alpha$ is low(0.3) which means the weight of knowledge distillation is low and CrossEntropyLoss is still the majority. But when $\alpha$ gets larger, both ADVKD-ENH and ADVKD-REG are selected more frequently. When $\alpha$ reaches 0.7, even the selected count of ADVKD-REG is higher than ADVKD-ENH, but its ASR is still lower because ADVKD-REG damages the ASR, which is consistent with previous result in FedAvg.

Fig\ref{fig:fig_Attack_Krum_c} and Fig\ref{fig:fig_Attack_Krum_d} show the ASR and count of DBA and DBA+ADVKD. We can see that the count of DBA is low, so its ASR is also lower than itself in FedAvg(Fig\ref{fig:fig_Attack_NoDef_b}). For ADVKD, the count of ADVKD gets larger when $\alpha$ gets larger. But the limit of ADVKD-REG still damage its ASR so the ASR of ADVKD-REG grows fast in the beginning but finally only reaches 0.6-0.8, which is not better than DBA. However, ADVKD-ENH doesn't have such limit, so the ASR of ADVKD-ENH not only grows fast but also can reach almost 1.0 in the end.

\begin{figure}[htbp]
  \centering
  \subfloat[ASR of Naive Methods]{
    \includegraphics[width=0.48\textwidth]{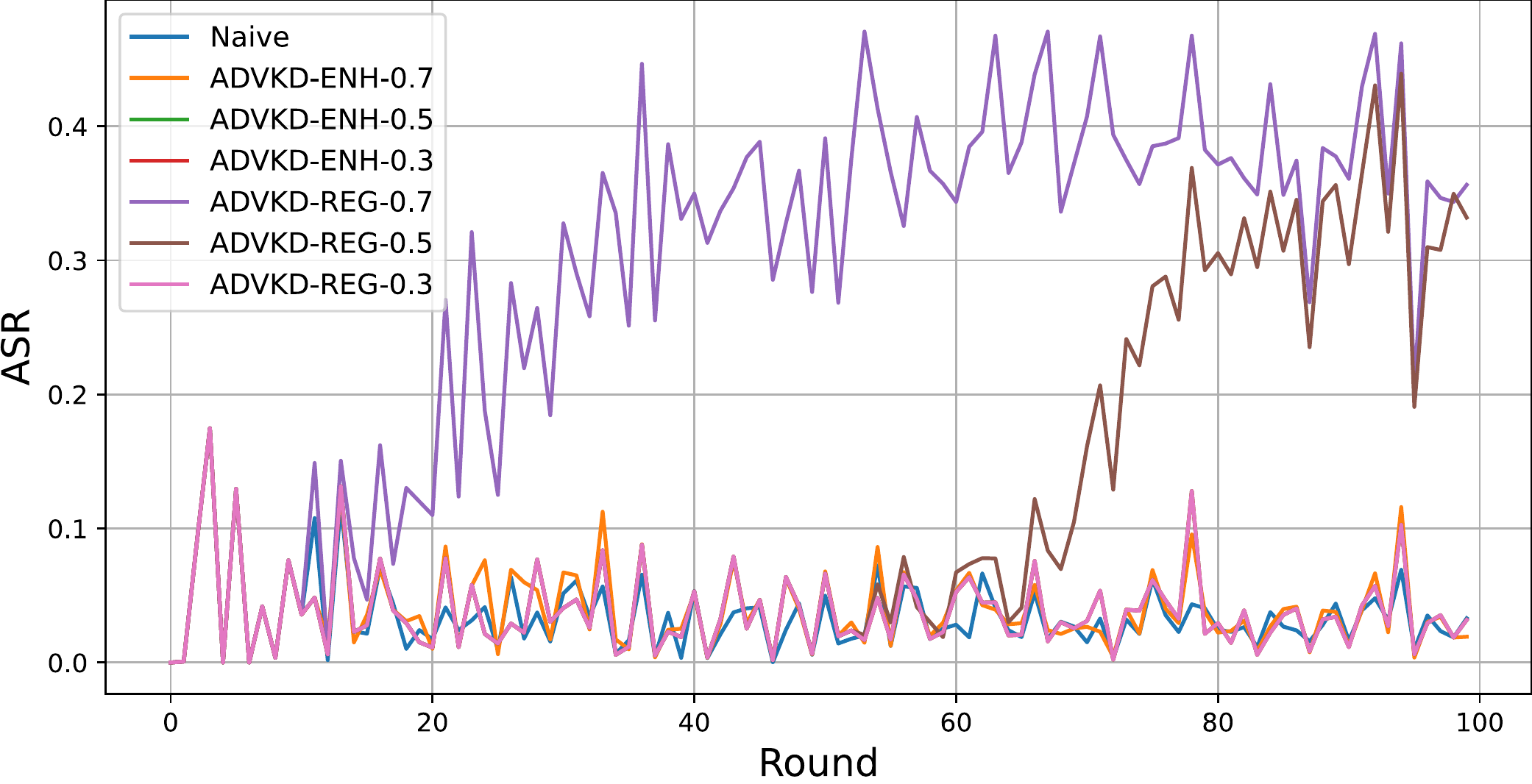}
    \label{fig:fig_Attack_Krum_fmnist_a}
  }
  \subfloat[Count of Naive Methods]{
    \includegraphics[width=0.48\textwidth]{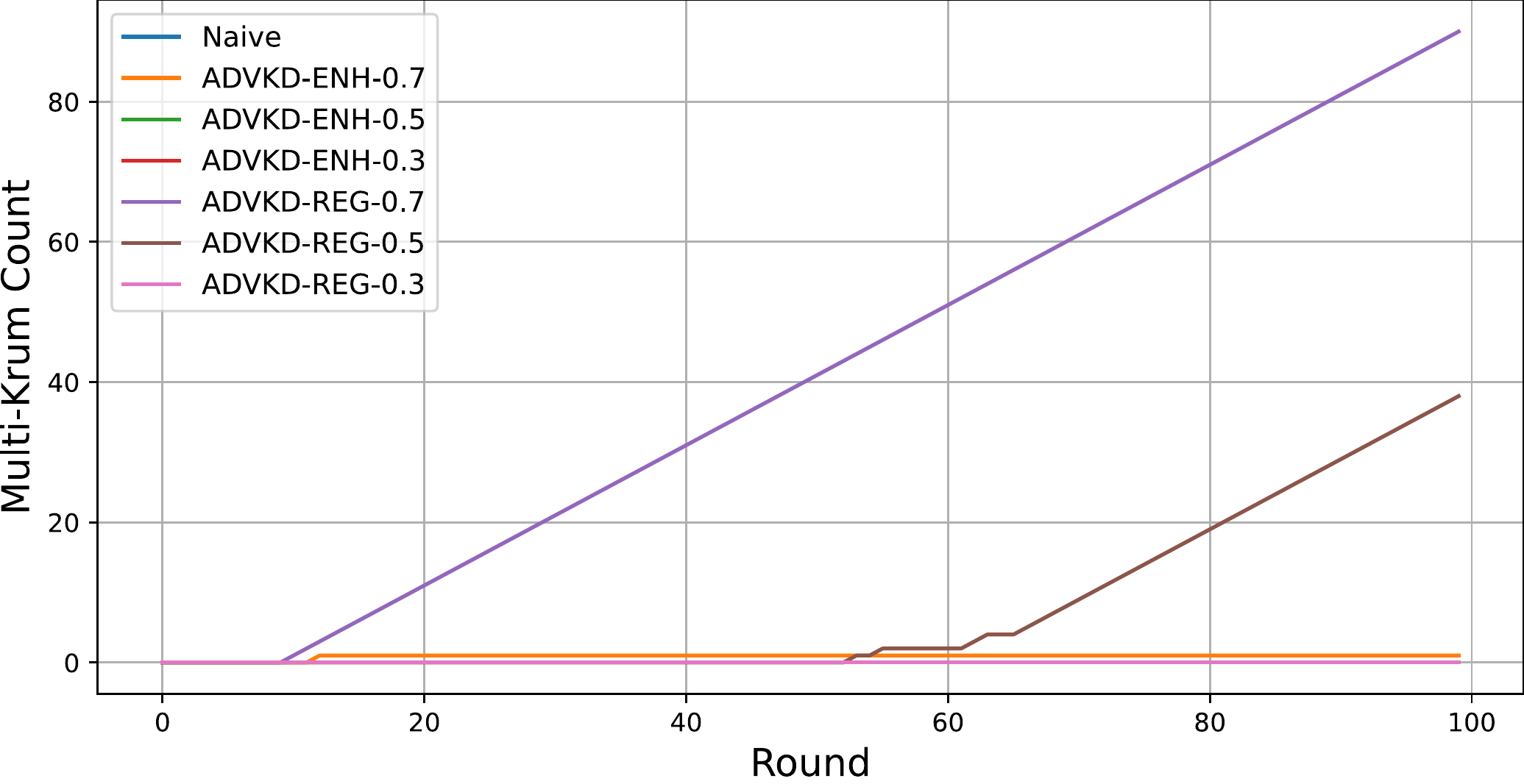}
    \label{fig:fig_Attack_Krum_fmnist_b}
  }

  \subfloat[ASR of DBA Methods]{
    \includegraphics[width=0.48\textwidth]{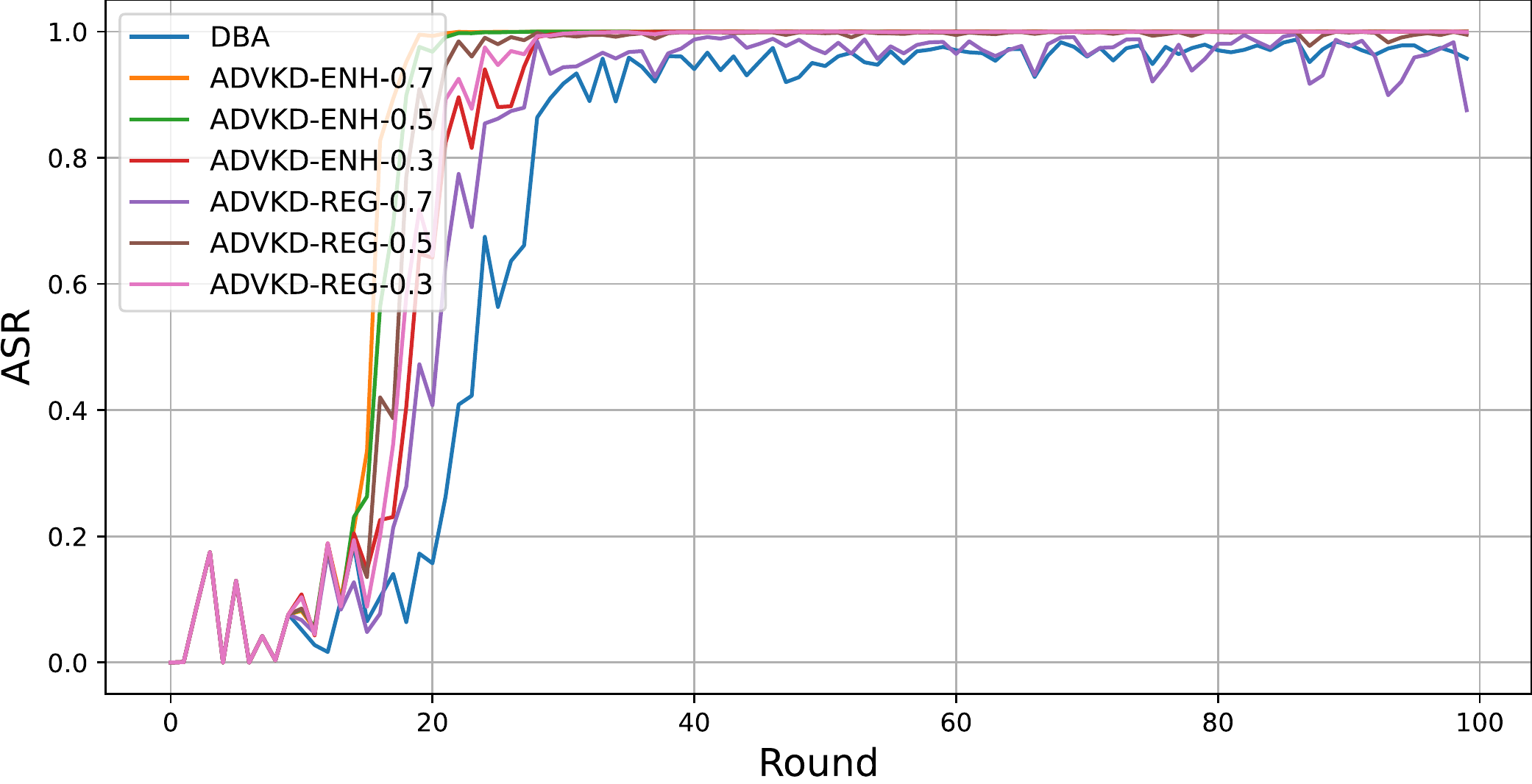}
    \label{fig:fig_Attack_Krum_fmnist_c}
  }
  \subfloat[Count of DBA Methods]{
    \includegraphics[width=0.48\textwidth]{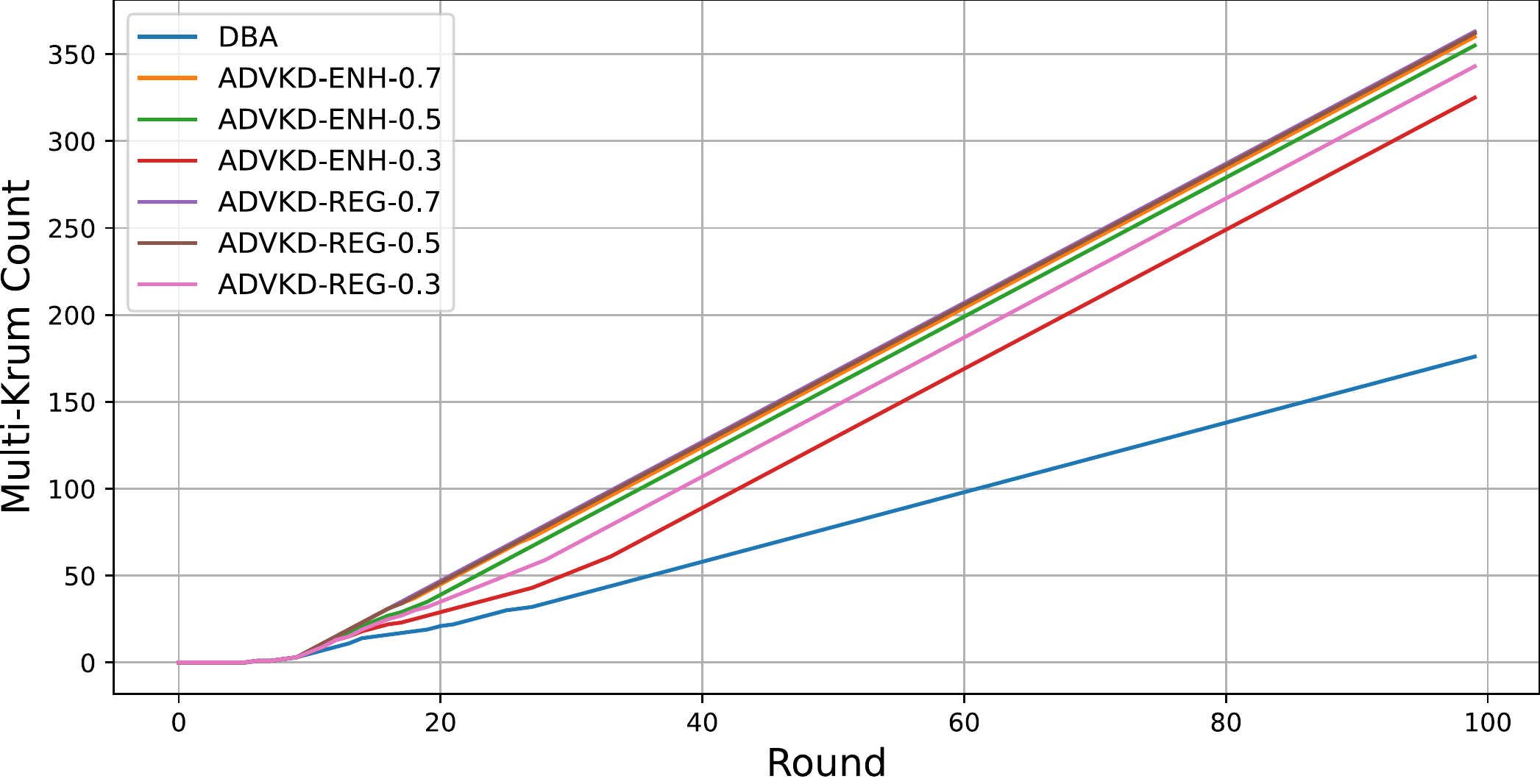}
    \label{fig:fig_Attack_Krum_fmnist_d}
  }
  \caption{Fashion-MNIST in Multi-Krum}
  \label{fig:fig_Attack_Krum_fmnist}
\end{figure}

Fig\ref{fig:fig_Attack_Krum_fmnist} shows the result on Fashion-MNIST. As we can see, the model updates of naive method and the model updates of naive+ADVKD-ENH can hardly been selected, so their ASR is close to 0. For naive+ADVKD-REG, only when the $\alpha$ is high enough(0.5 or 0.7), it can pass the Multi-Krum defense. In DBA and DBA+ADVKD, as the model updates become more stealthy and the number of adversary gets more, more adversarial model update can pass the Multi-Krum Defense, and so all of these methods can get a good ASR in the end.

\begin{figure}[htbp]
  \centering
  \subfloat[ASR of Naive Methods]{
    \includegraphics[width=0.48\textwidth]{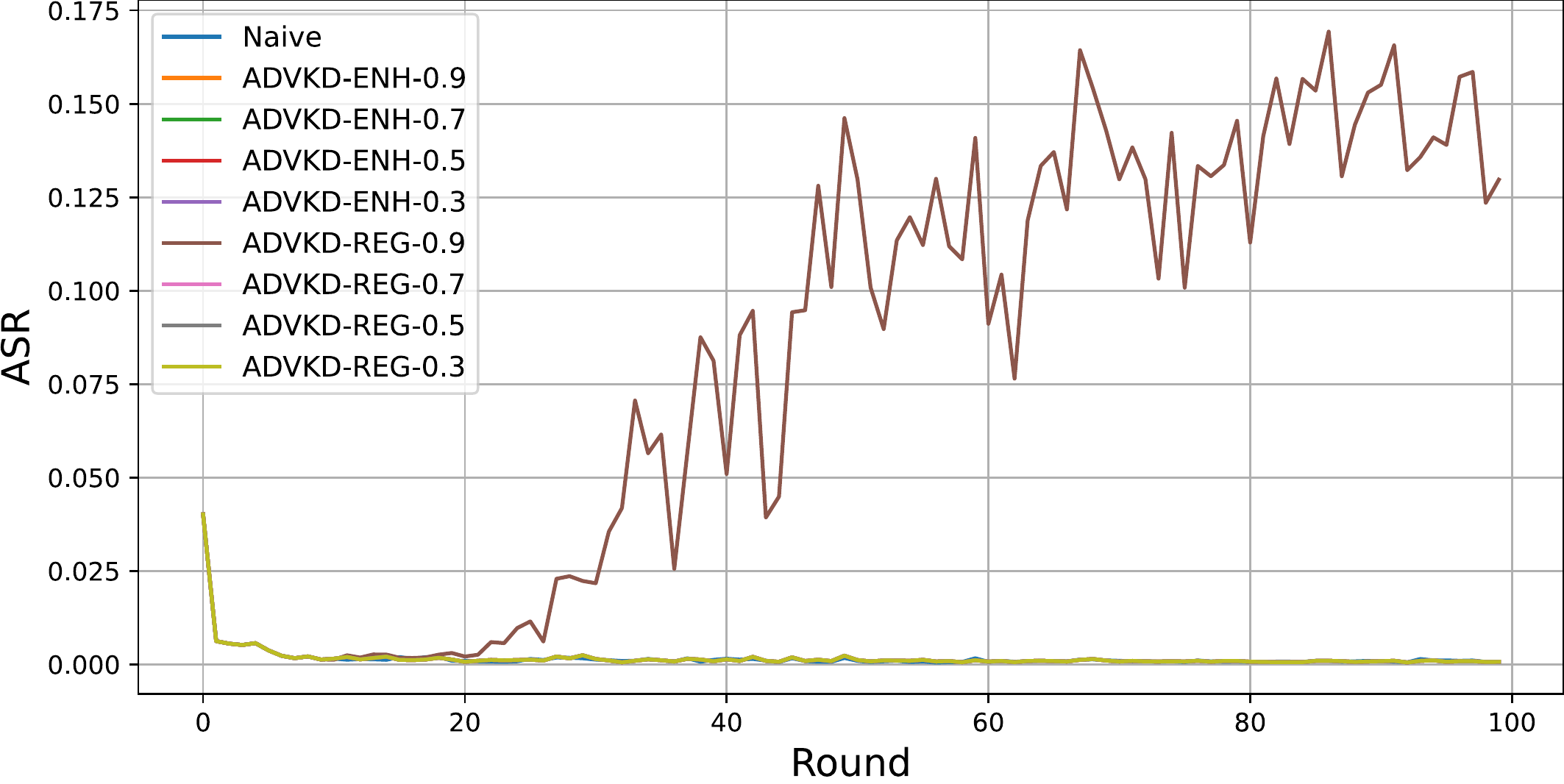}
    \label{fig:fig_Attack_Krum_emnist_a}
  }
  \subfloat[ASR of DBA Methods]{
    \includegraphics[width=0.48\textwidth]{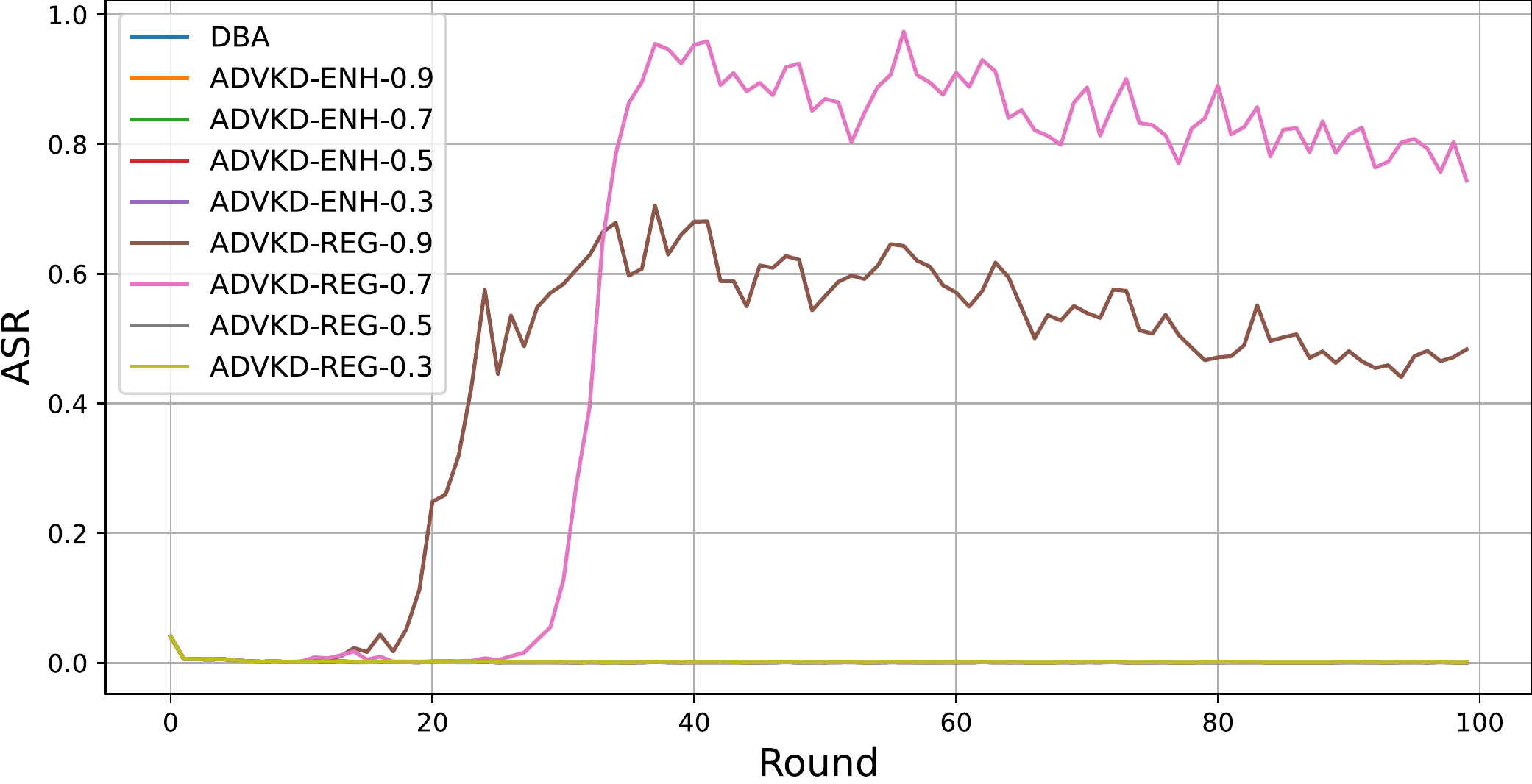}
    \label{fig:fig_Attack_Krum_emnist_c}
  }
  \caption{EMNIST in Multi-Krum}
  \label{fig:fig_Attack_Krum_emnist}
\end{figure}

From the experiment result of EMNIST dataset shown in Fig\ref{fig:fig_Attack_Krum_emnist}, we can find the problem in Fashion-MNIST(only ADVKD-REG can bypass defense) gets worse in EMNIST dataset. When the $\alpha$ can only be 0.3,0.5 or 0.7, only DBA+ADVKD-REG with $\alpha=0.7$ can bypass Multi-Krum and get a high ASR, and the model updates of other methods are discarded by Multi-Krum defense so that their ASR are always close to 0. One reason of this phenomenon may be that with the dataset and model getting more simple, the distances between adversarial updates and benign updates become relatively larger, so Multi-Krum can find these adversarial model update and prune them. On the other hand, as ADVKD-REG adds additional regularization on backdoor model training, with larger $\alpha$, the adversarial model updates would get nearer to other benign model updates and hence they can pass the detection of Multi-Krum. For this phenomenon, we additionally test the effect of ADVKD-REG with $\alpha=0.9$. The results are given in Fig\ref{fig:fig_Attack_NoDef_emnist} and Fig\ref{fig:fig_Attack_Krum_emnist}, the results show that assigning too much weight to regularization damages the performance of backdoor, so the ASR is also low. Finally, DBA+ADVKD-REG with $\alpha = 0.7$ gets the best ASR under this scenario.

\paragraph{FLAME}

\begin{figure}[htbp]
  \centering
  \subfloat[ASR of Naive Methods]{
    \includegraphics[width=0.48\textwidth]{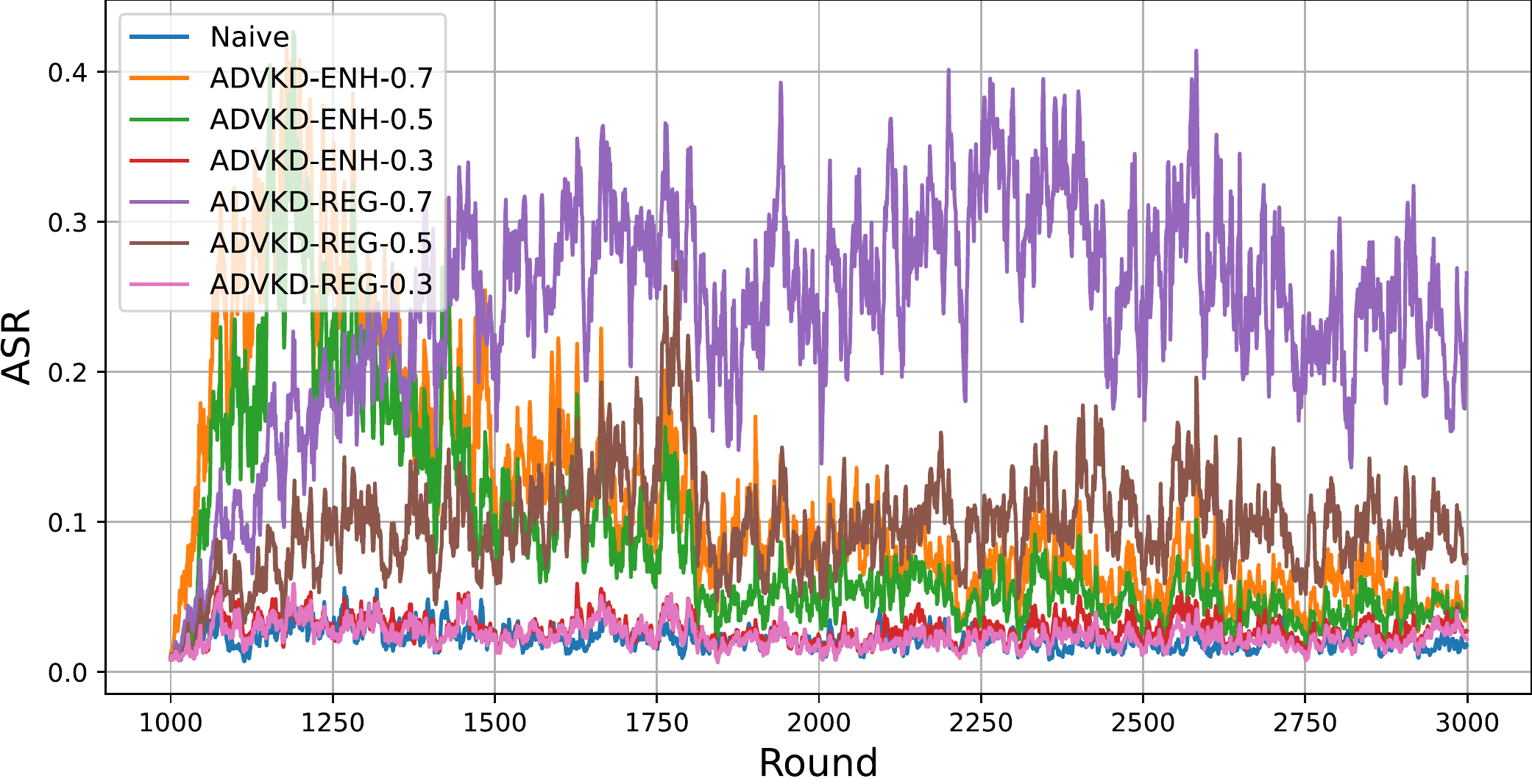}
    \label{fig:fig_Attack_FLAME_cifar10_a}
  }
  \subfloat[ASR of DBA Methods]{
    \includegraphics[width=0.48\textwidth]{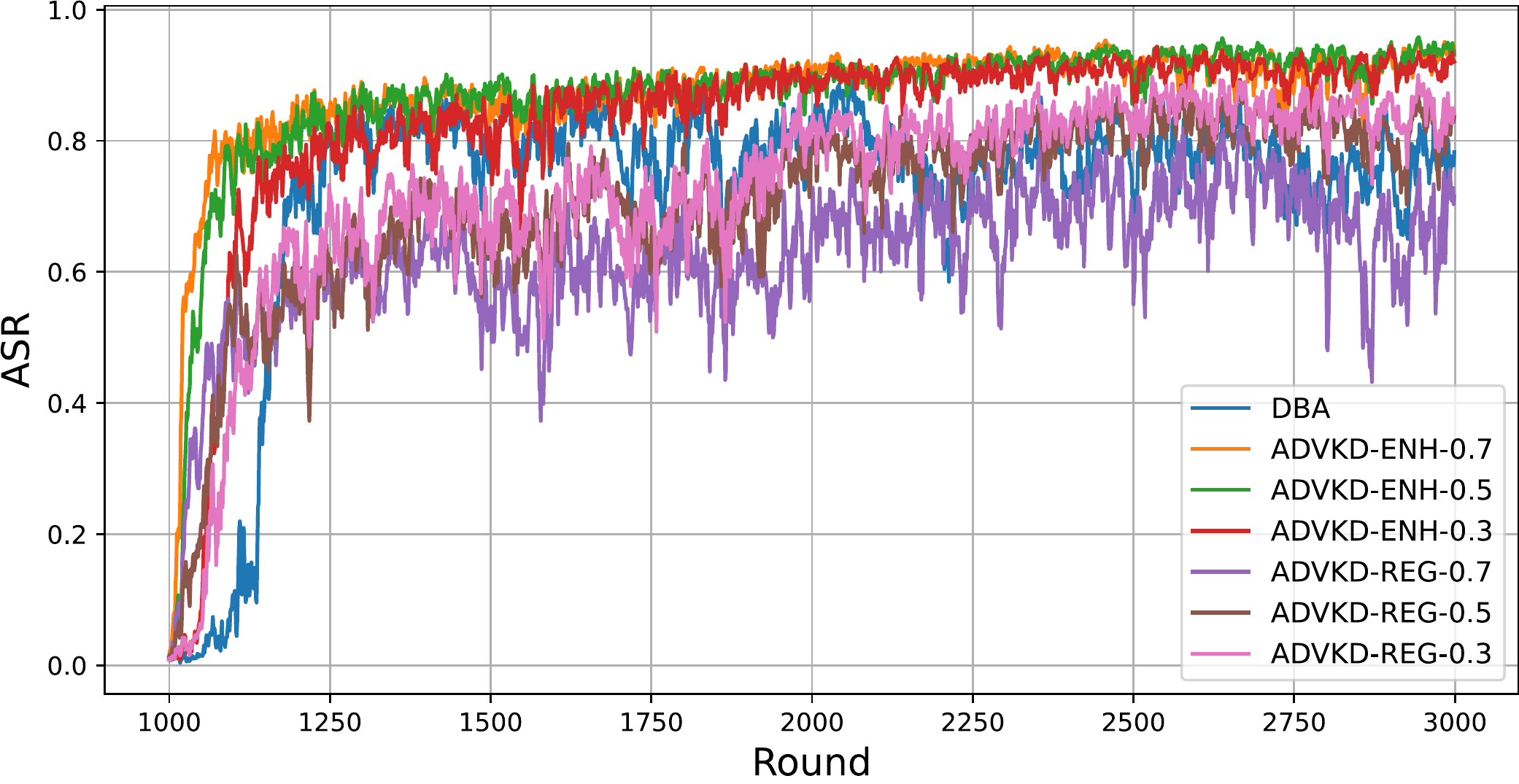}
    \label{fig:fig_Attack_FLAME_cifar10_b}
  }

  \subfloat[Smoothed ASR of Naive Methods]{
    \includegraphics[width=0.48\textwidth]{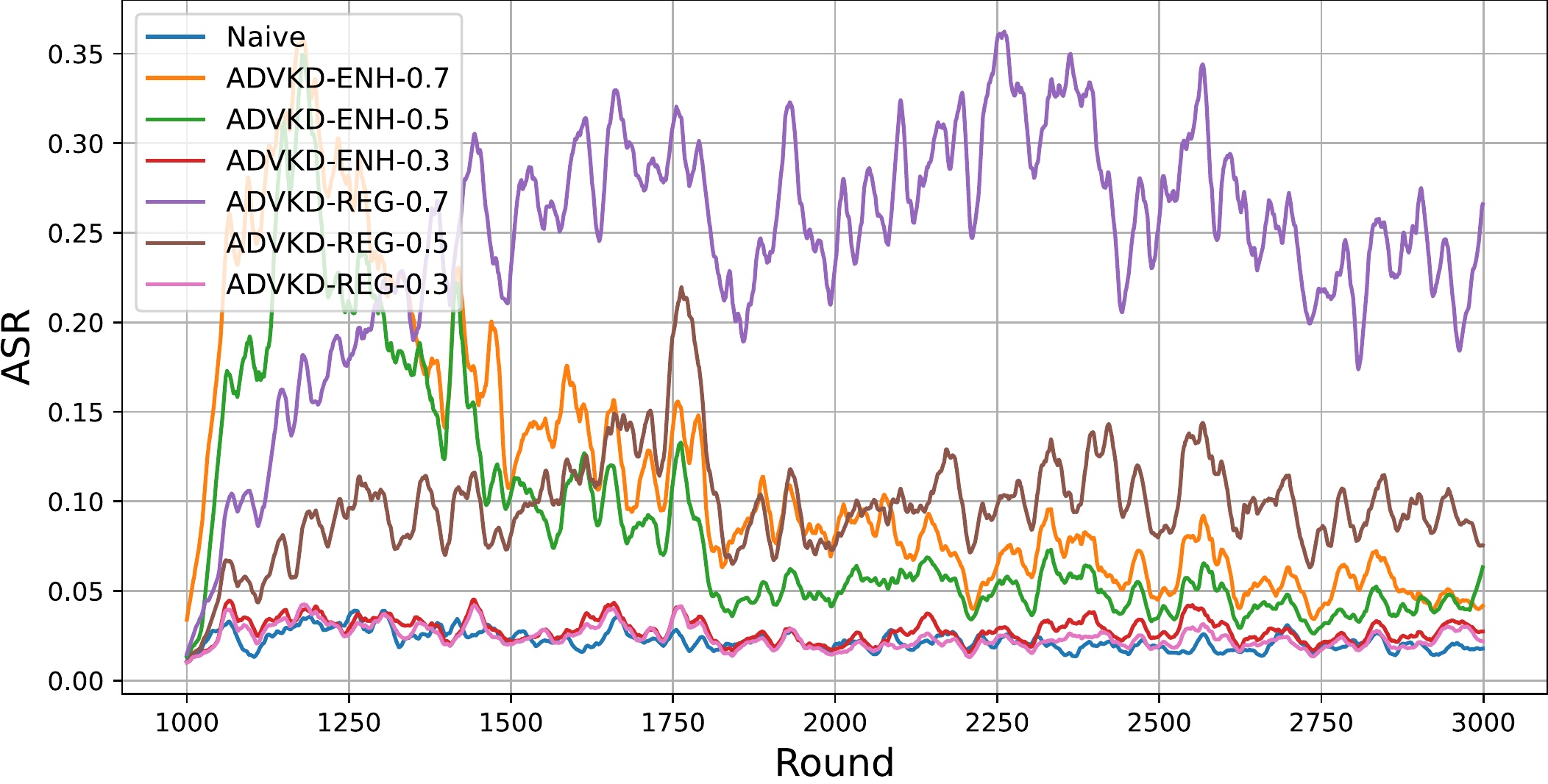}
    \label{fig:fig_Attack_FLAME_cifar10_c}
  }
  \subfloat[Smoothed ASR of DBA Methods]{
    \includegraphics[width=0.48\textwidth]{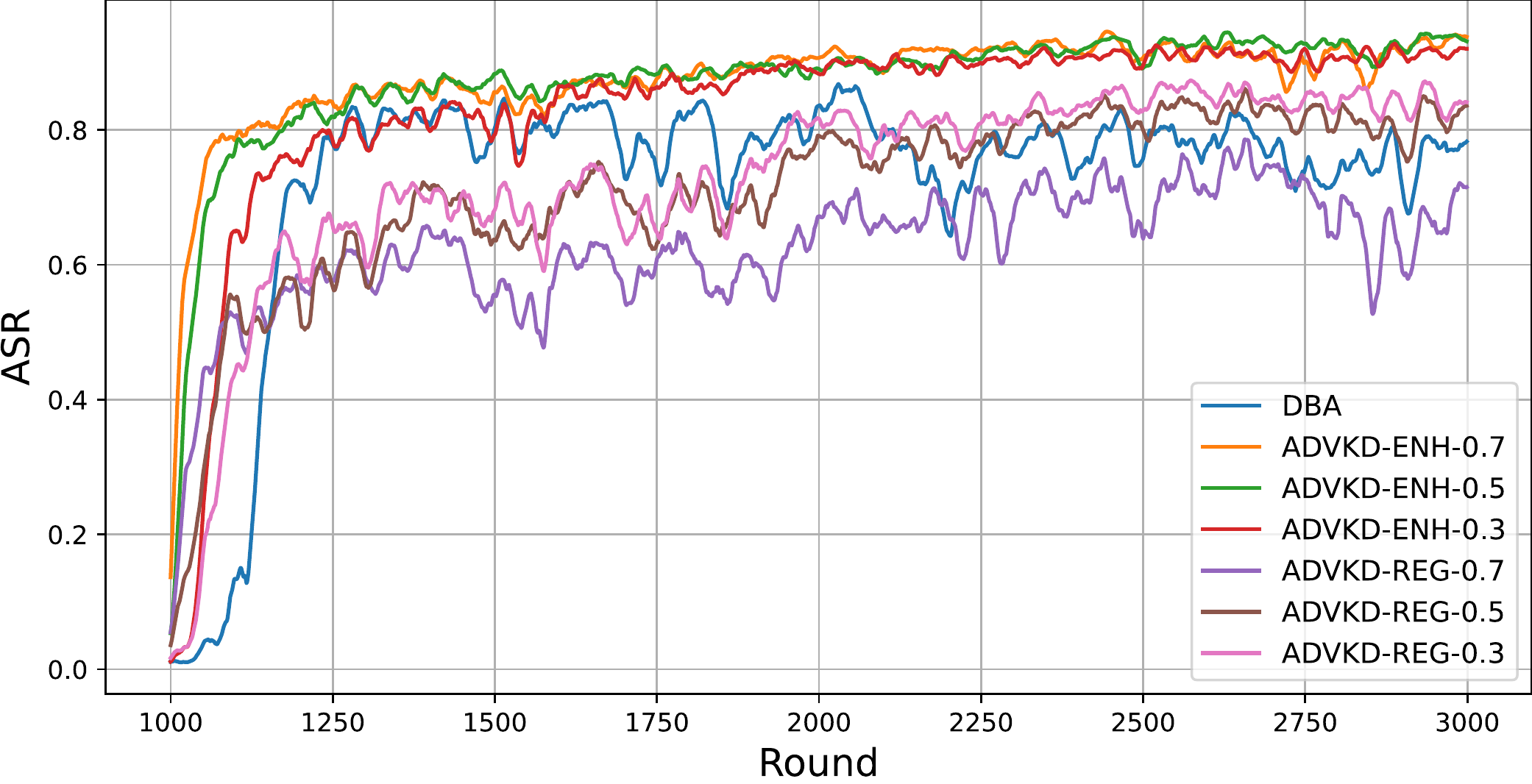}
    \label{fig:fig_Attack_FLAME_cifar10_d}
  }
  \caption{CIFAR10 in FLAME}
  \label{fig:fig_Attack_FLAME_cifar10}
\end{figure}

The experiment results of CIFAR10 dataset are shown in Fig\ref{fig:fig_Attack_FLAME_cifar10_a} and Fig\ref{fig:fig_Attack_FLAME_cifar10_b}. As FLAME adds Gaussian noise in every round, the performance of backdoor in complex model such as ResNet-18 becomes unstable. So we apply rolling average on the ASR, in other word, convolving ASR data with a uniform kernel. The smoothed ASR are shown in Fig\ref{fig:fig_Attack_FLAME_cifar10_c} and Fig\ref{fig:fig_Attack_FLAME_cifar10_d}. The performance of naive method and naive+ADVKD in this scenario is not so good, and the highest ASR is only 30\%. But in the experiment of DBA and DBA+ADVKD, we can see that it is quite similar to the result of Multi-Krum in Fig\ref{fig:fig_Attack_Krum_c}. The final ASR of DBA is about 80\%, DBA+ADVKD-ENH is about 92\% and DBA+ADVKD-REG is about 70\% to 85\% as it is still affected by its regularization.

\begin{figure}[htbp]
  \centering
  \subfloat[ASR of Naive Methods]{
    \includegraphics[width=0.48\textwidth]{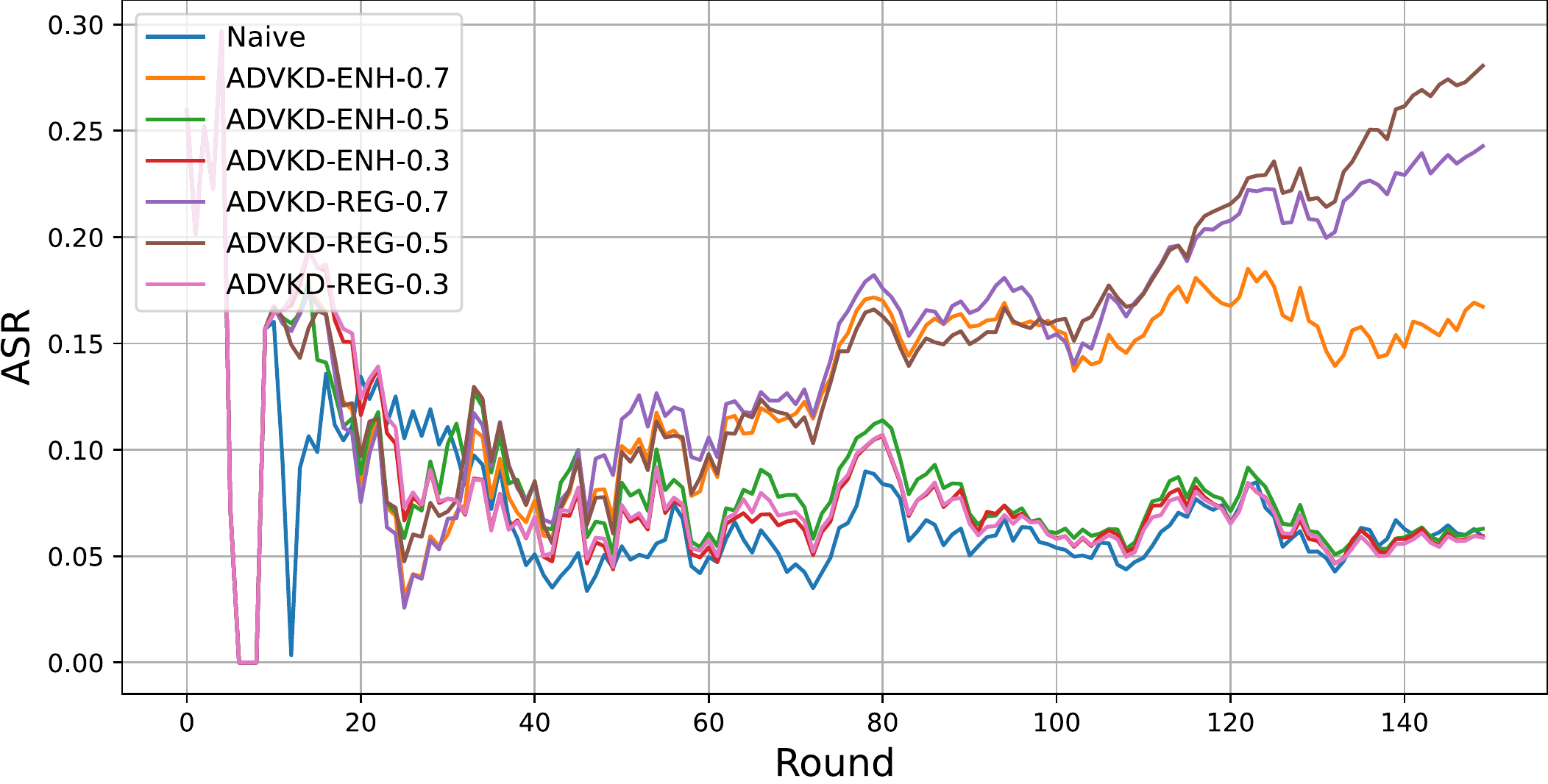}
    \label{fig:fig_Attack_FLAME_fmnist_a}
  }
  \subfloat[ASR of DBA Methods]{
    \includegraphics[width=0.48\textwidth]{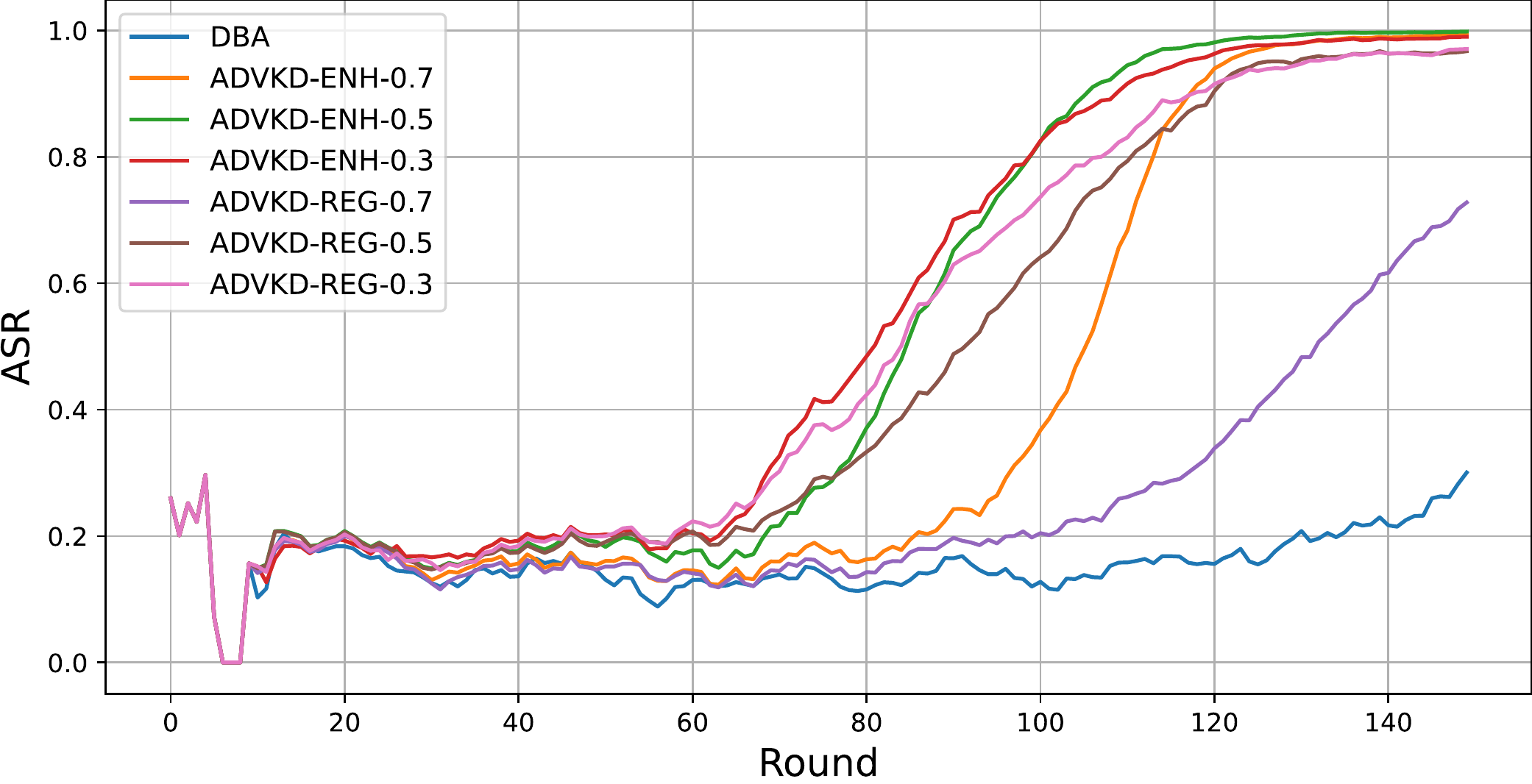}
    \label{fig:fig_Attack_FLAME_fmnist_b}
  }
  \caption{Fashion-MNIST in FLAME}
  \label{fig:fig_Attack_FLAME_fmnist}
\end{figure}

\begin{figure}[htbp]
  \centering
  \subfloat[ASR of Naive Methods]{
    \includegraphics[width=0.48\textwidth]{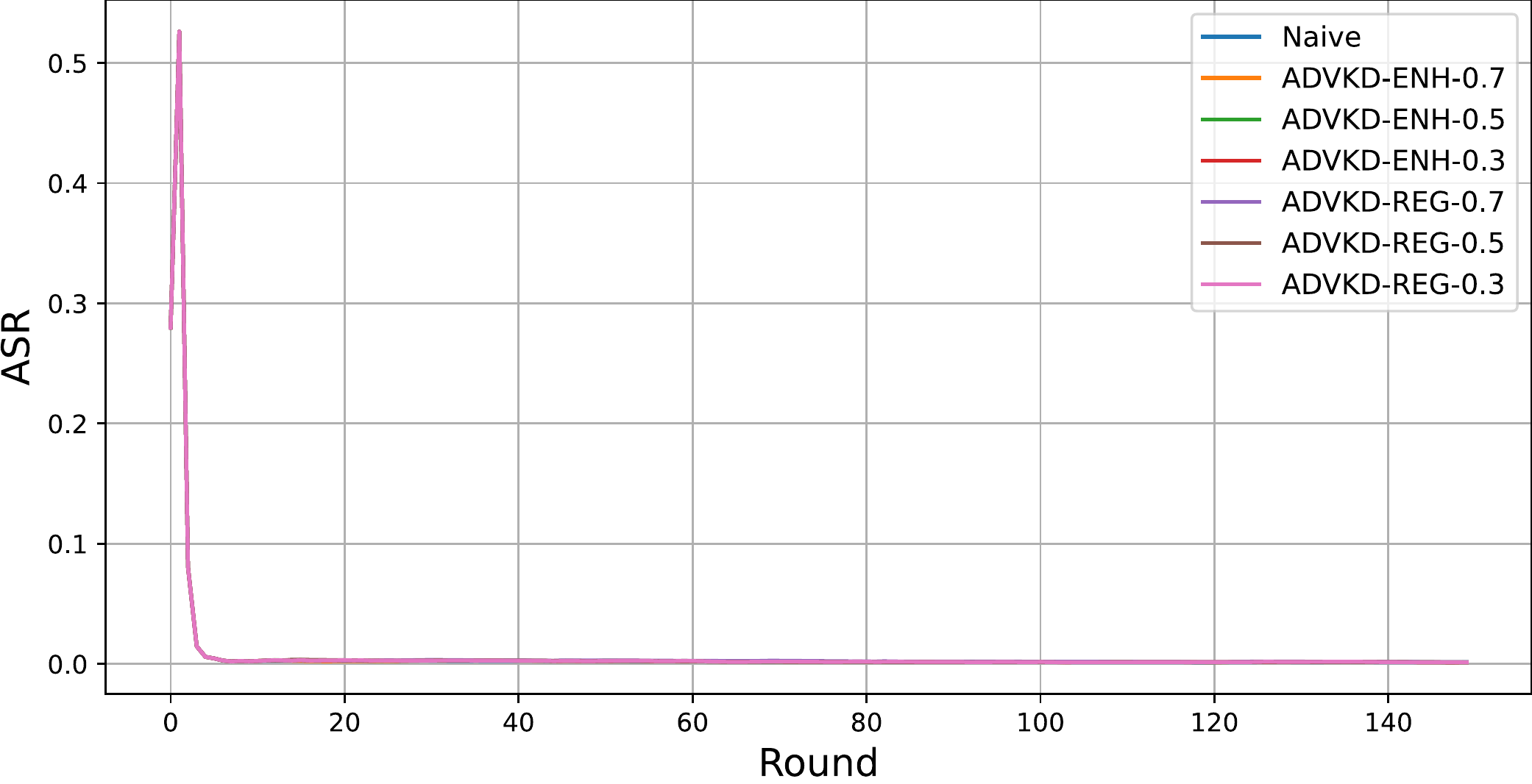}
    \label{fig:fig_Attack_FLAME_emnist_a}
  }
  \subfloat[ASR of DBA Methods]{
    \includegraphics[width=0.48\textwidth]{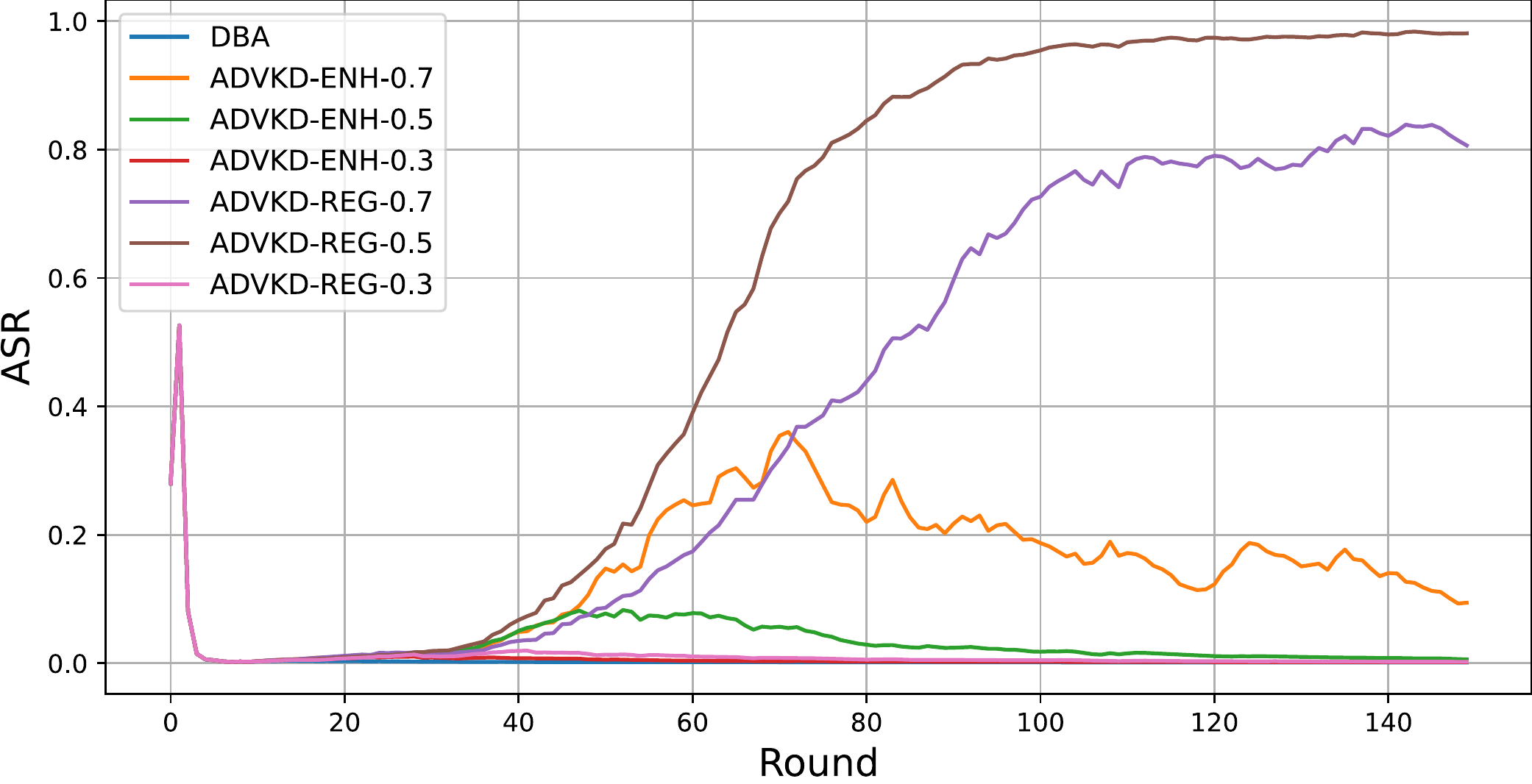}
    \label{fig:fig_Attack_FLAME_emnist_b}
  }
  \caption{EMNIST in FLAME}
  \label{fig:fig_Attack_FLAME_emnist}
\end{figure}

As shown in Fig\ref{fig:fig_Attack_FLAME_fmnist} and Fig\ref{fig:fig_Attack_FLAME_emnist}, we can find that the experiment results of FLAME is similar to the results of Multi-Krum. In experiments on Fashion-MNIST dataset, the ASR of naive method and naive+ADVKD is low, even the performance of ADVKD gets better when $\alpha$ gets larger, the highest ASR is only 28\%. However, for DBA and DBA+ADVKD, even the ASR of DBA is still low, the ASR of DBA+ADVKD are close to 100\% and only ADVKD-REG with large $\alpha$ holds a lower ASR due to its regularization. In experiments on EMNIST, both naive method and naive+ADVKD are failed to pass the detection of FLAME so that the ASR is close to 0. The DBA method and DBA+ADVKD-ENH are also affected by FLAME and their attack finally failed. But, when $\alpha$ is high(0.5 or 0.7), DBA+ADVKD-REG can break the defense of FLAME, and reach a high ASR (80\% and 98\%) in final.

\paragraph{Effect of Gamma and Beta}

\begin{figure}[htbp]
  \centering
  \subfloat[ASR of ADVKD-ENH with lower parameters]{
    \includegraphics[width=0.48\textwidth]{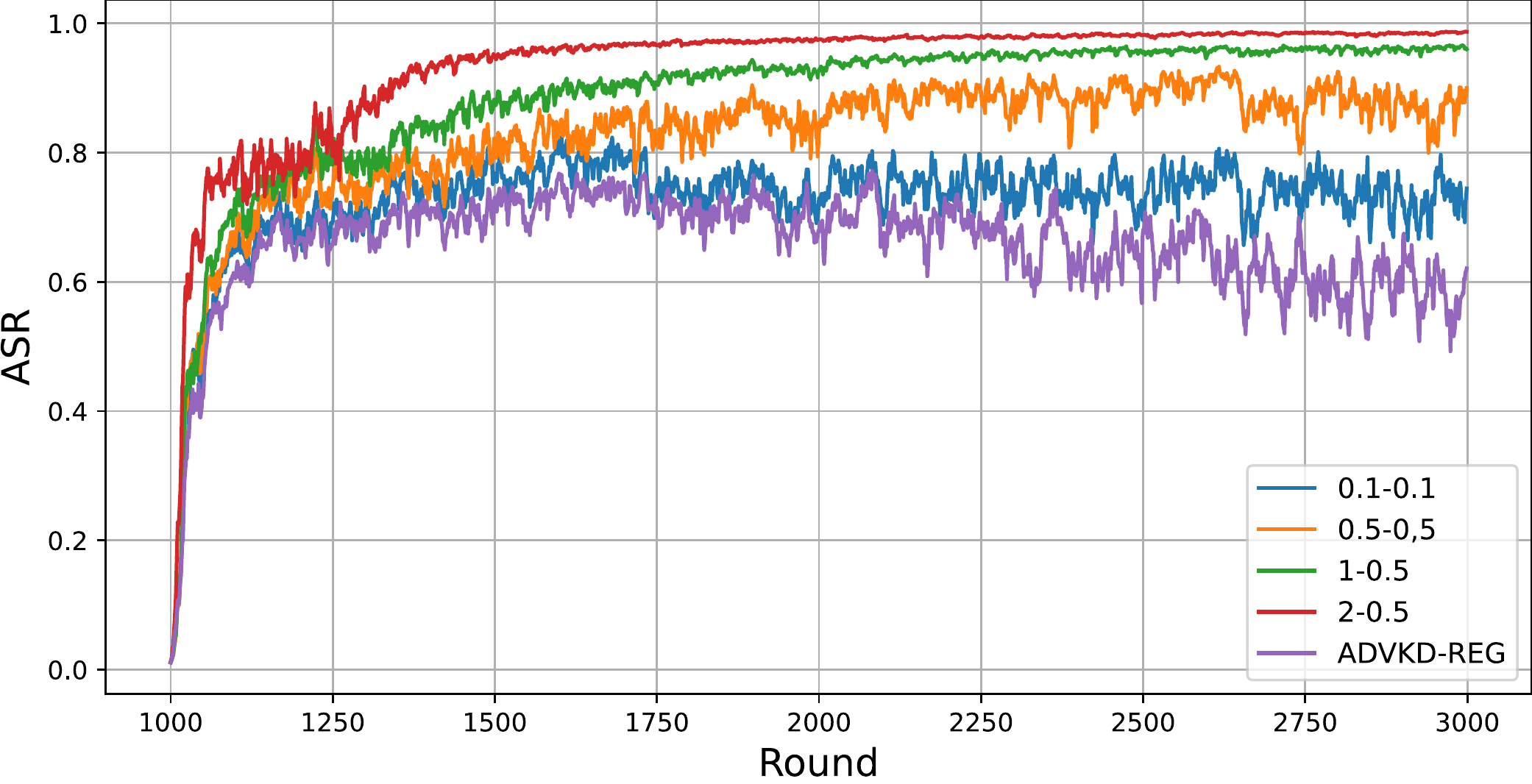}
    \label{fig:fig_gb_less}
  }
  \subfloat[Smoothed ASR of ADVKD-ENH with higher parameters]{
    \includegraphics[width=0.48\textwidth]{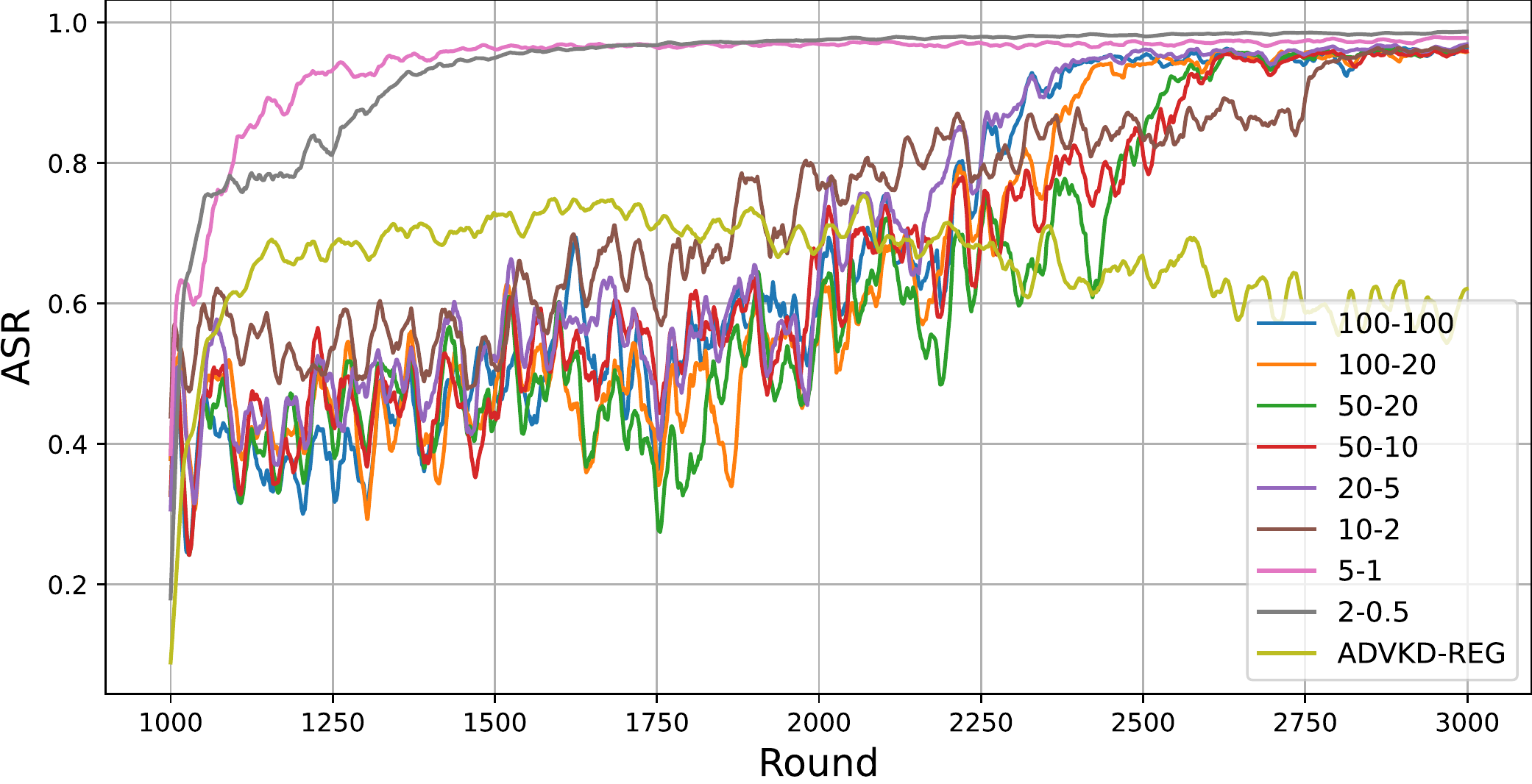}
    \label{fig:fig_gb_greater}
  }

  \subfloat[Count of ADVKD-ENH with higher parameters]{
    \includegraphics[width=0.48\textwidth]{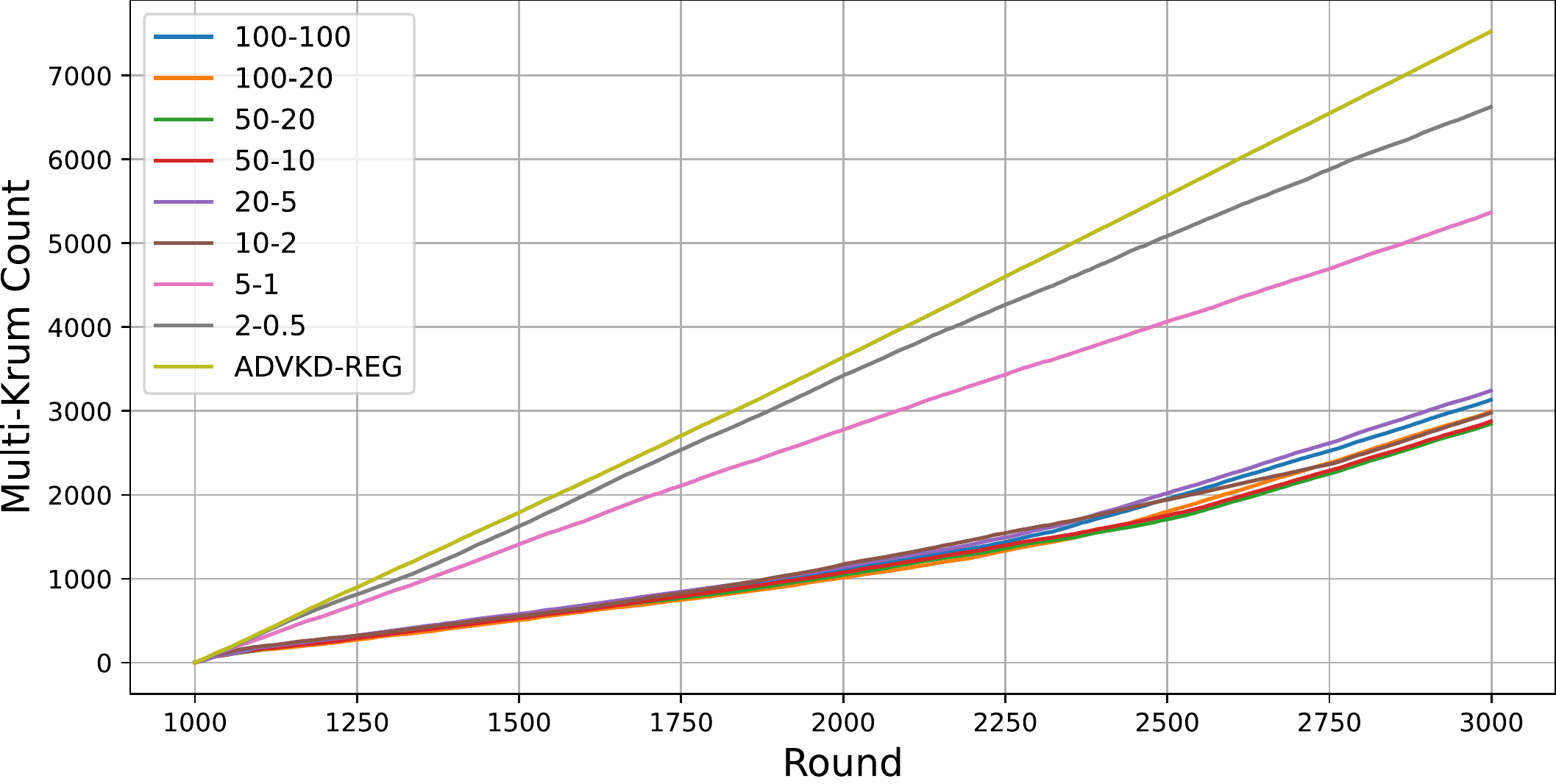}
    \label{fig:fig_gb_greater_count}
  }
  \subfloat[Accuracy of ADVKD-ENH with higher parameters]{
    \includegraphics[width=0.48\textwidth]{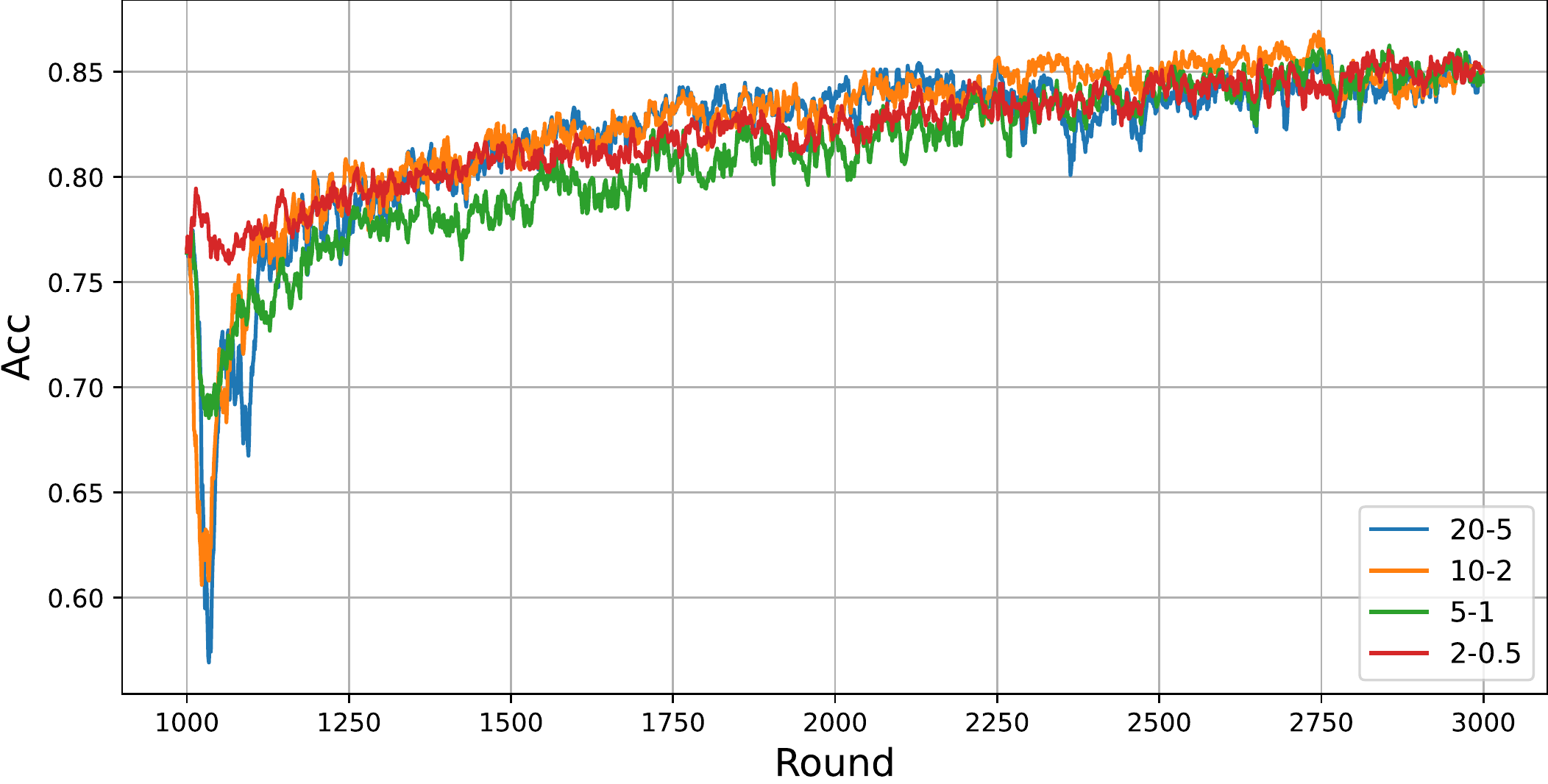}
    \label{fig:fig_gb_greater_acc}
  }
  \caption{Experiments of Gamma and Beta}
  \label{fig:fig_gb}
\end{figure}

To verify the effect of the parameters $\gamma$ and $\beta$ in ADVKD-ENH, we select Multi-Krum as aggregation method and conduct experiments on Resnet-18 with CIFAR10 dataset. We use ADVKD-ENH with different $\gamma$ and $\beta$ and a fixed $\alpha=0.7$ to launch attacks. The results of experiments are shown in Fig\ref{fig:fig_gb}.

We can find from Fig\ref{fig:fig_gb_less} that when we use lower $\gamma$ and $\beta$, the ASR would also decrease and get close to the ASR of ADVKD-REG.

On the contrary, when we select higher $\gamma$ and $\beta$, the ASR of model would become unstable and need more time to reach the highest ASR. We can find the reason in Fig\ref{fig:fig_gb_greater_count}. With $\gamma$ and $\beta$ getting higher, the distances between model updates from adversary and model updates from benign participants are getting larger, so it becomes less frequently selected. That is why the ASR grows slower and become unstable. However, there is one exception that when $\gamma=5$ and $\beta=1$, the ASR grows even faster than the ASR of $\gamma=2$ and $\beta=0.5$. If we only focus on the ASR, we can say the former performs better. Nevertheless, Fig\ref{fig:fig_gb_greater_acc} shows that when $\gamma$ and $\beta$ getting higher, the accuracy of model would drop in the beginning. There is no obvious drop of accuracy for $\gamma=2$ and $\beta=0.5$, but when $\gamma=5$ and $\beta=1$, the accuracy decreased 7\%. Hence, $\gamma=2$ and $\beta=0.5$ are suitable parameters that can not only keep high ASR but also prevent damaging the accuracy of model.

\paragraph*{Effect on Model Updates}

We now aim to more directly compare different backdoor attack methods' effects or contributions to the neurons or parameters related to the original task. So that we can explain why our attack enhances the stealth of backdoor attack and why the attack success. Here we also divide the dataset(EMNIST here) into 100 local datasets to simulate 100 participants. We use a pre-trained model as current global model, use different backdoor attack methods(Naive, ADVKD-ENH and ADVKD-REG) with local datasets to launch backdoor attacks on the model to generate model updates with backdoor. We also directly use the local datasets for local training to generate the model updates without backdoor.

To quantify and compare the contributions/effects of the model updates, we propose and use $UpdateGain$(Eq\ref{UpdateGain}) and $UpdateSignGain$(Eq\ref{UpdateSignGain}) as the metrics. Using a model update with backdoor and one without backdoor as the input, when the directions of updates are the same on some parameters, the result would become larger. Conversely, the result would become smaller when the directions are different. A larger result indicates the model update with backdoor also has a larger positive effect on original task. To avoid the effect of the heterogeneous between local datasets and the effect of the difference between global models, when evaluating the effect of backdoor, we limit the pair of model updates (with and without backdoor) using the same global model and the same local dataset in their local training procedures.

According to previous analysis, we focus on reducing the negative effect on the neurons or parameters which is important for original task. Hence, we only need to preserve the most important parameters (the parameters with larger absolute value) in the model, here we select the top-1000 parameters. For each model updates, we also only preserve the updates of the selected parameters.

\begin{equation}
  UpdateGain(\Delta\omega_{poison}, \Delta\omega_{clean}) = \sum_{i=1}^{\vert\Delta\omega\vert}{(\Delta\omega_{poison,i} * \Delta\omega_{clean,i})}
  \label{UpdateGain}
\end{equation}

\begin{equation}
  UpdateSignGain(\Delta\omega_{poison}, \Delta\omega_{clean}) = \sum_{i=1}^{\vert\Delta\omega\vert}{(Sign(\Delta\omega_{poison,i}) * Sign(\Delta\omega_{clean,i}))}
  \label{UpdateSignGain}
\end{equation}

where $\Delta\omega_{poison}$ and $\Delta\omega_{clean}$ are the model update with backdoor and the clean model update, $\Delta\omega_{poison,i}$ is the $i$-th value of $\Delta\omega_{poison}$, and $Sign()$ is a function returns the sign of the value.

\begin{figure}[htbp]
  \centering
  \subfloat[Update Gain of Naive]{
    \includegraphics[width=0.33\textwidth]{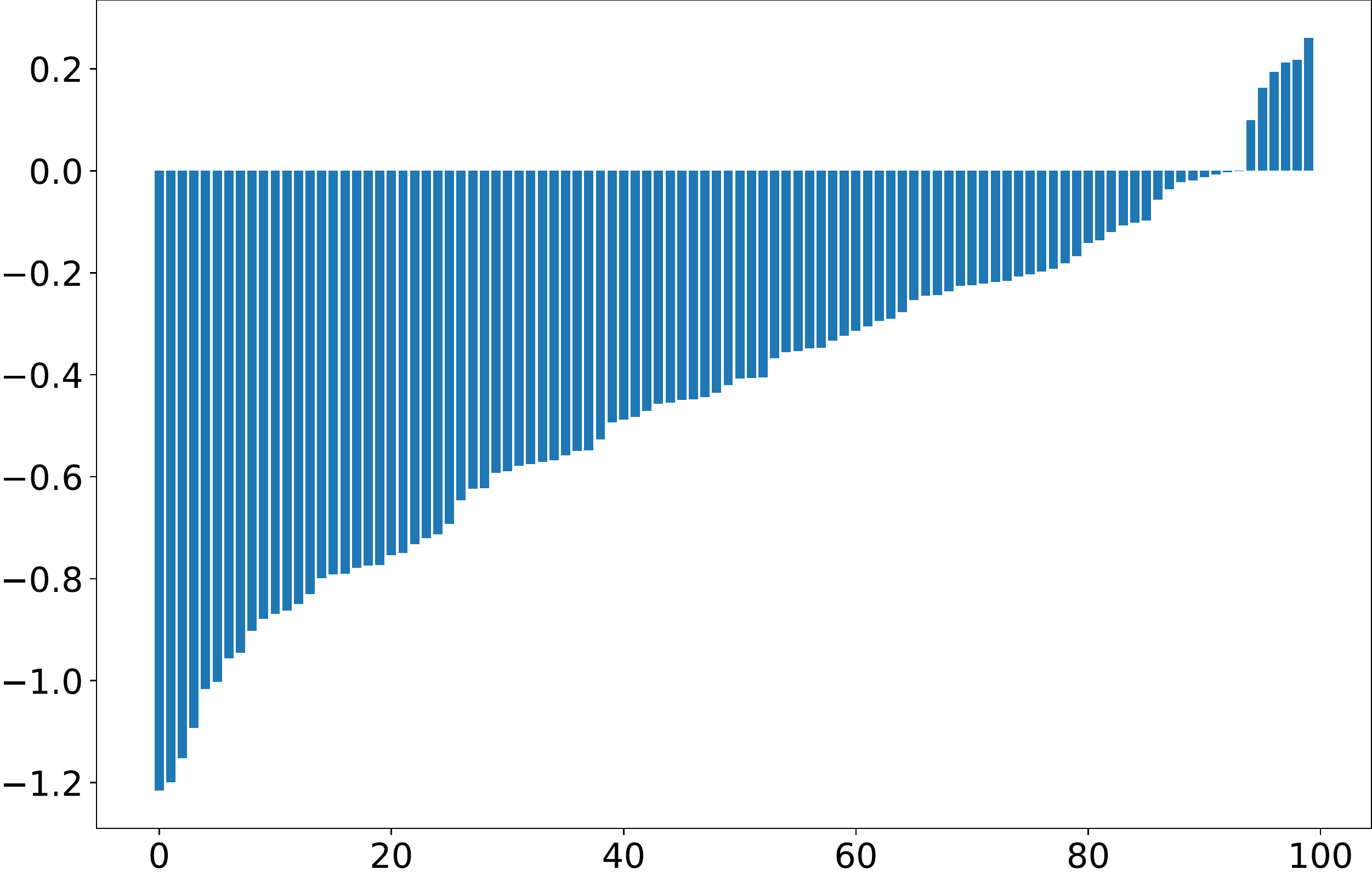}
    \label{fig:fig_gradient_gain_naive}
  }
  \subfloat[Update Gain of ADVKD-ENH]{
    \includegraphics[width=0.33\textwidth]{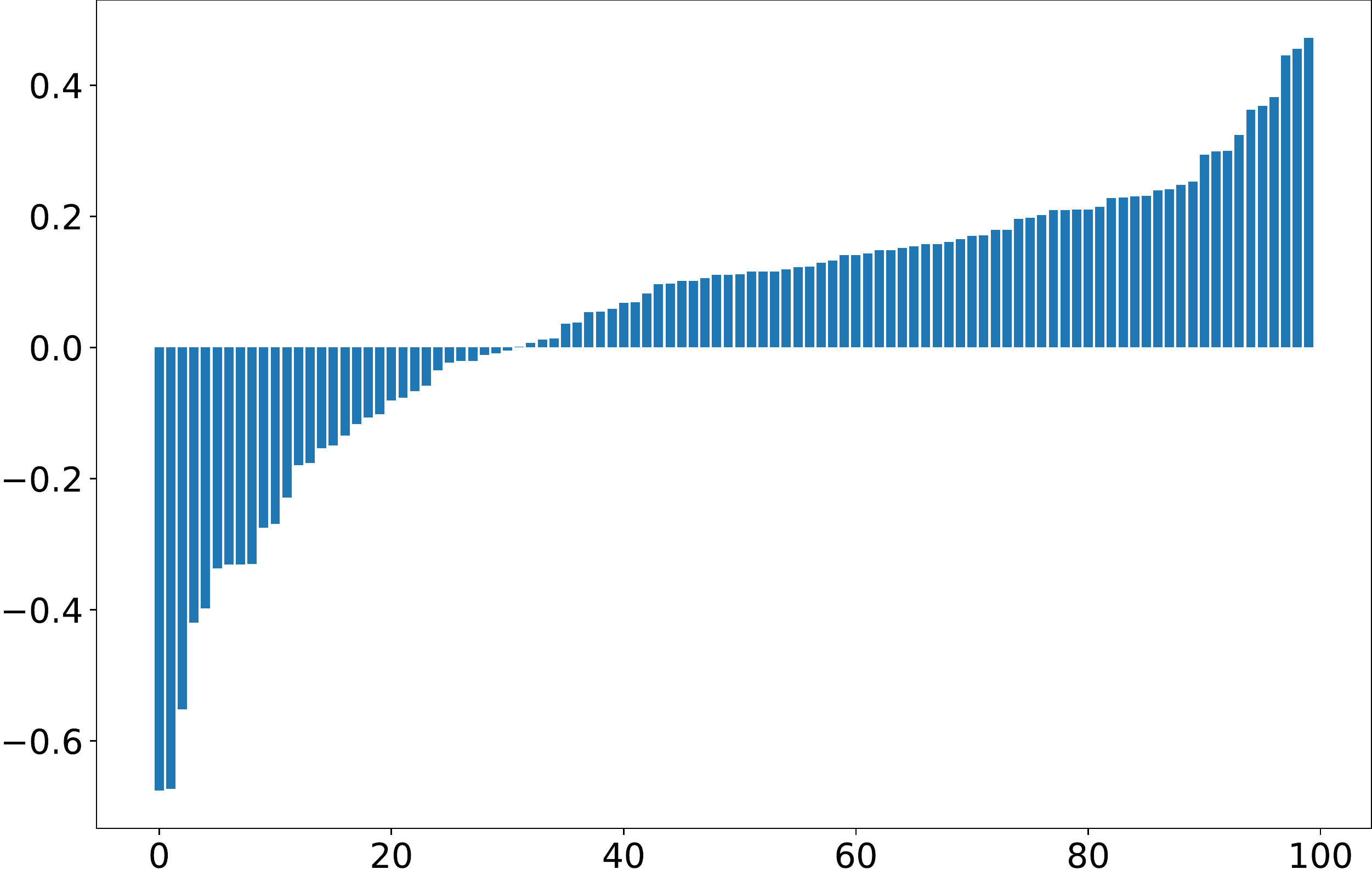}
    \label{fig:fig_gradient_gain_104}
  }
  \subfloat[Update Gain of ADVKD-REG]{
    \includegraphics[width=0.33\textwidth]{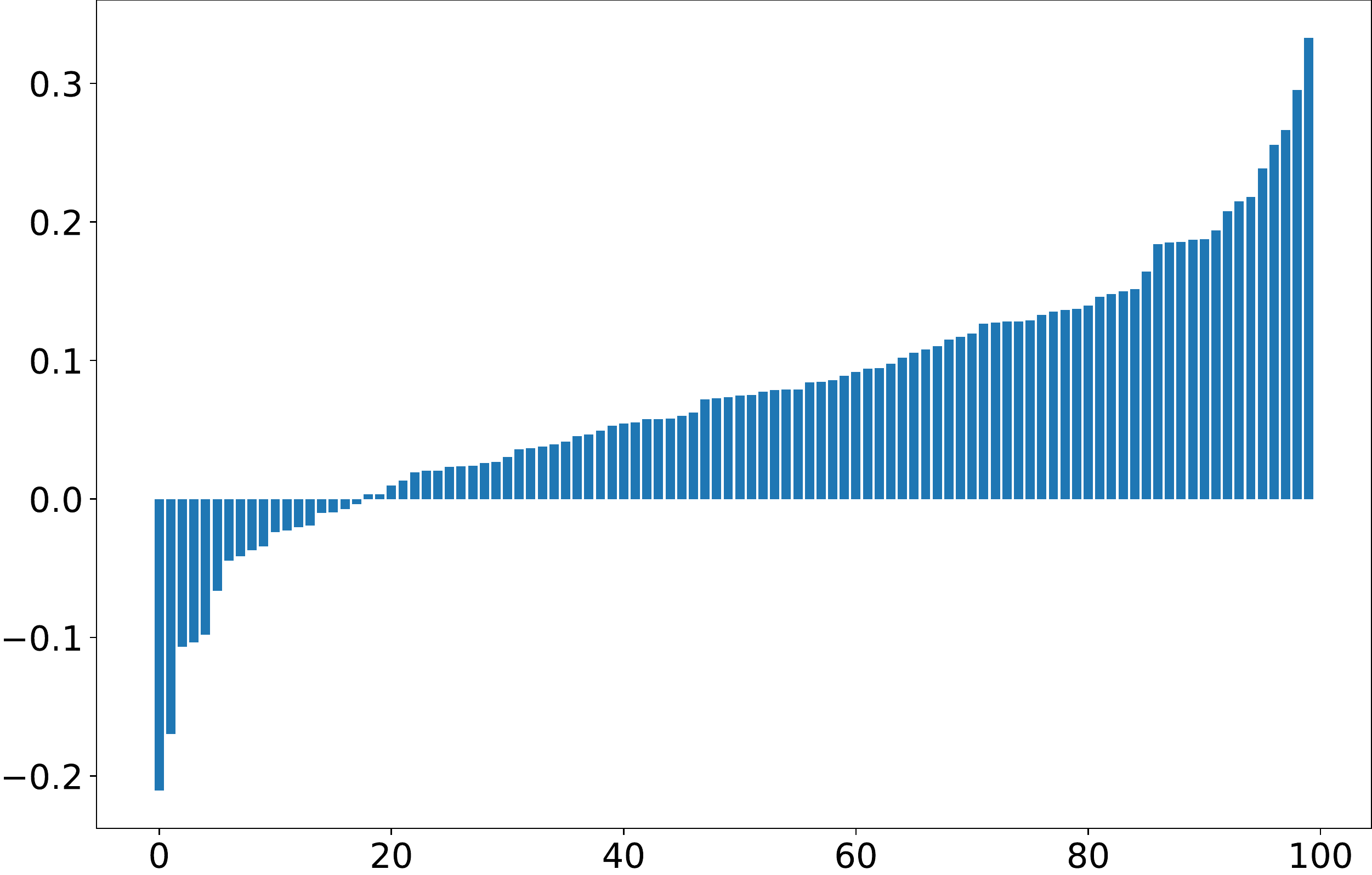}
    \label{fig:fig_gradient_gain_111}
  }

  \subfloat[Update Sign Gain of Naive]{
    \includegraphics[width=0.33\textwidth]{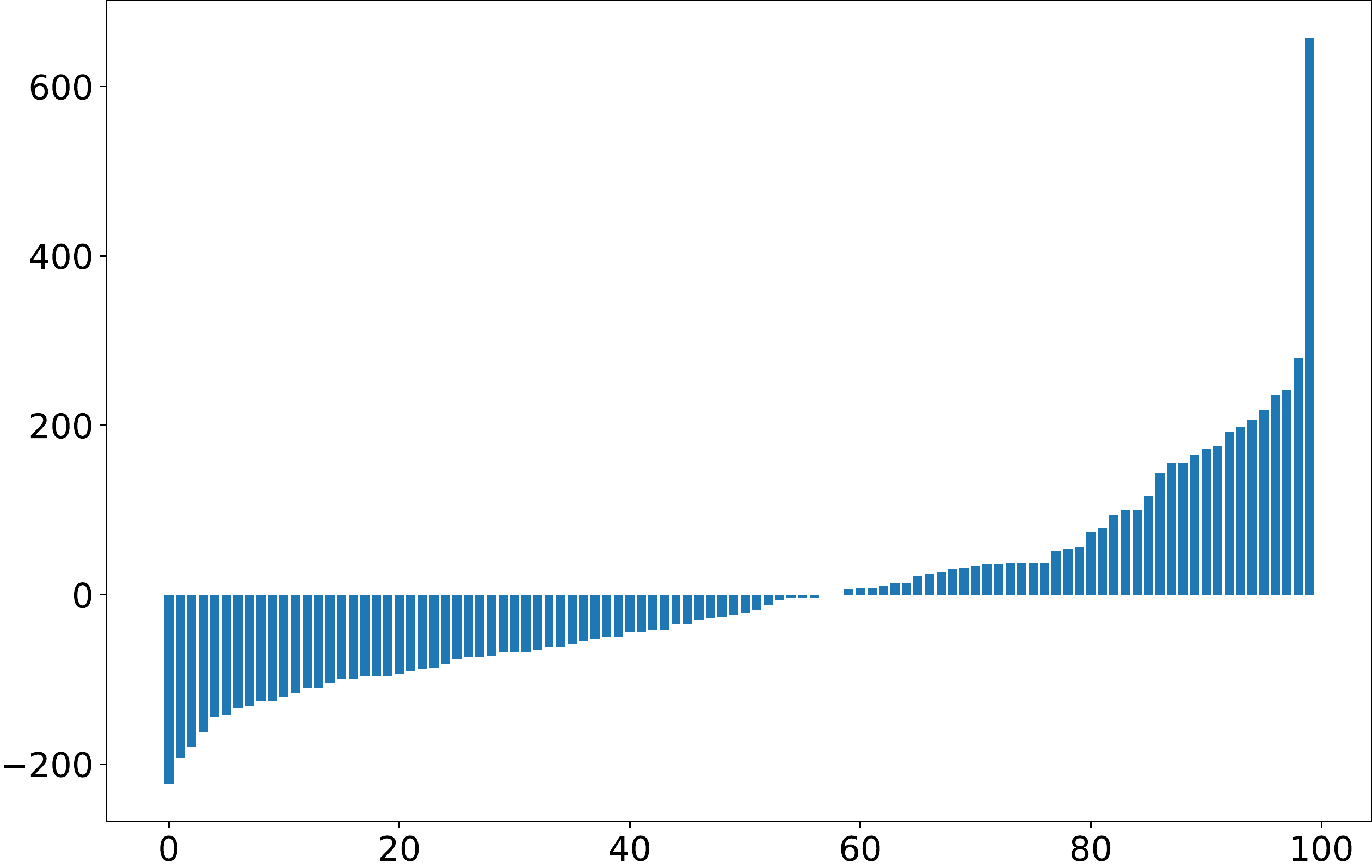}
    \label{fig:fig_gradient_gain_naive_s}
  }
  \subfloat[Update Sign Gain of ADVKD-ENH]{
    \includegraphics[width=0.33\textwidth]{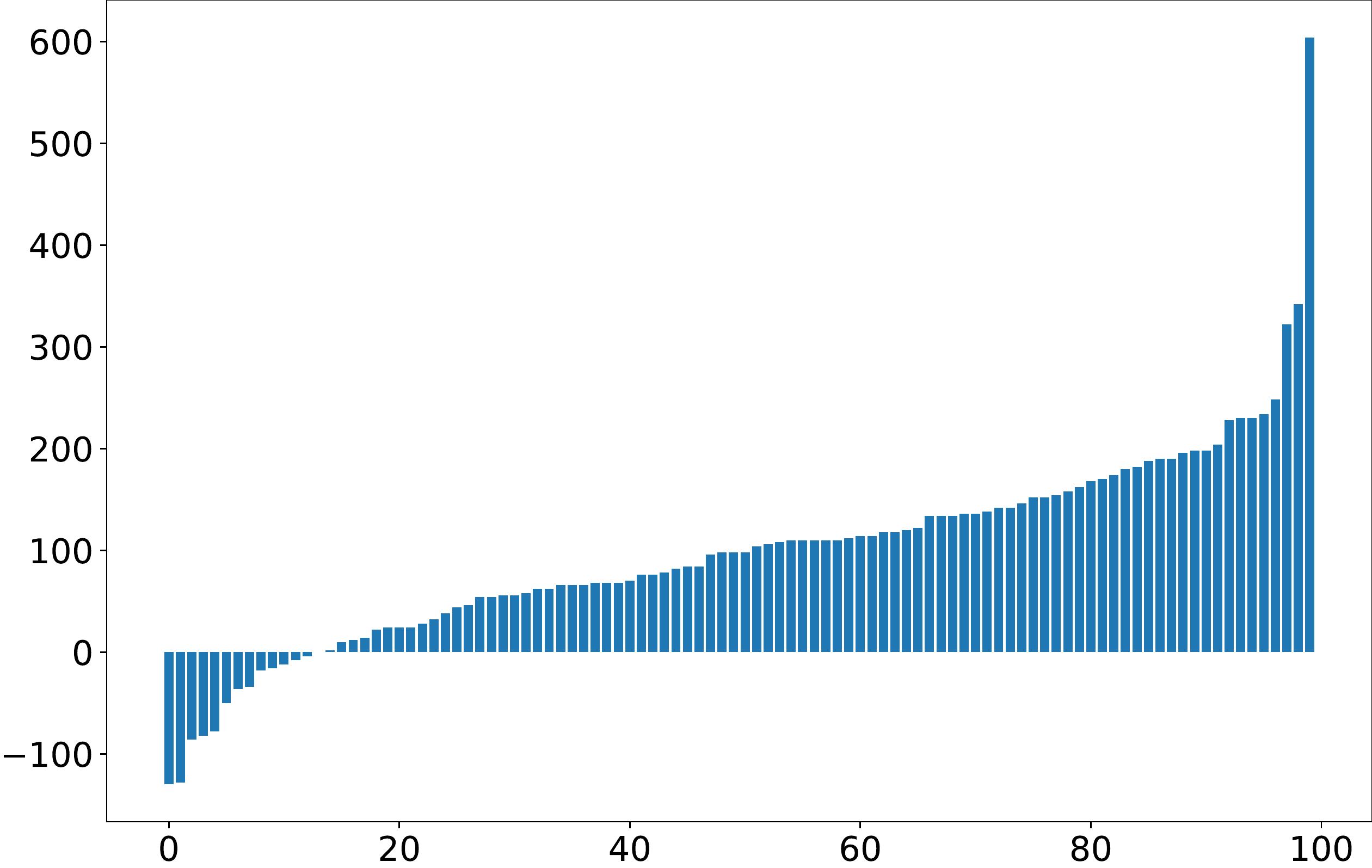}
    \label{fig:fig_gradient_gain_104_s}
  }
  \subfloat[Update Sign Gain of ADVKD-REG]{
    \includegraphics[width=0.33\textwidth]{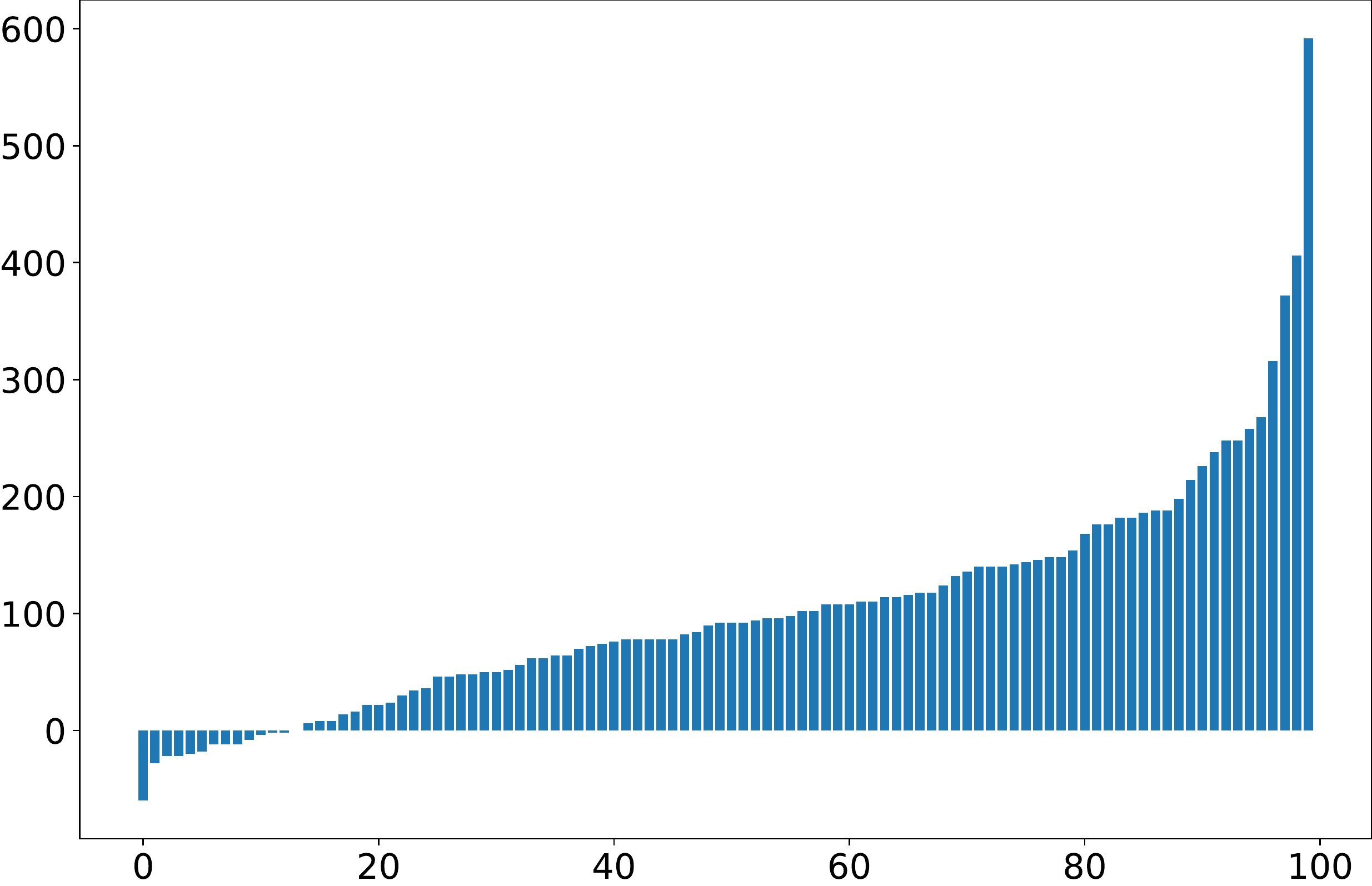}
    \label{fig:fig_gradient_gain_111_s}
  }
  \caption{Update Gain and Update Sign Gain}
  \label{fig:fig_gradient_gain}
\end{figure}

As we divided dataset into 100 local datasets, for each backdoor attack and each metric, we have 100 different results. The results of the same attack and the same metric are sorted in ascending order and shown in Fig\ref{fig:fig_gradient_gain}. We can see that under both $UpdateGain$ and $UpdateSignGain$, the model updates of Naive method tend to make a negative effect on the important parameters. Such phenomenon is reduced in ADVKD-ENH, and in ADVKD-REG most of the model updates tend to affect the model positively on the original task. The results indicate that ADVKD does can reduce the negative effect or the penalty on the important neurons or parameters, demonstrate that ADVKD can reduce the abnormal characteristics of the model updates with backdoor.

\paragraph{Effect on Model Activates}

Moreover, we can also compare the activations of the model under different backdoor attacks to observe the effect of different methods. In this experiment, we still use EMNIST dataset and CNN model with two convolutional layers followed by two fully
connected layers. Using many testing samples as the input of each model, and take the averaged activations (after max-pooling and ReLU activation function) of the second convolutional layer of these models, we can visualize and compare the difference among the models. As the activations of the samples with different label would vary greatly, we only select the samples with the same label.

\begin{figure}[htbp]
  \centering
  \subfloat[Comparing of Activations]{
    \includegraphics[width=0.7\textwidth]{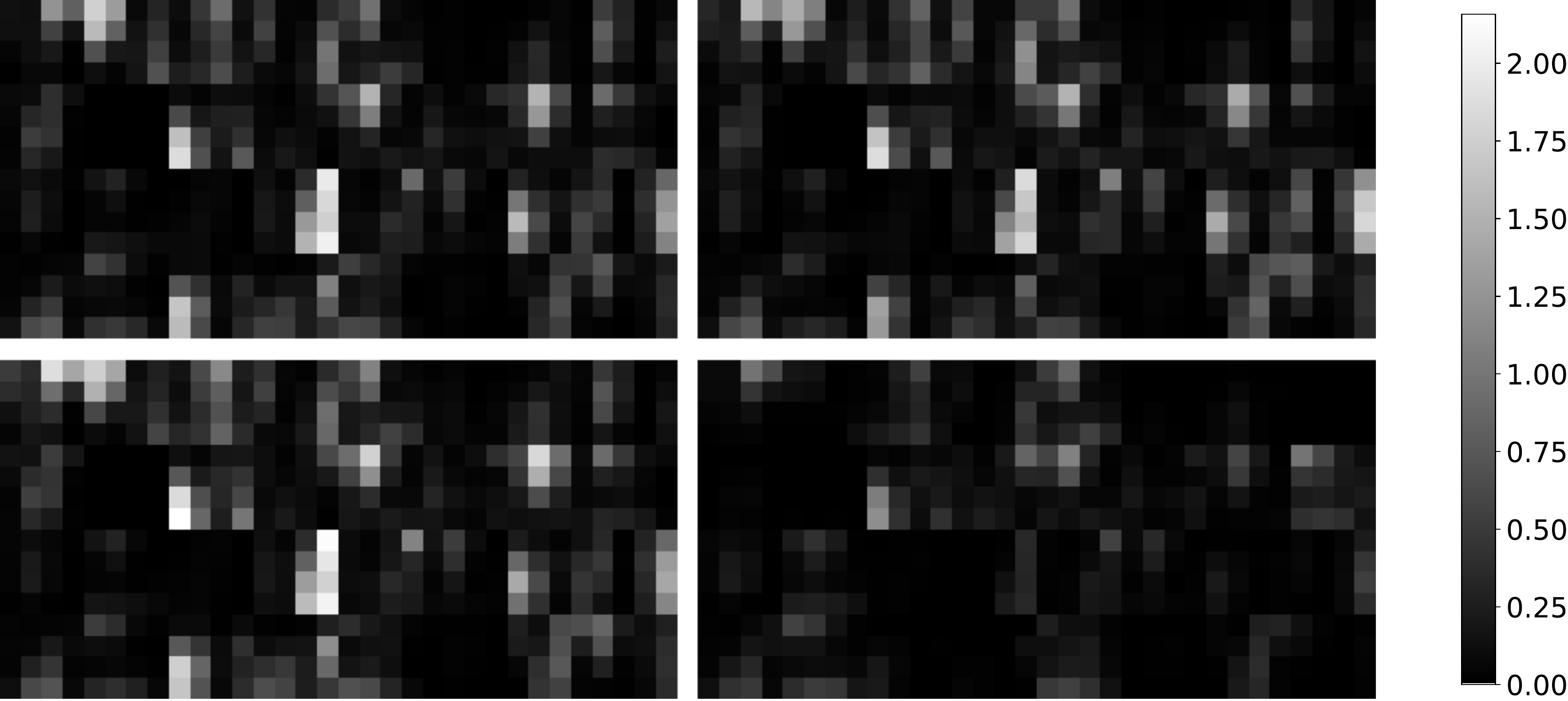}
    \label{fig:fig_activate_1_2}
  }

  \subfloat[Activations of Original Model]{
    \includegraphics[width=0.4\textwidth]{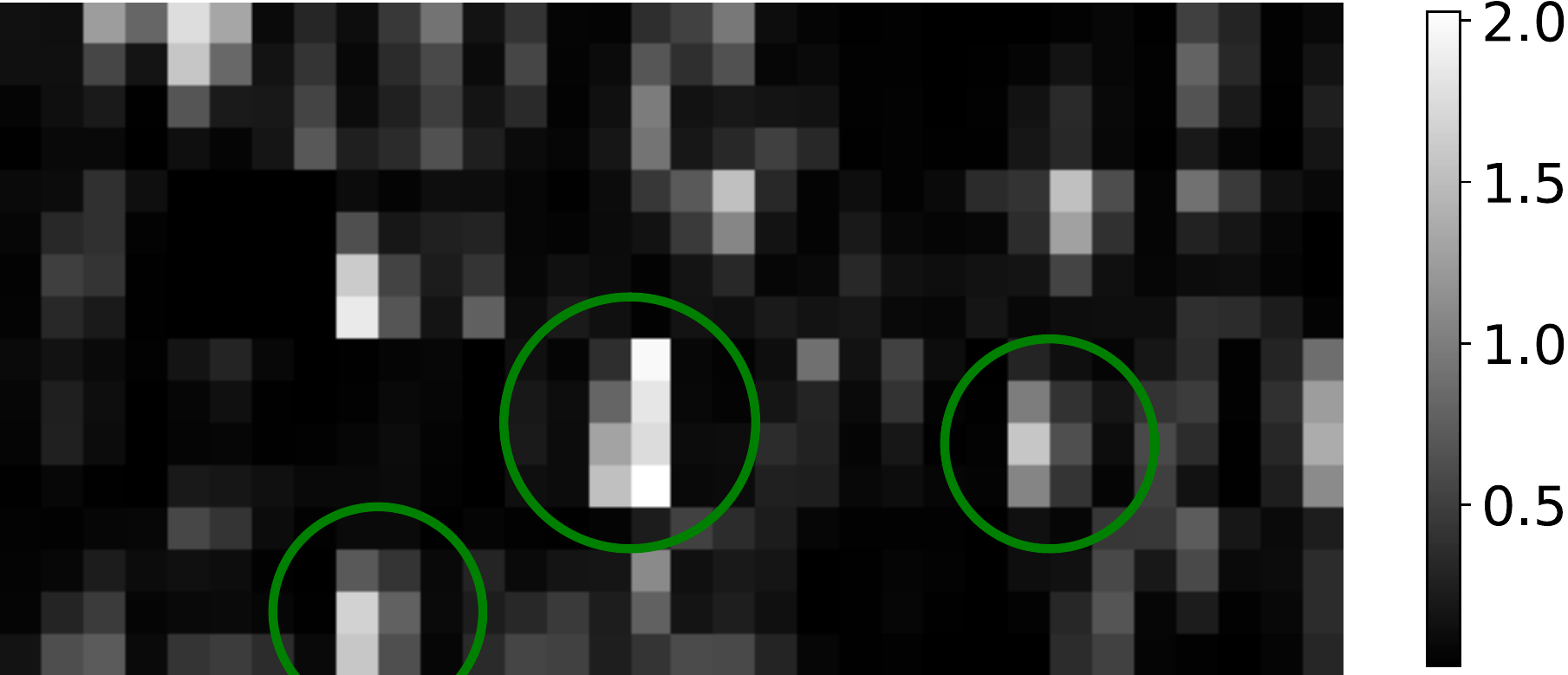}
    \label{fig:fig_activate_1_2_clean}
  }
  \subfloat[Activations of ADVKD-ENH]{
    \includegraphics[width=0.4\textwidth]{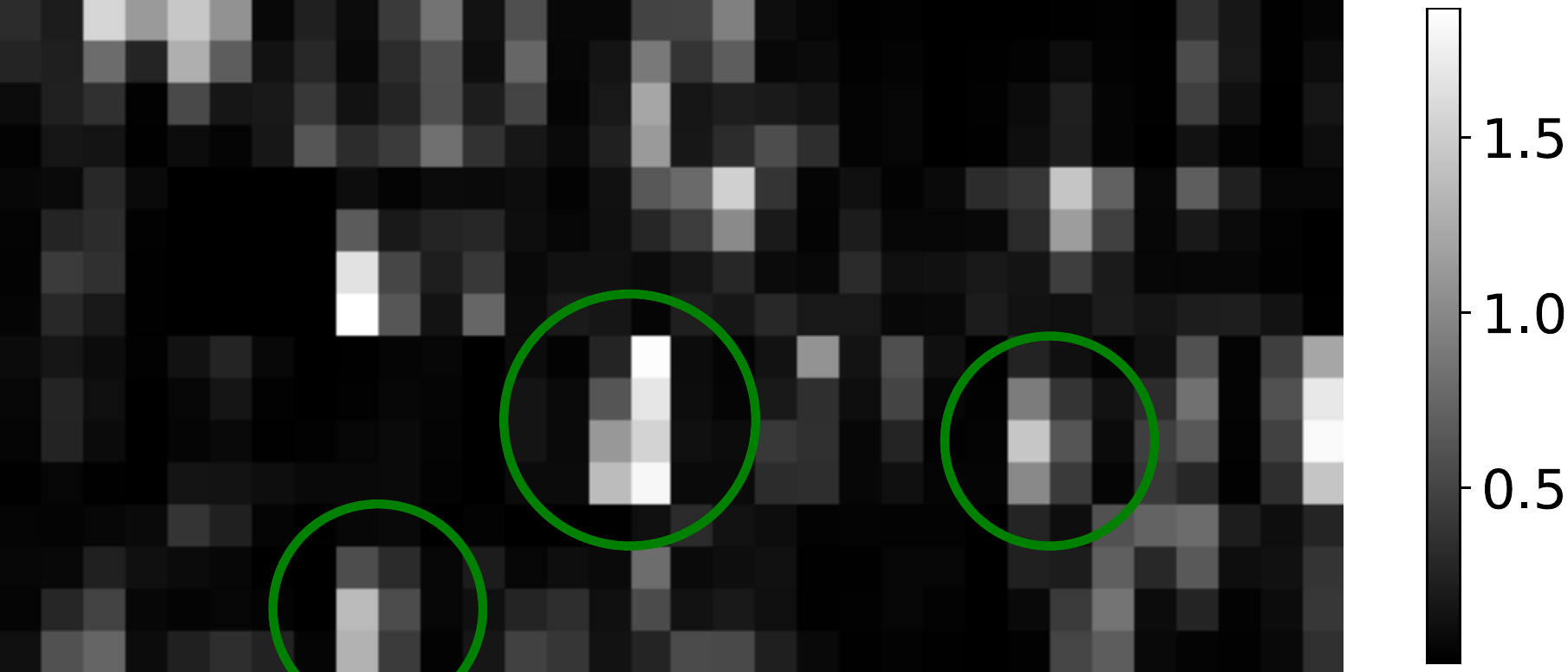}
    \label{fig:fig_activate_1_2_advkd104}
  }

  \subfloat[Activations of ADVKD-REG]{
    \includegraphics[width=0.4\textwidth]{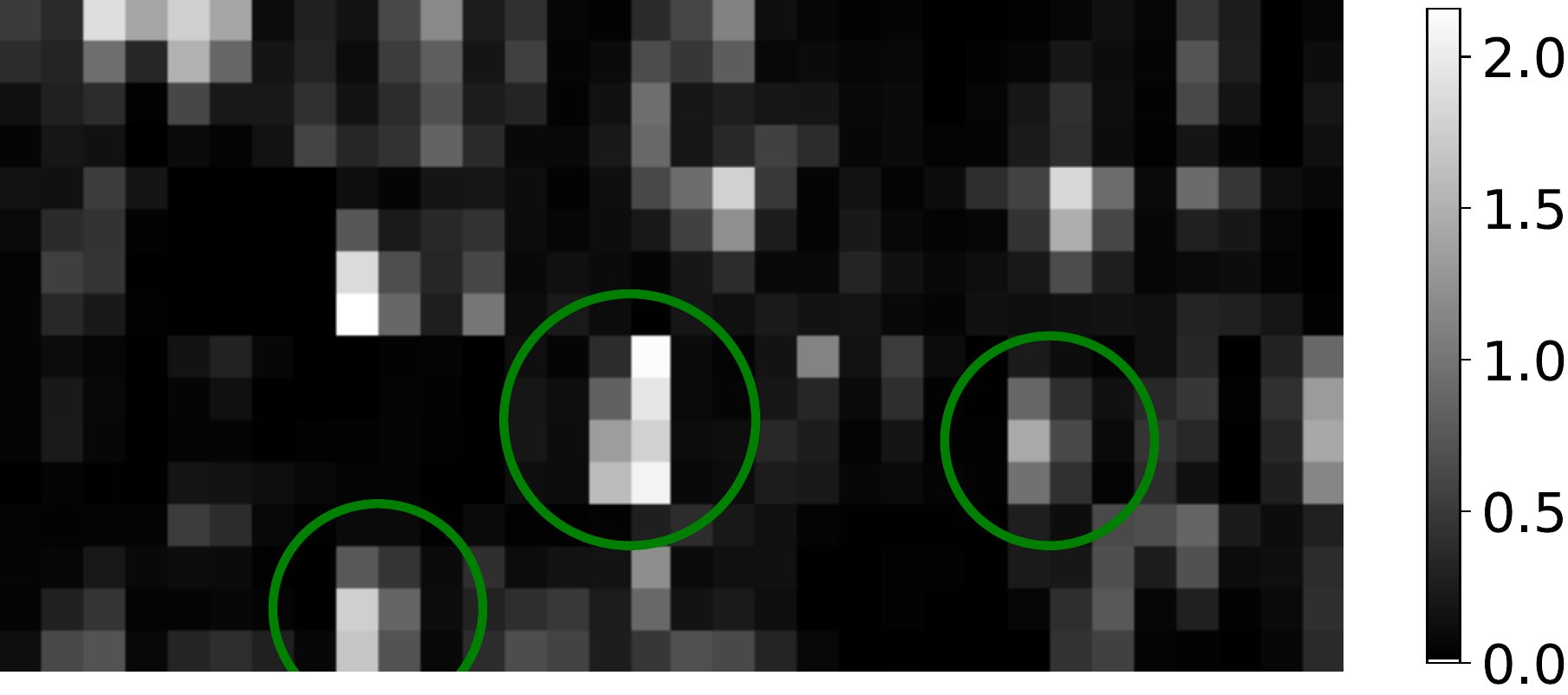}
    \label{fig:fig_activate_1_2_advkd111}
  }
  \subfloat[Activations of Naive]{
    \includegraphics[width=0.4\textwidth]{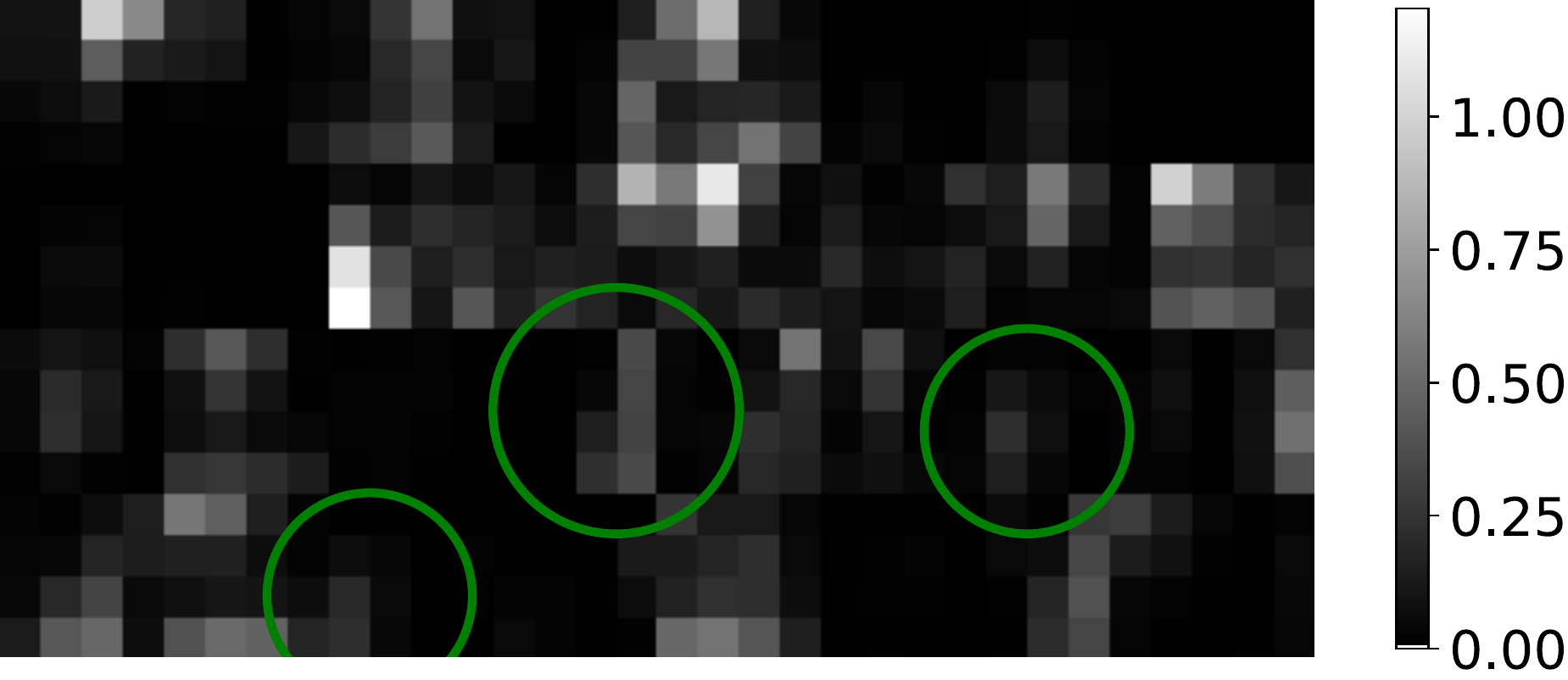}
    \label{fig:fig_activate_1_2_naive}
  }

  \caption{Activations of the Last Convolution Layer of a CNN}
  \label{fig:fig_activate}
\end{figure}

To remove the effect of different global model, we also use a pre-trained model. For each attack (Naive, ADVKD-ENH and ADVKD-REG), we simulate one round of training with adversarial participants to get a model with backdoor. Including the pre-trained model (without backdoor), we visualize and compare the activations of these models in Fig\ref{fig:fig_activate}. Fig\ref{fig:fig_activate_1_2} compares the activations of pre-trained model(upper left), ADVKD-ENH(upper right), ADVKD-REG(lower left) and Naive method(lower right). Comparing with the pre-trained model, ADVKD-ENH and ADVKD-REG only have small difference. However, in the model with the backdoor of Naive method, the activations show an obvious drop. Then, we separately display the activations of models in Fig\ref{fig:fig_activate_1_2_clean}, Fig\ref{fig:fig_activate_1_2_advkd104}, Fig\ref{fig:fig_activate_1_2_advkd111} and Fig\ref{fig:fig_activate_1_2_naive}. The activations of ADVKD-ENH and ADVKD-REG are still similar to the original pre-trained model and only have little difference. For the model of Naive method, on one hand we can find that even its activations are small, the structures of activations are still similar to the original pre-trained model, which means it can still successfully classify the clean samples. However, on the other hand, we can also find that some structures of activations in the original model are dimmed or even disappeared in Naive method, here we use green circles to mark some of such structures. Both results show that directly using backdoor attack would penalize the neurons or parameters which are important to original task, and the result also confirm that ADVKD can reduce such phenomenon.

\paragraph{Euclidean Distance and Cosine Distance}

\begin{figure}[hbtp]
  \centering
  \subfloat[Euclidean Distance of naive method]{
    \includegraphics[width=0.35\textwidth]{assets/update_dmat_ce_r1000_N100.pdf}
    \label{fig:fig_dmat_CE_comparing}
  }
  \subfloat[Cosine Distance of naive method]{
    \includegraphics[width=0.35\textwidth]{assets/update_cosmat_ce_r1000_N100.pdf}
    \label{fig:fig_cosmat_CE_comparing}
  }

  \subfloat[Euclidean Distance of ADVKD-ENH]{
    \includegraphics[width=0.35\textwidth]{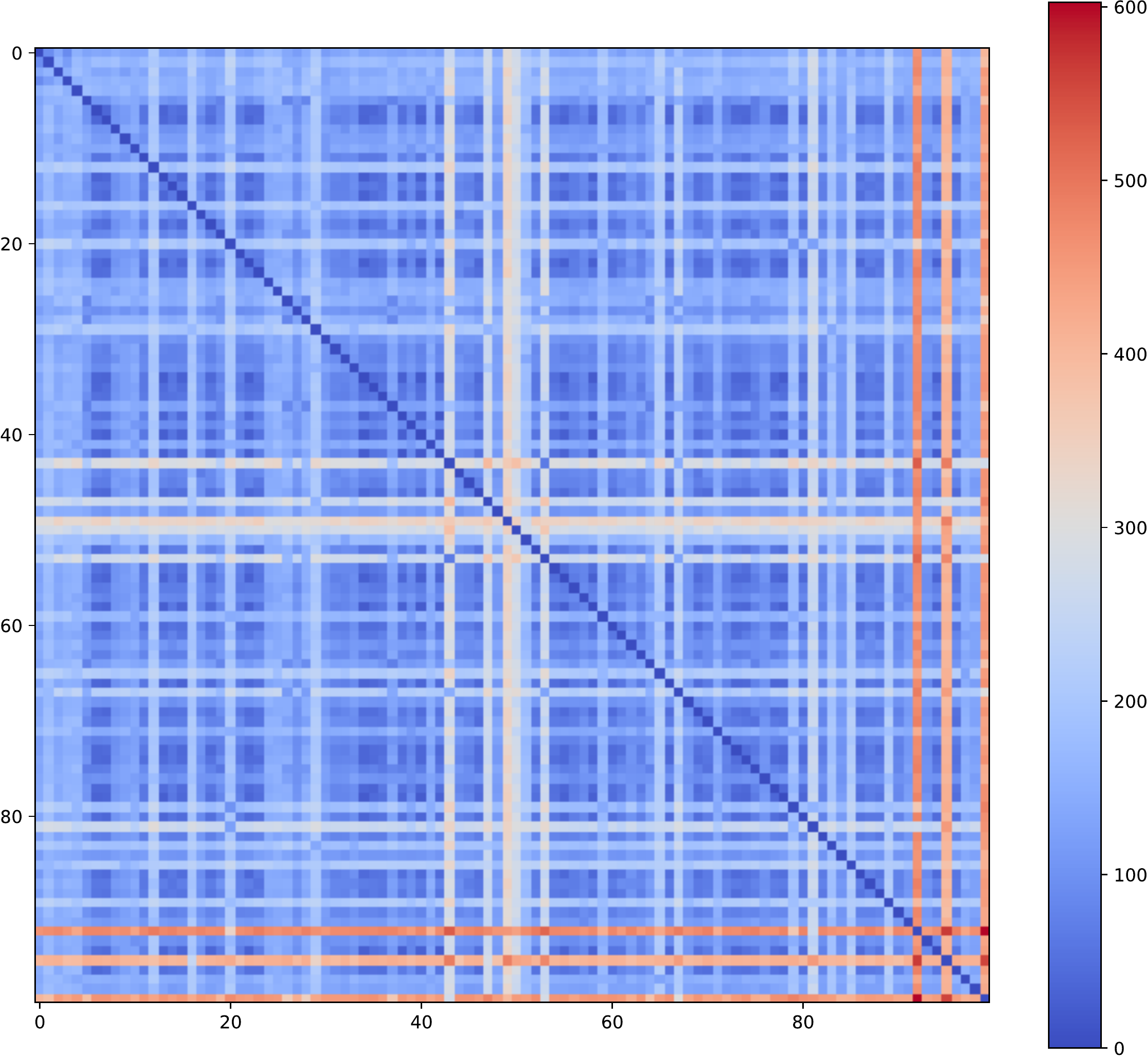}
    \label{fig:fig_dmat_KD73_comparing}
  }
  \subfloat[Cosine Distance of ADVKD-ENH]{
    \includegraphics[width=0.35\textwidth]{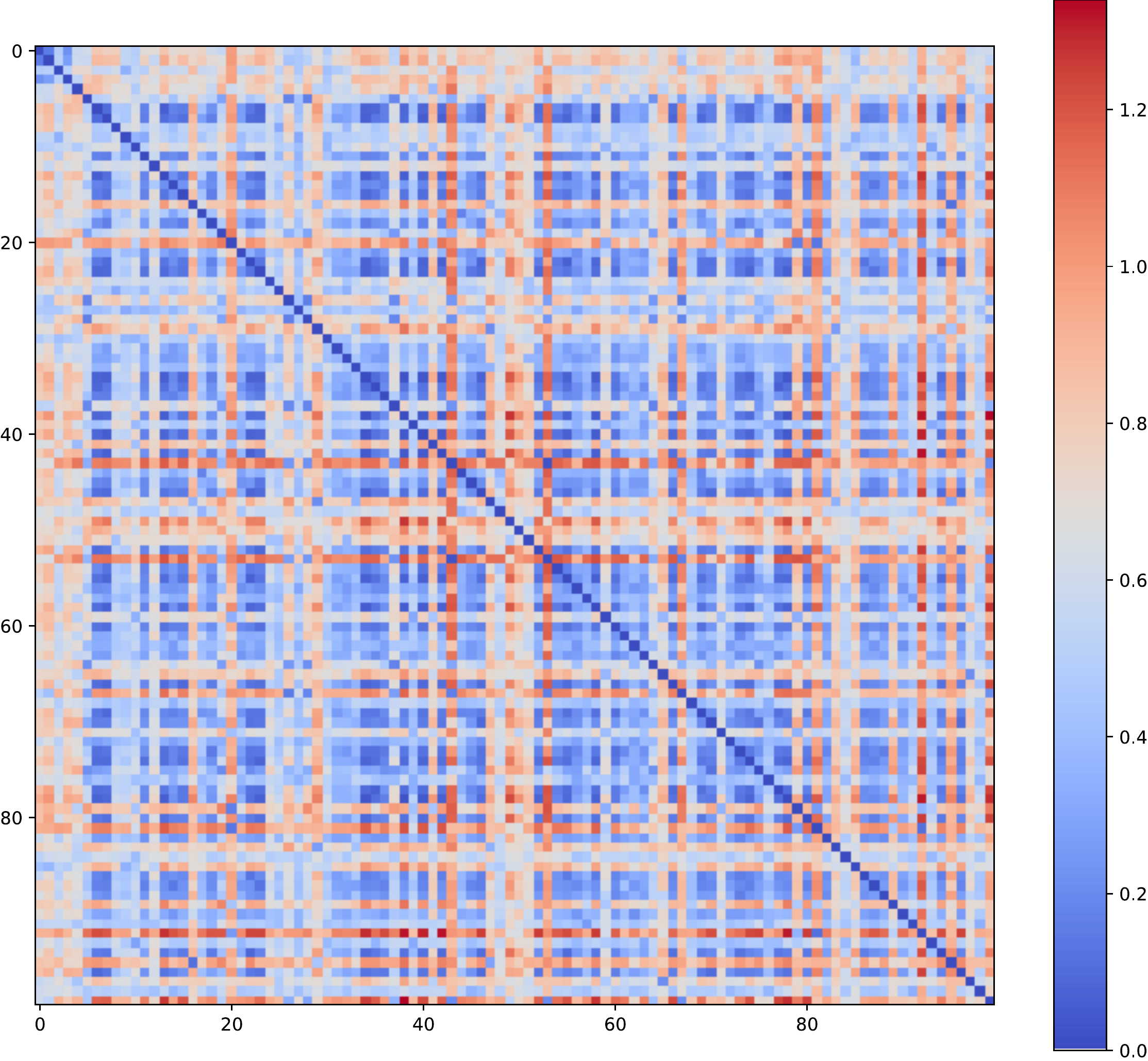}
    \label{fig:fig_cosmat_KD73_comparing}
  }
  \caption{Euclidean Distance and Cosine Distance of naive method and ADVKD}
  \label{fig:fig_mat_comparing}
\end{figure}

After evaluating the performance of ADVKD under different scenario, we look back on the problem of Euclidean Distance and Cosine Distance mentioned in the beginning. As mentioned above, when we launch a backdoor attack with naive method, we can find that there are large difference between the model update submitted by adversary and by benign participants, which are high Euclidean distance and cosine distance, so we proposed ADVKD. To compare with the result of naive method, we still conduct experiment with ResNet-18 on CIFAR10 dataset by simulating one round of federated learning with 4 adversaries and 96 benign participants and then calculate the Euclidean distance and cosine distance between updates. The results are shown in Fig\ref{fig:fig_mat_comparing}. Fig\ref{fig:fig_dmat_CE_comparing} and Fig\ref{fig:fig_cosmat_CE_comparing} are results of naive method, which are identical to Fig\ref{fig:fig_mat_CE}. Fig\ref{fig:fig_dmat_KD73_comparing} and Fig\ref{fig:fig_cosmat_KD73_comparing} are results of ADVKD-ENH with $\alpha=0.7$. Different from naive method, it's obvious that the Euclidean distance between model update of ADVKD adversary and benign participants is even smaller than the Euclidean distance between some benign outlier and others. And the cosine distance also gets smaller and become similar to benign one. So we can say that it's hard to discriminate between an update generated by ADVKD and one from benign participant.

According to the experiments above, we can find that with the dataset and model become more complex, it becomes harder to inject backdoor into global model. So the ADVKD-REG often fails as the restriction on backdoor is too strong. On the other hand, with the dataset and model become easier, it also becomes easier to inject backdoor into global model. However, the difference between a model with backdoor and other regular model also becomes larger. Hence, other backdoor attack and ADVKD-ENH may fail to pass the robust aggregation on server. Nevertheless, ADVKD-REG can not only pass the defense but also successfully inject backdoor into global model under this scenario. So, with an appropriate adjustment, ADVKD can launch successful backdoor attacks under different scenarios.

\section{Conclusion}

In this paper, we propose a novel backdoor attack(ADVKD) to inject backdoor into global model in FL. We first analyze why backdoor attack in FL would fail to pass the defending methods, and we find that directly flapping the label in dataset poisoning would cause the backdoor model different from a regular model too much and become an outlier. Inspired by this, we combine knowledge distillation with backdoor attack in FL and propose a novel backdoor attack method. By conducting experiments on three public datasets, we find that our proposed attack method can not only successfully inject backdoor into global model in FL when no defense applied, but also bypass the detection of defense method/robust aggregation method and inject backdoor even other baseline methods fail. We analyze the effect of the parameters of ADVKD to further explore the properties and features of ADVKD. We also use several ways to visualize the effect of different attacks to proof that ADVKD can reduce the abnormal characteristics in the model updates with backdoor. Our results suggest that ADVKD is a new powerful and stealthy backdoor attack in FL.


\bibliographystyle{unsrt}  
\bibliography{references}  

\end{document}